\documentclass[journal]{IEEEtran}

\usepackage{cite}
\usepackage{amsmath,amssymb,amsfonts}
\usepackage{textcomp}

\usepackage{booktabs}
\usepackage[pdftex]{graphicx}
\usepackage{amsmath,amsthm, amssymb}
\usepackage{mathtools}
\usepackage{url}
\usepackage{color}
\usepackage{multirow}

\usepackage{dsfont}
\usepackage{stmaryrd}
\usepackage{algorithm}
\usepackage{algorithmic}

 \usepackage[table,xcdraw]{xcolor}

\usepackage[caption=false,font=footnotesize]{subfig}

\usepackage{array}
\usepackage{caption}
\newcommand\cincludegraphics[2][]{\raisebox{-0.1\height}{\includegraphics[#1]{#2}}}

\newcommand{\changes}[1]{\textcolor[rgb]{0,0,0}{#1}}

\begin{document}

\title{An Introduction to Deep Morphological Networks}

\author{
	Keiller~Nogueira,
	Jocelyn~Chanussot, \textit{Fellow, IEEE},
	Mauro~Dalla~Mura, \textit{Senior, IEEE},\\
	%	William~Robson~Schwartz,
	Jefersson~A.~dos~Santos, \textit{Member, IEEE}%
	\thanks{
		K.~Nogueira is with the Department of Computer Science, Universidade Federal de Minas Gerais, Brazil, and also with Computing Science and Mathematics, University of Stirling, Stirling, FK9 4LA, Scotland, UK. Email: keiller.nogueira@stir.ac.uk.
		J.~A.~dos~Santos is with the Department of Computer Science, Universidade Federal de Minas Gerais, Brazil. Emails: jefersson@dcc.ufmg.br.
		\newline J.~Chanussot is with the Univ. Grenoble Alpes, Inria, CNRS, Grenoble INP, LJK, Grenoble, 38000, France. Email: jocelyn.chanussot@gipsa-lab.grenoble-inp.fr.
		\newline M.~Dalla~Mura is with University of Grenoble Alpes, CNRS, Grenoble INP, GIPSA-lab, 38000 Grenoble, France, and also with Tokyo Tech World Research Hub Initiative (WRHI), School of Computing, Tokyo Institute of Technology, Tokyo 152-8550, Japan. Email: mauro.dalla-mura@gipsa-lab.grenoble-inp.fr.
	}%
	
	\thanks{
		The authors thank FAPEMIG (APQ-00449-17), CNPq (grant \#424700/2018-2), and CAPES (grant \#88881.131682/2016-01, \#88881.145912/2017-01).
	}
}

\maketitle

\begin{abstract}
	Over the past decade, Convolutional Networks (ConvNets) have renewed the perspectives of the research and industrial communities. Although this deep learning technique may be composed of multiple layers, its core operation is the convolution, an important linear filtering process. Easy and fast to implement, convolutions actually play a major role, not only in ConvNets, but in digital image processing and analysis as a whole, being effective for several tasks. However, aside from convolutions, researchers also proposed and developed non-linear filters, such as operators provided by mathematical morphology. Even though these are not so computationally efficient as the linear filters, in general, they are able to capture different patterns and tackle distinct problems when compared to the convolutions. In this paper, we propose a new paradigm for deep networks where convolutions are replaced by non-linear morphological filters. Aside from performing the operation, the proposed Deep Morphological Network (DeepMorphNet) is also able to learn the morphological filters (and consequently the features) based on the input data. While this process raises challenging issues regarding training and actual implementation, the proposed DeepMorphNet proves to be able to extract features and solve problems that traditional architectures with standard convolution filters cannot.
\end{abstract}


\IEEEpeerreviewmaketitle

\section{Introduction}

Over the past decade, Convolutional Networks (ConvNet)~\cite{goodfellow2016deep} have been a game changer in the computer vision community, achieving state-of-the-art in several computer-vision applications, including image classification~\cite{krizhevsky2012imagenet,nogueira2017towards}, object and scene recognition~\cite{li2016visual,zhang2017beyond,lee2017going,zhang2018diverse,nogueira2018dynamic}, and many others.
Although this deep learning technique may be composed of several distinct components (such as convolutional and pooling layers, non-linear activation functions, etc), its core operation is the convolution, a linear filtering process whose weights, in this case, are to be learned based on the input data. %, an important procedure known as feature learning.
Easy and fast to implement, convolutions actually play a major role, not only in ConvNets~\cite{goodfellow2016deep}, but in digital image processing and analysis~\cite{jain1989fundamentals,szeliski2010computer} as a whole, being effective for many tasks (including image denoising~\cite{motwani2004survey}, edge detection~\cite{oskoei2010survey}, etc) and employed by several techniques (such as the filtering approaches~\cite{szeliski2010computer}).

\begin{figure}[t!]
	\centering
	\includegraphics[width=0.95\columnwidth]{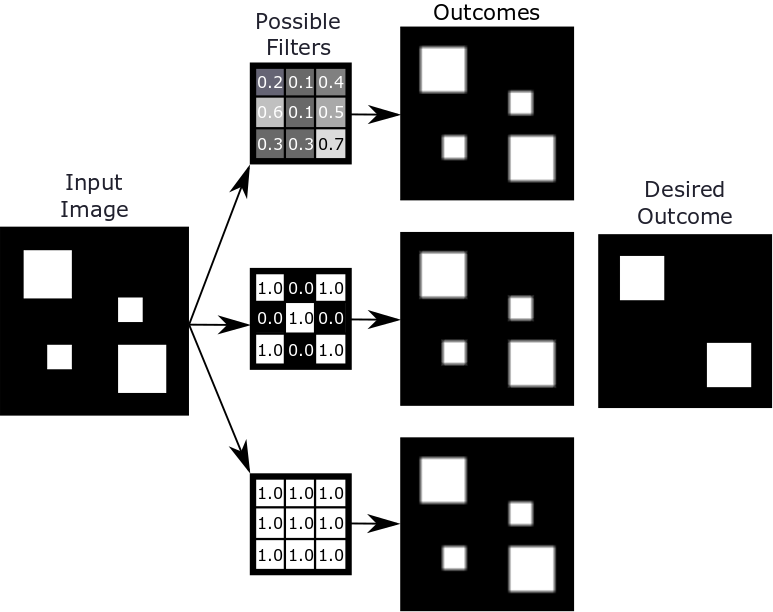}
	\caption{
		%Illustration showing several possible learned filters and outcomes generated by a ConvNet that is unsuccessfully trying to conceive a filter that produces an outcome similar to the expected one, which was produced using an $3\times3$ square structuring element.
		Illustration showing the lack of ability of the convolution filter to produce certain outcomes that are easily generated by non-linear operations.
		The goal here is to preserve only the larger squares of the input image, as presented in the desired outcome.
		Towards such objective, this image is processed by three distinct convolution filters producing different outputs, none of them similar to the desired outcome.
		However, a simple morphological opening with a $3\times3$ structuring element is capable of generating such output.
		% Note that none of the operations outcomes is similar to the desired output, which was generated using a morphological opening with a $3\times3$ structuring element, equal to the first filter employed by the ConvNet, in this example.
		% This shows that, regardless of the filter, the Convolutional Network is not able to produce the desired output mainly due to its core linear operations, i.e., the convolutions.
	}
	\label{fig:comparison_linear_non_linear}
\end{figure}

Aside from convolutions, researchers also proposed and developed non-linear filters, such as operators provided by mathematical morphology.
Even though these are not so computationally efficient as the linear filters, in general, they are able to capture different patterns and tackle distinct problems when compared to the convolutions.
For instance, suppose one desires to preserve only the large objects of an input image with $4\times 4$ and $2\times 2$ squares. 
As presented in Figure~\ref{fig:comparison_linear_non_linear}, despite having a myriad of possible configurations, the convolution is not able to produce such an outcome that can be easily obtained, for example, by a non-linear morphological opening.
% and, therefore, may not be able to produce the desired output.
%this non-linear filter is very unlikely to be obtained with a ConvNet.
%%In fact, because of its linear processing, ConvNets can not learn non-linear filters, as the previous one.
In fact, supported by this capacity of extracting distinct features, some non-linear filters, such as the morphological operations~\cite{serra2012mathematical}, are still very popular and state-of-the-art in some scenarios~\cite{fauvel2007spectral,xia2015random,kimori2016quantifying,seo2018morphological}.

In this paper, we propose a novel paradigm for deep networks where linear convolutions are replaced by the aforementioned non-linear morphological operations.
Furthermore, differently from the current literature, wherein distinct morphological filters must be evaluated in order to find the most suitable ones for each application, the proposed technique, called Deep Morphological Network (DeepMorphNet), learns the filters (and consequently the features) based on the input data.
Technically, the processing of each layer of the proposed approach can be divided into three steps/operations:
(i) depthwise convolution~\cite{chollet2017xception}, employed to rearrange the input pixels according to the binary filters, 
(ii) depthwise pooling, used to select some pixels and generate an eroded or dilated outcome, and
(iii) pointwise convolution~\cite{chollet2017xception}, employed to combine the generated maps producing one final morphological map (per neuron).
This process resembles the depthwise separable convolutions~\cite{chollet2017xception} but using binary filters and one more step (the second one) between the convolutions.
Note that, to the best of our knowledge, this is the first proof of concept work related to networks capable of performing and optimizing exact (non-approximate) morphological operations with flat filters.
Moreover, while this replacement process raises challenging issues regarding training and actual implementation, the proposed DeepMorphNet proves to be able to solve problems that traditional architectures with standard convolution filters cannot.

In practice, we can summarize the main contributions of this paper as follows:
\begin{itemize}
	\item a novel paradigm for deep networks where linear convolutions are replaced by the non-linear morphological operations, and
	\item a technique, called Deep Morphological Network (DeepMorphNet), capable of performing and optimization morphological operations.
\end{itemize}

The remainder of this paper is organized as follows. 
Section~\ref{sec:back_relwork} introduces some background concepts and presents the related work.
The proposed method is presented in Section~\ref{sec:methodology}. 
The experimental setup is introduced in Section~\ref{sec:experiments} while Section~\ref{sec:results} presents the obtained results. 
Finally, Section~\ref{sec:conclusion} concludes the paper.

\section{Background Knowledge and Related Work} \label{sec:back_relwork}

This section introduces the basic principles underlying mathematical morphology, and reviews the main approaches that exploit such operations for distinct image tasks.
% Section~\ref{sec:background} presents the background while Section~\ref{sec:relwork} discusses the related works.

\subsection{Mathematical Morphology} \label{sec:background}
% https://www.cs.auckland.ac.nz/courses/compsci773s1c/lectures/ImageProcessing-html/topic4.htm

Morphological operations, commonly employed in the image processing area, are strongly based on mathematical morphology.
%, systematically examined by Matheron and Serra in the 1960s and are an extension of Minkowski's set theory.
Since its introduction to the image domain, these morphological operations have been generalized from the analysis of a single band image to hyperspectral images made up of hundreds of spectral channels and has become one of the state-of-the-art techniques for a wide range of applications~\cite{serra2012mathematical}.
This study area includes several different operations (such as erosion, dilation, opening, closing, top-hats, and reconstruction), which can be applied to binary and grayscale images in any number of dimensions~\cite{serra2012mathematical}.

Formally, consider a grayscale 2D image $I(\cdot)$ as a mapping from the coordinates ($\mathds{Z}^2$) to the pixel-value domain ($\mathds{Z}$).
%: $I: \mathds{Z}^2 \shortrightarrow \mathds{Z}$.
% Most morphological transformations first require the definition of a structuring element (SE) to, then, process the input image $I$.
Most morphological transformations process this input image $I$ using a structuring element (SE) (usually defined prior to the operation).
A flat\footnote{A flat SE is binary and only defines which pixels of the neighborhood should be taken into account. On the other hand, a non-flat SE contains finite values used as additive offsets in the morphological computation.} SE $B(\cdot)$ can be defined as a function that, based on its format (or shape), returns a set of neighbors of a pixel $(i,j)$.
This neighborhood of pixels is taken into account during the morphological operation, i.e., while probing the image $I$.
%Normally, a SE is defined by two components:
%(i) shape, which is usually a discrete representation of continuous shapes, such as square, circle,
%(ii) center, that identifies the pixel on which the SE is superposed when probing the image.
% Figure~\ref{fig:se_examples} presents some examples of common SEs employed in the literature.
As introduced, the definition of the SE is of vital importance for the process to extract relevant features.
However, in literature~\cite{liu2017multimorphological,wang2018lidar}, this definition is performed experimentally (with common shapes being squares, disks, diamonds, crosses, and x-shapes), an expensive process that does not guarantee a good descriptive representation.
%gu2016nonlinear

%\newcommand{\seImageSize}{0.08}
%
%\begin{figure}[h!]
%	\centering
%	\scriptsize
%	\subfloat[Square]{
%		\includegraphics[width=\seImageSize\textwidth, keepaspectratio=true]{compared_nonLinear_linear/se_square.png}
%	}
%	\hspace{1mm}
%	\subfloat[Disk]{
%		\includegraphics[width=\seImageSize\textwidth, keepaspectratio=true]{compared_nonLinear_linear/se_disk.png}
%	}
%	\hspace{1mm}
%	\subfloat[Diamond]{
%		\includegraphics[width=\seImageSize\textwidth, keepaspectratio=true]{compared_nonLinear_linear/se_diamond.png}
%	}
%	\hspace{1mm}
%	\subfloat[Cross]{
%		\includegraphics[width=\seImageSize\textwidth, keepaspectratio=true]{compared_nonLinear_linear/se_cross.png}
%	}
%	\hspace{1mm}
%	\subfloat[X shape]{
%		\includegraphics[width=\seImageSize\textwidth, keepaspectratio=true]{compared_nonLinear_linear/se_x_shape.png}
%	}
%	\caption{Examples of common structuring elements employed in the literature.}
%	\label{fig:se_examples}
%\end{figure}

%After the definition of the SE, morphological operations can be carried out.
After its definition, the SE can be then employed in several morphological processes. %, from basic to complex ones.
Most of these operations are usually supported by two basic morphological transformations: \textit{erosion} $\mathcal{E}(\cdot)$ and \textit{dilation} $\mathcal{\delta}(\cdot)$.
%(which are the basic procedures of morphology)
%Erosion and dilation are fundamental neighborhood transformations in mathematical morphology. 
%%an input image by giving as output for each pixel the compression or expansion of the intensity values of the set of pixels included by the structuring element when it is centered on that pixel.
%%It is important to note that compression and the expansion are the minimum and maximum values, respectively.
Such operations receive basically the same input: an image $I$ and the SE $B$.
While erosion transformations process each pixel $(i,j)$ using the supremum
% (the smallest upper bound, $\wedge$) 
function $\wedge$, as denoted in Equation~\ref{eq:erosion}, the dilation operations process the pixels using the infimum $\vee$
% (the greatest lower bound, $\vee$) 
function, as presented in Equation~\ref{eq:dilation}.
Intuitively, these two operations probe an input image using the SE, i.e., they position the structuring element at all possible locations in the image and analyze the neighborhood pixels.
This process, somehow similar to the convolution procedure, outputs another image with regions compressed or expanded. % (depending on the operation).
Some examples of erosion and dilation are presented in Figure~\ref{fig:ucmerced_mp}, in which it is possible to notice the behavior of each operation.
As can be noticed erosion affects brighter structures while dilation influences darker ones (w.r.t. the neighborhood defined by the SE).

\begin{eqnarray} \label{eq:erosion} % \bigvee
	%\mathcal{E} (I, B) \wedge_{b \in B} I_{-b} -- \underset{p \in I}{\forall}
	%\mathcal{E}(B, I) =  \forall_{p \in I} \{\wedge I(p') | p' \in B(p) \cup I(p)\}
	\mathcal{E}(B, I)_{(i,j)} =  \{\wedge I((i,j)') | (i,j)' \in B(i,j) \cup I(i,j)\}
\end{eqnarray}
\begin{eqnarray} \label{eq:dilation}
	%\mathcal{\delta}_B (I) \vee_{b \in B} I_{-b}
	\mathcal{\delta}(B, I)_{(i,j)} = \{\vee I((i,j)') | (i,j)' \in B(i,j) \cup I(i,j)\}
\end{eqnarray}

If we have an ordered set, then the erosion and dilation operations can be simplified.
This is because the infimum and the supremum are respectively equivalent to the minimum and maximum functions when dealing with ordered sets.
In this case, erosion and dilation can be defined as presented in Equations~\ref{eq:erosion_orderedset} and~\ref{eq:dilation_orderedset}, respectively.

\begin{eqnarray} \label{eq:erosion_orderedset} % \bigvee
	\mathcal{E}(B, I)_{(i,j)} = \{\min_{(i,j)'\in B(i,j)} (I((i,j)'))\}
\end{eqnarray}
\begin{eqnarray} \label{eq:dilation_orderedset}
	\mathcal{\delta}(B, I)_{(i,j)} = \{\max_{(i,j)'\in B(i,j)} (I((i,j)'))\}
\end{eqnarray}

%As introduced, two operations are available: opening and closing.
%The first one is a sequential composition of erosion and dilation while the latter is composed by dilations followed by erosions.
Based on these two fundamental transformations, other more complex morphological operations may be computed.
The morphological \textit{opening}, denoted as $\mathcal{\gamma}(\cdot)$ and defined in Equation~\ref{eq:opening}, is simply an erosion operation followed by the dilation (using the same SE). % of the eroded output.
In contrast, a morphological \textit{closing} $\mathcal{\varphi}(\cdot)$ of an image, defined in Equation~\ref{eq:closing}, is a dilation followed by the erosion (using the same SE). % of the dilated output.
%https://books.google.com.br/books?id=kIMlNp50wW0C&pg=PA127&lpg=PA127&dq=an+opening+flattens+bright+objects+that+are+smaller+than+the+size+of+the+SE+and,+because+of+the+dilation,+mostly+preserves+the+bright+large+areas&source=bl&ots=kjx5l6Pc1R&sig=ACfU3U0VBqPIwtpbksVFNkTydl0EAXLHtQ&hl=pt-BR&sa=X&ved=2ahUKEwiKmrKm-IvhAhXPGLkGHd_uARQQ6AEwAHoECAkQAQ#v=onepage&q=an%20opening%20flattens%20bright%20objects%20that%20are%20smaller%20than%20the%20size%20of%20the%20SE%20and%2C%20because%20of%20the%20dilation%2C%20mostly%20preserves%20the%20bright%20large%20areas&f=false
%Optical Remote Sensing: Advances in Signal Processing and Exploitation ... p. 127
% Intuitively, while an erosion would affect all brighter structures, 
Intuitively, an opening flattens bright objects that are smaller than the size of the SE and, because of dilation, mostly preserves the bright large areas.
A similar conclusion can be drawn for darker structures when closing is performed.
Examples of this behavior can be seen in Figure~\ref{fig:ucmerced_mp}.
It is important to highlight that by using erosion and dilation transformations, opening and closing perform geodesic reconstruction in the image.
Operations based on this paradigm belongs to the class of filters that operate only on connected components (flat regions) and cannot introduce any new edge to the image.
Furthermore, if a segment (or component) in the image is larger than the SE then it will be unaffected, otherwise, it will be merged to a brighter or darker adjacent region depending upon whether a closing or opening is applied.
This process is crucial because it avoids the generation of distorted structures, which is obviously an undesirable effect.

\newcommand{\exMorphFigSize}{0.09}

\begin{figure}[h]
	\centering
	\scriptsize
	\subfloat[Original]{
		\includegraphics[width=\exMorphFigSize\textwidth]{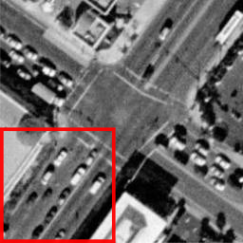}
	}
	\subfloat[Original Highlight]{
		 % trim={<left> <lower> <right> <upper>}
		\includegraphics[trim={0 0 20em 20em} ,clip, width=\exMorphFigSize\textwidth]{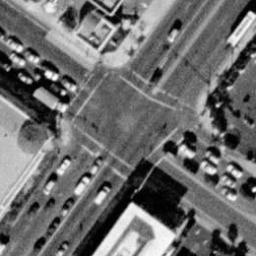}
	}
	%\hspace{1mm}
	\subfloat[Erosion]{
		\includegraphics[trim={0 0 20em 20em} ,clip, width=\exMorphFigSize\textwidth]{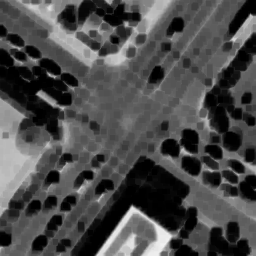}
	}
	\subfloat[Dilation]{
		\includegraphics[trim={0 0 20em 20em} ,clip, width=\exMorphFigSize\textwidth]{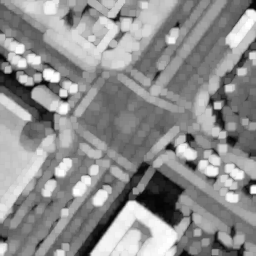}
	}
	%\hspace{1mm}
	\subfloat[Opening]{
		\includegraphics[trim={0 0 20em 20em} ,clip, width=\exMorphFigSize\textwidth]{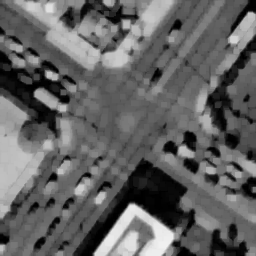}
	}
	\hspace{1mm}
	\subfloat[Closing]{
		\includegraphics[trim={0 0 20em 20em} ,clip, width=\exMorphFigSize\textwidth]{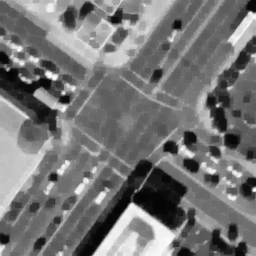}
	}
	\subfloat[White Top-hat]{
		\includegraphics[trim={0 0 20em 20em} ,clip, width=\exMorphFigSize\textwidth]{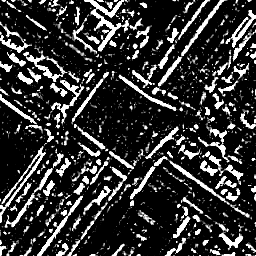}
	}
	\subfloat[Black Top-hat]{
		\includegraphics[trim={0 0 20em 20em} ,clip, width=\exMorphFigSize\textwidth]{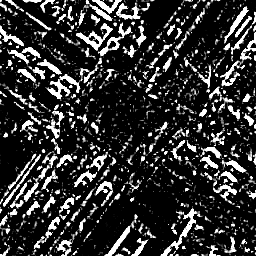}
	}
	%\hspace{1mm}
	\subfloat[Closing by Reconstruction]{
		\includegraphics[trim={0 0 20em 20em} ,clip, width=\exMorphFigSize\textwidth]{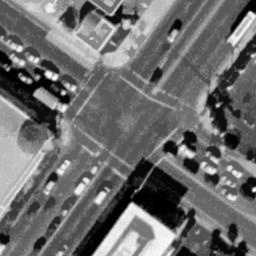}
	}
	\subfloat[Opening by Reconstruction]{
		\includegraphics[trim={0 0 20em 20em} ,clip, width=\exMorphFigSize\textwidth]{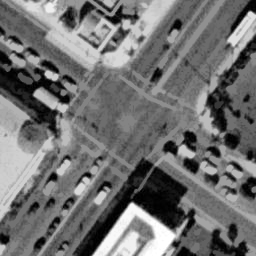}
	}
	\caption{Examples of morphological images generated for the UCMerced Land-use Dataset.
		For better viewing and understanding, images (b)-(j) only present a left-bottom zoom of the original image (a).
		All these images were processed using a $5\times5$ square as structuring element.}	
	\label{fig:ucmerced_mp}
\end{figure}

\begin{eqnarray} \label{eq:opening}
	\mathcal{\gamma}(B, I) = \mathcal{\delta}(B, \mathcal{E}(B, I))
\end{eqnarray}
\begin{eqnarray} \label{eq:closing}
	\mathcal{\varphi}(B, I) = \mathcal{E}(B, \mathcal{\delta}(B, I))
\end{eqnarray}

Other important morphological operations are the \textit{top-hats}.
Top-hat transform is an operation that extracts small elements and details from given images.
There are two types of top-hat transformations:
(i) the white one $\mathcal{T}^w(\cdot)$, defined in Equation~\ref{eq:white_top_hat}, in which the difference between the input image and its opening is calculated, and
(ii) the black one, denoted as $\mathcal{T}^b(\cdot)$ and defined in Equation~\ref{eq:black_top_hat}, in which the difference between the closing and the input image is performed.
White top-hat operation preserves elements of the input image brighter than their surroundings but smaller than the SE.
On the other hand, black top-hat maintains objects smaller than the SE with brighter surroundings.
Examples of these two operations can be seen in Figure~\ref{fig:ucmerced_mp}.

\begin{eqnarray} \label{eq:white_top_hat}
	\mathcal{T}^w(B, I) = I - \mathcal{\gamma}(B, I)
\end{eqnarray}
\begin{eqnarray} \label{eq:black_top_hat}
	\mathcal{T}^b(B, I) = \mathcal{\varphi}(B, I) - I
\end{eqnarray}

Another important morphological operation based on erosions and dilations is the \textit{geodesic reconstruction}.
% $\rho(\cdot)$. %\mathcal{R}(\cdot)
%This operation is obtained by iteratively performing a specific transformation over the image until idempotence~\footnote{There is no more change from the current iteration and the last one.} is reached~\citep{pesaresi2001new}.
There are two types of geodesic reconstruction: by erosion and by dilation.
%For simplicity, only the former one is formally detailed here, however, the latter one can be obtained, by duality, using the same reasoning.
The geodesic reconstruction by erosion $\rho^{\mathcal{E}} (\cdot)$, mathematically defined in Equation~\ref{eq:reconstruction_erosion}, receives two parameters as input: an image $I$ and a SE $B$.
The image $I$ (also referenced in this operation as mask image) is dilated by the SE $B$ ($\mathcal{\delta}(B, I)$) creating the marker image $Y$ ($Y \in I$), responsible for delimiting which objects will be reconstructed during the process.
A SE $B'$ (usually with any elementary composition~\cite{serra2012mathematical}) and the marker image $Y$ are provided for the reconstruction operation $\mathcal{R}^{\mathcal{E}}_{I}(\cdot)$.
This transformation, defined in Equation~\ref{eq:reconstruction}, reconstructs the marker image $Y$ (with respect to the mask image $I$) by recursively employing geodesic erosion (with the elementary SE $B'$) until idempotence is reached (i.e., $\mathcal{E}^{(n)}_{I}(\cdot) = \mathcal{E}^{(n+1)}_{I}(\cdot)$).
In this case, a geodesic erosion $\mathcal{E}^{(1)}_{I}(\cdot)$, defined in Equation~\ref{eq:geodesic_reconstruction}, consists of a pixel-wise maximum operation between an eroded (with elementary SE $B'$) marker image $Y$ and the mask image $I$.
By duality, a geodesic reconstruction by dilation can be defined, as presented in Equation~\ref{eq:reconstruction_dilation}.
These two crucial operations try to preserve all large (than the SE) objects of the image removing bright and dark small areas, such as noises.
Some examples of these operations can be seen in Figure~\ref{fig:ucmerced_mp}.

\begin{eqnarray} \label{eq:reconstruction_erosion}
%\rho^{\mathcal{E}} (B, I) = \wedge_n \mathcal{E}_{n}^{I} (Y) | \mathcal{E}_{n}^{I} = \mathcal{E}_{n+1}^{I}
\rho^{\mathcal{E}} (B, I) = \mathcal{R}^{\mathcal{E}}_{I}(B', Y) = \mathcal{R}^{\mathcal{E}}_{I}(B', \mathcal{\delta}(B, I))
\end{eqnarray}

\begin{eqnarray} \label{eq:reconstruction}
\begin{gathered}
\mathcal{R}^{\mathcal{E}}_{I}(B', Y) = \mathcal{E}^{(n)}_{I}(B', Y) = \\
= \mathrlap{\underbrace{\phantom{\mathcal{E}^{(1)}_{I}\bigg(B', \mathcal{E}^{(1)}_{I}\Big(B', \cdots \mathcal{E}^{(1)}_{I}\big(B', \mathcal{E}^{(1)}_{I}(B', Y)\bigr)\Bigr)\biggr)}}_{\text{\normalsize n times}}}
\mathcal{E}^{(1)}_{I}\bigg(B', \mathcal{E}^{(1)}_{I}\Big(B', \cdots \mathcal{E}^{(1)}_{I}\big(B', \mathcal{E}^{(1)}_{I}(B', Y)\bigr)\Bigr)\biggr)
\end{gathered}
\end{eqnarray}

\begin{eqnarray} \label{eq:geodesic_reconstruction}
\mathcal{E}^{(1)}_{I}(B', Y) = \max \{ \mathcal{E}(B', Y), I \}
\end{eqnarray}

\begin{eqnarray} \label{eq:reconstruction_dilation}
	%\rho^{\mathcal{\delta}} (B, I) = \vee_n \mathcal{\delta}_{n}^{I} (Y) | \mathcal{\delta}_{n}^{I} = \mathcal{\delta}_{n+1}^{I}
	\rho^{\mathcal{\delta}} (B, I) = \mathcal{R}^{\mathcal{\delta}}_{I}(B', Y) = \mathcal{R}^{\mathcal{\delta}}_{I}(B', \mathcal{E}(B, I))
\end{eqnarray}

%The reconstruction step is a process that allows a full retrieval of objects that are not completely suppressed by the erosion and it potentially needs several iterations before reaching stability.
Note that geodesic reconstruction operations require an iterative process until the convergence.
This procedure can be expensive, mainly when working with a large number of images. 
% (a common scenario when training neural networks~\cite{goodfellow2016deep}).
An approximation of such operations, presented in Equations~\ref{eq:simply_reconstruction_erosion} and~\ref{eq:simply_reconstruction_dilation}, can be achieved by performing just \textbf{one} transformation over the marker image with a large (than the SE used to create the marker image) structuring element.
In other words, suppose that $B$ is the SE used to create the marker image, then $B'$, the SE used in the reconstruction step, should be larger than $B$.
%, i.e., $B \subset B'$.
This process is faster since only one iteration is required, but may lead to worse results, given that the use of a large filter can make the reconstruction join objects that are close in the scene (a phenomenon known as \textit{leaking}~\cite{serra2012mathematical}).

\begin{eqnarray} \label{eq:simply_reconstruction_erosion}
	\tilde{\rho}^{\mathcal{E}} (B, I) = \mathcal{E}_{I}(B', \mathcal{\delta}(B, I))
\end{eqnarray}

\begin{eqnarray} \label{eq:simply_reconstruction_dilation}
	\tilde{\rho}^{\mathcal{\delta}} (B, I) = \mathcal{\delta}_{I}(B', \mathcal{E}(B, I))
\end{eqnarray}

%Although all previously defined morphological operations used a grayscale image $I$, they could have employed a binary image or even an image with several channels.
%In this case, morphological operations would be applied to each input channel independently and separately, generating an outcome with the same number of input channels.

\subsection{Related Work} \label{sec:relwork}

As introduced, such non-linear morphological operations~\cite{serra2012mathematical} have the ability to preserve some features that may be essential for some problems.
Supported by this, several tasks and applications have exploited the benefits of morphological operations, such as image analysis~\cite{yu2006medical,kimori2016quantifying,miri2011retinal,seo2018morphological,mondal2020image,franchi2020deep}, classification~\cite{masci2013learning,mellouli2017morph,borra2018classification}, segmentation~\cite{fauvel2007spectral,xia2015random,aptoula2016deep,kalshetti2017interactive,wang2018lidar}, and so on.
%mirzapour2015improving,gu2016nonlinear

Some of these techniques~\cite{yu2006medical,miri2011retinal,kimori2016quantifying,kalshetti2017interactive} are strongly based on mathematical morphology.
These approaches process the input images using only morphological operations.
The produced outcomes are then analyzed in order to extract high-level semantic information% related to the input images
, such as borders, area, geometry, volume, and more.
Other works~\cite{fauvel2007spectral,xia2015random,liu2017multimorphological,seo2018morphological,borra2018classification}
% mirzapour2015improving,gu2016nonlinear
go further and use morphology to extract robust features that are employed as input to machine learning techniques (such as Support Vector Machines, and decision trees) to perform image analysis, classification, and segmentation.
Usually, in these cases, the input images are processed using several different morphological transformations, each one employing a distinct structuring element, in order to improve the diversity of the extracted features.
All these features are then concatenated and used as input for the machine learning techniques.

More recently, ConvNets~\cite{goodfellow2016deep} started achieving outstanding results, mainly in applications related to images.
Therefore, it would be more than natural for researchers to propose works combining the benefits of ConvNets and morphological operations.
In fact, several works~\cite{masci2013learning,mellouli2017morph,aptoula2016deep,wang2018lidar} tried to combine these techniques to create a more robust model.
Some works~\cite{aptoula2016deep,wang2018lidar} employed morphological operations as a pre-processing step in order to extract the first set of discriminative features.
In these cases, pre-defined (hand-crafted) structuring elements are employed.
%the structuring elements of the morphological operations are not learned and pre-defined structuring elements are employed.
Those techniques use such features as input for a ConvNet responsible to perform the classification.
Based on the fact that morphology generates interesting features that are not captured by the convolutional networks, such works achieved outstanding results on pixelwise classification.

Other works~\cite{masci2013learning,mellouli2017morph,mondal2020image,franchi2020deep} introduced morphological operations into neural networks, creating a framework in which the structuring elements are optimized.
%Ritter and Sussner~\cite{ritter1996introduction} were the first to propose a network that performs and optimizes morphological operations.
%This work was developed before the ``deep learning boom'', therefore, is based upon simple multi-layer perceptions~\cite{goodfellow2016deep}.
%Proposed based on great mathematical cleverness, the network, composed of only a single layer, was used for binary classification, producing interesting results.
%Considering the new generation of deep networks, 
Masci \emph{et al.}~\cite{masci2013learning} proposed a convolutional network that aggregates pseudo-morphological operations.
%, such as pseudo-erosion, pseudo-dilation, pseudo-opening, pseudo-closing, and pseudo-top-hats.
Specifically, their proposed approach uses the counter-harmonic mean, which allows the convolutional layer to perform its traditional linear process, or approximations of morphological operations.
%counter-harmonic mean~\cite{bullen2013handbook}
They show that the approach produces outcomes very similar to real morphological operations.
Mellouli \emph{et al.}~\cite{mellouli2017morph} performed a more extensive validation of the previous method, proposing different deeper networks that are used to perform image classification.
In their experiments, the proposed network achieved promising results for two datasets of digit recognition.
\changes{
	In~\cite{mondal2020image}, the authors proposed a new network capable of performing some morphological operations (including erosion, dilation, opening, and closing) while optimizing non-flat structuring elements.
	Their proposed network, evaluated for image de-raining and de-hazing, produced results similar to those of a ConvNet but using much fewer parameters.
	Finally, Franchi \emph{et al.}~\cite{franchi2020deep} proposed a new deep learning framework capable of performing non-approximated mathematical morphology operations (including erosion, dilation) while optimizing non-flat structuring elements.
	Their method produced competitive results for edge detection and image denoising when compared to classical techniques and standard ConvNets.
}
%This proposed network and a new version that performs only approximate morphological operations were further evaluated in~\cite{mondal2019dense}.
%In this work, several networks were compared for the image classification task using five distinct datasets.
%The obtained results showed that the proposed networks (mainly the one performing approximate operations) produce results very similar to the baselines, which did not include any ConvNets.

%situation of this paper
In this work, we proposed a new network capable of performing and optimizing several morphological operations, including erosion, dilation, openings, closing, top-hats, and an approximation of geodesic reconstructions.
Several differences may be pointed out between the proposed approach and the aforementioned works:
(i) differently from~\cite{masci2013learning,mellouli2017morph}, the proposed technique really carries out morphological operations without any approximation (except for the reconstruction),
(ii) the morphological operations incorporated into the proposed network use flat SEs (which may be used with any input image) instead of the non-flat ones, as exploited by~\cite{mondal2020image,franchi2020deep} and that can only be used with grayscale input data, and
(iii) to the best of our knowledge, this is the first approach to implement (approximate) morphological geodesic reconstruction within deep-learning based models.

\section{Deep Morphological Networks} \label{sec:methodology}

In this section, we present the proposed approach, called Deep Morphological Networks (or simply DeepMorphNets), that replaces the linear convolutional filters with optimizable non-linear morphological operations.
% capable of doing morphological operations while optimizing the (flat) SEs.
%Technically, this new technique is strongly based on ConvNets mainly because of the similarity between the morphological operations and convolutional layers, since both employ a somehow similar processing operation.
%(morphological transformations perform operations similar to those performed in the convolutional layers), 
%Therefore, this new network seeks to efficiently combine morphological operations and deep learning, aggregating the ability to learn certain important types of image properties (such as borders and corners) of the former and the feature learning step of the latter.
%Such combination would bring advantages that could assist several applications in which borders and shape are considered essential.
%Such combination would aggregate the benefits of the morphological transformations (i.e., the learning of certain important types of image properties, such as borders and corners) and the advantages of the deep learning techniques (i.e., its crucial feature learning step).
%As introduced, the combination of non-linear morphological operations and deep learning would aggregate the benefits of the former and the advantages of deep learning-based methods, i.e., the feature learning process.
%However, there are several challenges in fully integrating morphological operations and deep learning, especially convolutional neural networks.
Such replacement allows the proposed network to capture distinct information when compared to previous existing models, an advantage that may assist different applications.
However, this process raises several challenging issues. % regarding the implementation and optimization.

One first challenge is due to the linear nature of the core operations of the existing networks.
The convolutional layers extract features from the input data using an optimizable filter by performing only linear operations not supporting non-linear ones.
%let us assume the input $y(\cdot)$ of a convolutional layer as a 3D 
Formally, let us assume a 3D input $y(\cdot)$ of a convolutional layer as a mapping from coordinates ($\mathds{Z}^3$) to the pixel-value domain ($\mathds{Z}$ or $\mathds{R}$).
Analogously, the trainable filter (or weight) $W(\cdot)$ of such layer can be seen as a mapping from 3D coordinates ($\mathds{Z}^3$) to the real-valued weights ($\mathds{R}$).
A standard convolutional layer performs a convolution 
%(denoted here as $\ast$) 
of the filter $W(\cdot)$ over the input $y(\cdot)$, according to Equation~\ref{eq:standard_conv}.
Note that the output of this operation is the summation of the linear combination between input and filter (across both space and depth).
Also, observe the difference between this operation and the morphological ones stated in Section~\ref{sec:background}.
This shows that replacing the convolutional filters with morphological operations is not straightforward.
%This shows that the integration between morphological operations and convolutional layers is not straightforward.
%This shows that the combination of morphological operations and convolutional layers is not straightforward. %, being very challenging 

\begin{eqnarray} \label{eq:standard_conv}
	\begin{gathered}
		S(W, y)_{(i, j)} = \sum_m \sum_n \sum_l W(m, n, l) y(i+m, j+n, l)
	\end{gathered}
\end{eqnarray}

Another important challenge is due to the optimization of non-linear operations by the network.
Precisely, in ConvNets, a loss function $\mathcal{L}$ is used to optimize the model.
%a loss function $\mathcal{L}$ is defined to allow the evaluation of the network's current state and its optimization towards a better state.
Nevertheless, the objective of any network is to minimize this loss function by adjusting the trainable parameters (or filters) $W$.
Such optimization is traditionally based on the derivatives of the loss function $\mathcal{L}$ with respect to the weights $W$.
For instance, suppose Stochastic Gradient Descent (SGD)~\cite{goodfellow2016deep} is used to optimize a ConvNet.
As presented in Equation~\ref{eq:sgd_op}, the optimization of the filters depends directly on the partial derivatives of the loss function $\mathcal{L}$ with respect to the weights $W$ (employed with a pre-defined learning rate $\alpha$).
Those partial derivatives are usually obtained using the backpropagation algorithm~\cite{goodfellow2016deep}, which is strongly supported by the fact that all operations of the network are easily differentiable,
% (with respect to the filters)
including the convolution presented in Equation~\ref{eq:standard_conv}.
However, this algorithm can not be directly applied to non-linear operations, such as the presented morphological ones, because those operations do not have easy derivatives.
% being, therefore, not allowing a simple optimization.

\begin{eqnarray} \label{eq:sgd_op}
	W = W - \alpha \frac{\partial \mathcal{L}}{\partial W}
\end{eqnarray}

Overcoming such challenges, we propose a network that employs depthwise and pointwise convolution with depthwise pooling to recreate and optimize morphological operations.
%, from basic to complex ones.
%We proposed a novel method that overcomes such challenges and allows the introduction and optimization of morphological operations inside ConvNets.
%First, Section~\ref{sec:basic_concept} introduce the basic concepts used as a foundation for the proposed DeepMorphNet.
%% Deep Morphological Network.
%Section~\ref{sec:neurons} presents the proposed morphological neurons.
%% responsible to perform morphological operations.
%The proposed morphological layer is presented in Section~\ref{sec:layers}.
%%, composed of the proposed neurons,
%The optimization of the filters (or SEs) of such layers is explained in Section~\ref{sec:optmization}.
%Finally, the proposed DeepMorphNet architectures are introduced in Section~\ref{sec:arch}.
%
%As introduced, the proposed technique is based on the idea of combining morphological operations and deep learning while keeping the optimization of the filters (also known as structuring elements).
The basic concepts of the proposed technique is presented in Section~\ref{sec:basic_concept}.
This concept is employed to create morphological neurons and layers, presented in Sections~\ref{sec:neurons} and~\ref{sec:layers}, respectively.
Section~\ref{sec:optmization} explains the optimization processed performed to learn the structure elements.
Finally, the morphological architecture exploited in this work are introduced in Section~\ref{sec:arch}.

%\subsection{Morphological Neurons} \label{sec:neurons}
\subsection{Basic Morphological Framework} \label{sec:basic_concept}

%Ideally, the combination of morphological operations and deep learning should be subject to an essential condition: the new technique should be capable of conserving the end-to-end learning strategy. %, i.e., it should integrate the current training procedure.
%The reason for this condition is that the method would be able to to extract the benefits of the feature learning step while allowing the combination of morphological operations with any other operation explored by deep learning-based approaches.
Towards the preservation of the end-to-end learning strategy, we propose a new framework, capable of performing morphological erosion and dilation, that uses operations already employed in other existing deep learning-based methods.
%, i.e., neurons based on this framework can be easily integrated into the standard training process.
%The processing of these morphological neurons can be separated into two steps.
The processing of this framework can be separated into two steps.
The \textbf{first} one employs depthwise convolution~\cite{chollet2017xception} to perform a delimitation of features, based on the neighborhood (or filter).
As defined in Equation~\ref{eq:depthwise_conv}, this type of convolution differs from standard ones since it handles the input depth independently, using the same filter $W$ to every input channel.
In other words, suppose that a layer performing depthwise convolution has $k$ filters and receives an input with $l$ channels, then the processed outcome would be an image of $k\times l$ channels, since each $k$-th filter would be applied to each $l$-th input channel.
The use of depthwise convolution simplifies the introduction of morphological operations into the deep network since the linear combination performed by this convolution does not consider the depth (as in standard convolutions presented in Equation~\ref{eq:standard_conv}).
This process is fundamental for the recreation of morphological operations since such transformations can only process one single channel at a time.
% (as aforementioned in Section~\ref{sec:background}).

\begin{eqnarray} \label{eq:depthwise_conv}
	S_l(W, y)_{(i, j)} = \sum_m \sum_n W(m, n) y(i+m, j+n, l)
\end{eqnarray}

%\begin{figure}[h]
%	\centering{
%		\def\svgwidth{\columnwidth}
%		\input{Figures/method/method_v2.pdf_tex}
%		\caption{Illustration of the concept employed in the morphological layers.}
%		\label{fig:layer}
%	}
%\end{figure}

However, just using this type of convolution does not allow the reproduction of morphological transformations, given that a spatial linear combination is still performed by this convolutional operation.
To overcome this, all filters $W$ are first converted into binary and then used in the depthwise convolution operation.
This binarization process, referenced hereafter as max-binarize, activates only the highest value of the filter.
%, i.e., only the highest value is considered active, while all others are deactivated.
Formally, the max-binarize $b(\cdot)$ is a function that receives as input the real-valued weights $W$ and processes them according to Equation~\ref{eq:w_bin}, where $\mathds{1} \{ condition \}$ is the indicator function.
%, that returns 1 if the $condition$ is true and 0 otherwise.
This process outputs a binary version of the weights, $W^b$, in which \textbf{only} the highest value in $W$ is activated in $W^b$.
By using $W^b$, the linear combination performed by depthwise convolution can be seen as a simple operation that preserves the exact value of the single pixel activated by this binary filter.
%by the only value activated in the binary filter.
%This allows us a different interpretation of the linear combination presented in Equation~\ref{eq:depthwise_conv}.

\begin{eqnarray} \label{eq:w_bin}
	W^b_{(i, j)} = b(W(i, j)) = \mathds{1} \{ max_{m,n}(W(m,n)) = W(i,j) \}
\end{eqnarray}

%This binarization process allows the convolution layer to retrieve a single pixel without taken into account the neighborhood.
But preserving only one pixel with respect to the binary filter is not enough to reproduce the morphological operations, since they usually operate over a neighborhood (defined by the SE $B$).
In order to reproduce this neighborhood concept in the depthwise convolution operation, we decompose each filter $W$ into several ones, that when superimposed retrieve the final SE $B$.
More specifically, suppose a filter $W$ with size $s\times s$.
% there is a total of $s^2$ possible variations of this filter
Since only one position can be activated at a time, this filter has a total of $s^2$ possible activation variations.
%(because of the aforementioned binarization process)
%, each one with a single different activated position.
This analysis is the same if we consider each element of a $s\times s$ SE independently.
%Suppose also a structuring element with size $s\times s$.
%Such SE defines the neighborhood that should be taken into account during the morphological operation
% As explained in Section~\ref{sec:background}, such SE defines the pixel neighborhood and can have any feasible configuration.
%Considering each position of this SE independently, each one can be considered as activated or deactivated.
%If we consider each position of this SE independently, it also has $s^2$ possible configurations.
%(when that position of the neighborhood should be taken into account) or deactivated (when the neighboring position should not be taken into account).
% Therefore, a SE of size $s\times s$ has $s^2$ possible configurations when considering each position separately.
Based on all this, a set of $s^2$ max-binary filters with size $s\times s$ is able to cover all possible configurations of a SE with the same size.
%, i.e., with this set, it is possible to recreate any feasible configuration of a $s\times s$ SE.
Thus, a set of $s^2$ filters with size $s\times s$ can be seen as a decomposed representation of the structuring element,
%neighborhood concept (or of the SE) 
given that those $s^2$ filters (with only a single activated position) can be superimposed in order to retrieve any possible $s\times s$ neighborhood configuration defined by the SE. 
%would be required to cover all possible configurations of this filter (i.e., to have each position activated at least once).
%Since only one value is kept activated by the max-binarize, to cover all possible configurations, it is required $s^2$ filters.
%Supported by this idea, any $s\times s$ SE can be decomposed into $s^2$ filters, each one with size $s\times s$ and only one activated value.
%Since only one value of each of these filter is activated, the $s^2$ filters cover all possible configurations of the $s\times s$ SE.
By doing this, the concept of neighborhood introduced by the SE can be exploited by the depthwise convolution.
Technically, a $s^2$ set of $s\times s$ filters $W$ can be converted into binary weights $W^b$
%(via the aforementioned max-binarize function $b(\cdot)$) 
and then, used to process the input data.
When exploited by Equation~\ref{eq:depthwise_conv}, each of these $s^2$ binary filter $W^b$ will preserve only one pixel which is directly related to one specific position of the neighborhood.
As may be observed, this first step recreates, in depth, the neighborhood of a pixel delimited by a $s\times s$ SE $B$, which is essentially represented by $s^2$ binary filters $W^b$ of size $s\times s$.
%any possible configuration $s^2$ filters with size $s\times s$

%\begin{eqnarray} \label{eq:se_recover}
%	B_{(i, j)} = \mathds{1} \{ \sum_l W^b (i,j,l) \geq 1 \}
%\end{eqnarray}

Since the SE $B$ was decomposed in depth, in order to retrieve it, a depthwise operation
%, presented in Equation~\ref{eq:se_recover}, 
must be performed over the $s^2$ binary filters $W^b$.
Analogously, a depthwise operation is also required to retrieve the final outcome, i.e., the eroded or dilated image.
This is the \textbf{second} step of this proposed framework, which is responsible to extract the relevant information 
%(depending on the transformation) 
based on the depthwise neighborhood.
In this step, an operation, called depthwise pooling 
%(given its similar to standard pooling) and denoted 
$P(\cdot)$, processes the $s^2$ outcomes (of the decomposed filters), producing the final morphological outcome.
%$P(\cdot)$, performs a pixelwise minimum or maximum over the $s^2$ outcomes (of the decomposed filters) depending on the operation, producing the final morphological outcome.
This pooling operation is able to actually output the morphological erosion and dilation by using pixel and depthwise minimum and maximum functions, as presented in Equations~\ref{eq:minpooling} and~\ref{eq:maxpooling}, respectively.
%Note that the reproduction of morphological operations using minimum and maximum functions is only possible because the set created with each pixel position along the channels can be considered an ordered set (similar to the definition presented in Section~\ref{sec:background}).
The outcome of this second step is the final (eroded or dilated) feature map that will be exploited by any subsequent process.

\begin{eqnarray} \label{eq:minpooling}
	P^{\mathcal{E}}(y)_{(i, j)} = \min_l y(i,j,l) % = \min_l S_l(W, y)_{(i, j)}
\end{eqnarray}

\begin{eqnarray} \label{eq:maxpooling}
	P^{\mathcal{\delta}}(y)_{(i, j)} = \max_l y(i,j,l) % = \max_l S_l(W, y)_{(i, j)}
\end{eqnarray}

Equations~\ref{eq:minpooling2} and~\ref{eq:maxpooling2} compile the two steps performed by the proposed framework for morphological erosion and dilation, respectively.
This operation, denoted here as $M(\cdot)$, performs a depthwise convolution (first step), which uses max-binary filters that decompose the representation of the neighborhood concept introduced by SEs, followed by a pixel and depthwise pooling operation (second step), outputting the final morphological (eroded or dilated) feature maps.
%performs a pixel and depthwise pooling operation over the outcome of a depthwise convolution layer, composed of filters decomposing the neighborhood (or the SE $B$), outputting the final morphological (eroded or dilated) feature maps.
Note the similarity between these functions and Equations~\ref{eq:erosion_orderedset} and~\ref{eq:dilation_orderedset} presented in Section~\ref{sec:background}.
The main difference between these equations is in the neighborhood definition.
While in the standard morphology, the neighborhood of a pixel is defined spatially (via SE $B$), in the proposed framework, the neighborhood is defined along the channels due to the decomposition of the SE $B$ into several filters and, therefore, minimum and maximum operations also operate over the channels.

\begin{eqnarray} \label{eq:minpooling2}
	M^{\mathcal{E}}(W, y)_{(i, j)} = P^{\mathcal{E}}(S_l(W, y))_{(i, j)} = \min_l S_l(W, y)_{(i, j)}
\end{eqnarray}

\begin{eqnarray} \label{eq:maxpooling2}
	M^{\mathcal{\delta}}(W, y)_{(i, j)} = P^{\mathcal{\delta}}(S_l(W, y))_{(i, j)} = \max_l S_l(W, y)_{(i, j)}
\end{eqnarray}

\begin{figure}[h]
	\centering
	\includegraphics[width=0.95\columnwidth]{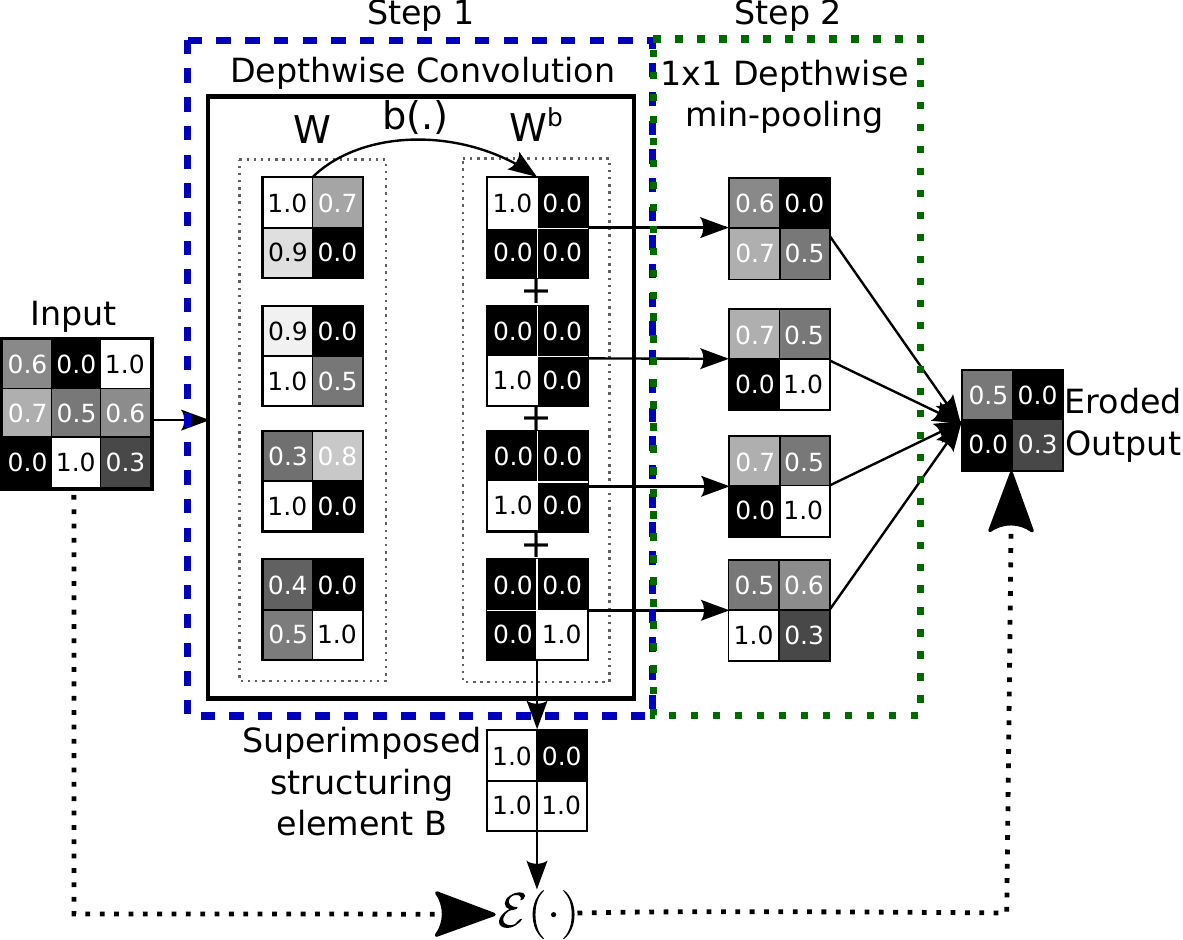}
	\caption{Example of a morphological erosion based on the proposed framework.
		The 4 filters $W$ (with size $4\times4$) actually represent a unique $4\times4$ SE.
		Each filter $W$ is first converted to binary $W^b$, and then used to process each input channel (step 1, blue dashed rectangle).
		The output is then processed via a pixel and depthwise min-pooling to produce the final eroded output (step 2, green dotted rectangle).
		Note that the binary filters $W^b$, when superimposed, retrieve the final SE $B$.
		The dotted line shows that the processing of the input with the superimposed SE $B$ using the standard morphological erosion results in the same eroded output image produced by the proposed morphological erosion.}
	\label{fig:layer}
\end{figure}

A visual example of the proposed framework being used for morphological erosion is presented in Figure~\ref{fig:layer}.
%The first step of this layer (blue square) is the processing of the input with the binary filters $W^b$, according to Equation~\ref{eq:depthwise_conv}, outputting a depthwise neighborhood of the pixels.
In this example, the depthwise convolution has 4 filters $W$ with size $4\times4$ which actually represent a unique $4\times4$ SE.
The filters $W$ are first converted into binary using the max-binarize function $b(\cdot)$, presented in Equation~\ref{eq:w_bin}.
Then, each binary filter $W^b$ is used to process (step 1, blue dashed rectangle) each input channel (which, for simplicity, is only one in this example) using Equation~\ref{eq:depthwise_conv}.
In this process, each binary filter $W^b$ outputs an image in which each pixel has a direct connection to the one position activated in that filter.
% (that, actually, represents a neighborhood position activated in the SE $B$).
%(i.e., a connection to the one position activated by that filter in the neighborhood defined by the superimposed SE $B$).
The output is then processed (step 2, green dotted rectangle) via a pixel and depthwise min-pooling $P(\cdot)^{\mathcal{E}}$ (according to Equation~\ref{eq:minpooling}) to produce the final eroded output.
Note that the binary filters $W^b$, when superimposed 
%(using Equation~\ref{eq:se_recover}), 
retrieve the final SE $B$.
The dotted line shows that the processing of the input with the superimposed SE $B$ using the standard erosion ($\mathcal{E}(\cdot)$ presented in Equation~\ref{eq:erosion_orderedset}) results in the same eroded output image produced by the proposed morphological erosion.

%\subsubsection{Complex Neurons} \label{sec:complex_neurons}
\subsection{Morphological Processing Units} \label{sec:neurons}

The presented framework is the foundation of all proposed morphological processing units (or neurons).
Before presenting them, it is important to observe that, although the proposed framework is able to reproduce morphological erosion and dilation, it has an important drawback: since it employs depthwise convolution, the number of outcomes can grow potentially, given that, each input channel is processed independently by each processing unit.
%As introduced, the proposed framework, based on depthwise convolution, processes each input channel separately, outputting more outcomes than the standard convolution.
%To avoid exponential growth on the number of outcomes, we propose to use a pointwise convolution~\cite{chollet2017xception}.
In order to overcome this issue and make the proposed technique more scalable, we propose to use a pointwise convolution~\cite{chollet2017xception} to force each processing unit to output only one image (or feature map).
% aside from the aforementioned steps (of the proposed framework),
Particularly, all neurons proposed in this work have the same design with two parts: 
(i) the core operation (fundamentally based on the proposed framework), in which the processing unit performs its morphological transformation outputting multiple outcomes, and
(ii) the pointwise convolution~\cite{chollet2017xception}, which performs a pixel and depthwise (linear) combination of the outputs producing only one outcome. % (per neuron). 
Observe that, even though the pointwise convolution performs a depthwise combination of the multiple outcomes, it does not learn any spatial feature, since it employs a pixelwise (or pointwise) operation, managing each pixel separately.
This design allows the morphological neuron to have the exact same input and output of a standard existing processing unit, i.e., it receives as input an image with any number of bands and outputs a single new representation.
It is interesting to notice that this processing unit design employs depthwise and pointwise convolution~\cite{chollet2017xception}, resembling very much the depthwise separable convolutions~\cite{chollet2017xception}, but with extra steps and binary decomposed filters.
Next Sections explain the \textbf{core operation} of all proposed morphological processing units.
%Note that these neurons were conceived to be equivalent in terms of operations.
%Therefore, all of them have, exactly, two operations (based on the proposed framework).
%Also 
Note that, although not mentioned in the next Sections, the pointwise convolution is present in all processing units as aforementioned. % proposed in this work.

\subsubsection{Composed Processing Units} \label{sec:composed_layers}

%The newly introduced framework allows a deep network to perform erosion and dilation, the two basic operations of morphology. %, over the input data.
%However, instead of using such operations independently, the first proposed processing unit is based on both morphological transformations.
%at the same time.
The first morphological neurons, called composed processing units, have, in their core, a morphological erosion followed by a dilation (or vice-versa), without any constraint on the weights.
% chanussot - mauro
The motivation behind the composed processing unit is based on the potential of the learned representation.
While erosion and dilation can learn simple representations, the combination of these operations is able to capture more complex information.
Equations~\ref{eq:composed_dilation_erosion} and~\ref{eq:composed_erosion_dilation} present the two possible configurations of the morphological composed neurons.
It is important to notice that the weights ($W_1$ and $W_2$) of each operation of this neuron are independent.
%Aside from this, a visual representation of one type of composed neuron can be seen in Figure~\ref{fig:composed}.

\begin{eqnarray} \label{eq:composed_dilation_erosion}
	M^{C_\mathcal{\delta}}(W, y) = M^{\mathcal{\delta}}(W_2, M^{\mathcal{E}}(W_1, y))
\end{eqnarray}
\begin{eqnarray} \label{eq:composed_erosion_dilation}
	M^{C_\mathcal{E}}(W, y) = M^{\mathcal{E}}(W_2, M^{\mathcal{\delta}}(W_1, y))
\end{eqnarray}

\subsubsection{Opening and Closing Processing Units} \label{sec:opening_closing_layers}

%Aside from implementing morphological erosion and dilation, the proposed framework is also able to support the implementation of other, more complex, morphological operations (or their approximations).
The proposed framework is also able to support the implementation of more complex morphological operations.
The most intuitive and simple transformations to be implemented are the opening and closing (presented in Section~\ref{sec:background}).
%As stated in Section~\ref{sec:background}, an opening is simply an erosion operation followed by a dilation of the eroded output (Equation~\ref{eq:opening}), while closing is the reverse operation (Equation~\ref{eq:closing}).
%In both cases, the two basic operations (erosion and dilation or vice-versa) use the same SE $B$.
In fact, the implementation of the opening and closing processing units, using the proposed framework, is straightforward.
The core of such neurons is very similar to that of the composed processing units, except that in this case a tie on the filters of the two basic morphological operations is required in order to make them use the same weights, i.e., the same SE $B$.
%A visual representation of the proposed opening neuron, presented in Figure~\ref{fig:open_layer}, allows a better view of the operation.
% of an opening or closing neuron using the proposed framework 
Equations~\ref{eq:morph_opening} and~\ref{eq:morph_closing} define the opening and closing morphological neurons, respectively.
Note the similarity between these functions and Equations~\ref{eq:opening} and~\ref{eq:closing}.

\begin{eqnarray} \label{eq:morph_opening}
	M^{\mathcal{\gamma}}(W, y) = M^{\mathcal{\delta}}(W, M^{\mathcal{E}}(W, y))
\end{eqnarray}
\begin{eqnarray} \label{eq:morph_closing}
	M^{\mathcal{\varphi}}(W, y) = M^{\mathcal{E}}(W, M^{\mathcal{\delta}}(W, y))
\end{eqnarray}

\subsubsection{Top-hat Processing Units} \label{sec:tophat_layers}

The implementation of other, more complex, morphological operations is a little more tricky.
%This is because complex operations, such as top-hats and geodesic reconstructions, require both the input data and the processed image to generate the final outcome.
This is the case of the top-hat operations, which require both the input and processed data to generate the final outcome.
Therefore, for such operations, a skip connection~\cite{goodfellow2016deep} (based on the identity mapping) is employed to support the forwarding of the input data, allowing it to be further processed.
%in order to reproduce this arrangement.
The core of the top-hat processing units is composed of three parts: 
(i) an opening or closing morphological processing unit, depending on the type of the top-hat, 
(ii) a skip connection, that allows the forwarding of the input data, and 
(iii) a subtraction function that operates over the data of both previous parts, generating the final outcome.
%Then, the processed and the input data (which is forwarded through the aforementioned skip connection) are then subtracted generating the final top-hat outcome.
%A visual concept of the white top-hat neuron is presented in Figure~\ref{fig:top_hat_layer}.
Such operation and its counterpart (the black top-hat) are defined in Equations~\ref{eq:morph_white_top_hat} and~\ref{eq:morph_black_top_hat}, respectively.
%Note that the order of this subtraction operation can vary depending on the top-hat transformation, as presented in Section~\ref{sec:background}.

\begin{eqnarray} \label{eq:morph_white_top_hat}
	M^{\mathcal{T}^w}(W, y) = y - M^{\mathcal{\gamma}}(W, y)
\end{eqnarray}
\begin{eqnarray} \label{eq:morph_black_top_hat}
	M^{\mathcal{T}^b}(W, y) = M^{\mathcal{\varphi}}(W, y) - y
\end{eqnarray}

\subsubsection{Geodesic Reconstruction Processing Units} \label{sec:reconstruction_layers}

Similarly to the previous processing units, the geodesic reconstruction also requires the input and processed data in order to produce the final outcome. 
Hence, the implementation of this important operation is also based on skip connections.
Aside from this, as presented in Section~\ref{sec:background}, reconstruction operations require an iterative process.
Although this procedure is capable of producing better outcomes, its introduction in a deep network is not straightforward, given that each process can take a different number of iterations.
Supported by this, the reconstruction processing units proposed in this work are an approximation, in which just one transformation over the marker image is performed. % (and not several iterations).
Based on this, the input is processed by two basic morphological operations (without any iteration) and an elementwise max- or min-operation (depending on the reconstruction) is performed over the input and processed images.
Such concept is presented in Equations~\ref{eq:morph_simply_reconstruction_erosion} and~\ref{eq:morph_simply_reconstruction_dilation} for reconstruction by erosion and dilation, respectively.
%A visual representation of the processing unit for reconstruction by erosion is presented in Figure~\ref{fig:rec_layer}.
Note that the SE used in the reconstruction of the marker image (denoted in Section~\ref{sec:background} by $B'$) is a dilated version of the SE employed to create such image.

\begin{eqnarray} \label{eq:morph_simply_reconstruction_erosion}
	M^{\tilde{\rho}^{\mathcal{E}}} (W, y) = M^{\mathcal{E}}_{y}(W, M^{\mathcal{\delta}}(W, y))
\end{eqnarray}
\begin{eqnarray} \label{eq:morph_simply_reconstruction_dilation}
	M^{\tilde{\rho}^{\mathcal{\delta}}} (W, y) = M^{\mathcal{\delta}}_{y}(W, M^{\mathcal{E}}(W, y))
\end{eqnarray}

\subsection{Morphological Layer} \label{sec:layers}

%Since the latter always has two operations (erosion followed by dilation, or vice-versa), in order to force a pattern in the MorphNet layer, we always forcing all MorphNet layers to have the same number of process in this part.
%In other words, if a MorphNet has a opening, internally it has two layers (one erosion and one dilation).
%If one want to use basic process (such as erosion and dilation), in order to make fair, 

%After definition the processing units, we are able to define the , which are composed of the presented neurons.
After defining the processing units, we are able to specify the morphological layers, which provide the essential tools for the creation of the DeepMorphNets.
Similar to the standard convolutional layer, this one is composed of several processing units.
However, the proposed morphological layer has two main differences when conceptually compared to the convolutional one.
%(i) the absence of activation functions, which causes the pixel to be not explicitly mapped to a non-linear space, and
%(ii) the possibility of use multiple types of processing units in the same layer, which allows it to aggregate more (possibly complementary) information.
The first one is related to the neurons that compose the layers.
Particularly, in convolutional layers, the neurons are able to perform the convolution operation. 
%Thus, those layers are usually composed of several neurons performing the same operation (with different filters) over the input data.
Though the filter of each neuron can be different, the operation performed by each processing unit in a convolutional layer is a simple convolution.
On the other hand, there are several types of morphological processing units, from opening and closing to geodesic reconstruction.
Therefore, a single morphological layer can be composed of several neurons that may be performing different operations.
This process allows the layer to produce distinct (and possibly complementary) outputs, increasing the heterogeneity of the network and, consequently, the generalization capacity.
The second difference is the absence of activation functions.
More specifically, in modern architectures, convolutional layers are usually composed of a convolution operation followed by an activation function (such as ReLU~\cite{nair2010rectified}), that explicitly maps the data into a non-linear space.
In morphological layers, there are only processing units and no activation function is employed.
%One can somehow see the type of the neuron (opening, closing, top-hat, etc) as its activation, however, it does not behave similarly, since no explicit mapping is performed through a function.

%Figure~\ref{fig:morph_layer} presents the concept of a single morphological layer.
%Observe that each neuron is performing a specific operation and outputting only one feature map.
%Also, note that, although the input has $c$ channels, supported by the pointwise convolution, each neuron outputs only one feature map.
%Hence, the number of outputted maps is directly connected to the number of neurons in that layer.
%In Figure~\ref{fig:morph_layer}, the layer has $n$ neurons that, consequently, produce $n$ feature maps.

%\begin{figure}[h]
%	\centering
%	\includegraphics[width=0.4\textwidth]{morph_layers.png}
%	\caption{Concept of a morphological layer.
%		Note that a single morphological layer can have neurons performing different operations.
%		This process is able to aggregate heterogeneous and complementary information.}
%	\label{fig:morph_layer}
%\end{figure}

%\begin{figure}[h]
%	\centering
%	\includegraphics[width=0.5\textwidth]{network.pdf}
%	\caption{Concept of a DeepMorphNet layer.}
%	\label{fig:network}
%\end{figure}

\subsection{Optimization} \label{sec:optmization}

Aside from defining the morphological layer, as introduced, we want to optimize its parameters, i.e., the filters $W$.
\textit{Since the proposed morphological layer uses common (differentiable) operations already employed in other existing deep learning-based methods, the optimization of the filters is straightforward.}
In fact, the same traditional existing techniques employed in the training of any deep learning-based approach, such feedforward, backpropagation and SGD~\cite{goodfellow2016deep}, can also be used for optimizing a network composed of morphological layers.

The training procedure is detailed in Algorithm~\ref{algo:training}. 
Given the training data $(y_0, y^{*})$, the \textbf{first step} is the feedforward, comprised in the loop from line 2 to 8.
In the first part of line 4, the weights of the first depthwise convolution are converted into binary (according to Equation~\ref{eq:w_bin}).
Then, in the second part, the first depthwise convolution (denoted here as $\ast$) is performed, with the first depthwise pooling being executed in the third part of this line.
The same operations are repeated in line 6 for the second depthwise convolution and pooling.
Finally, in line 7, the pointwise convolution is carried out.
After the forward propagation, the total error of the network can be estimated.
With this error, the gradients of the last layer can be directly estimated (line 10).
These gradients can be used by the backpropagation algorithm to calculate the gradients of the inner layers.
In fact, this is the process performed in the \textbf{second training step}, comprised in the loop from line 11 to 15.
It is important to highlight that during the backpropagation process, the gradients are calculated normally, using real-valued numbers (and not binary).
Precisely, line 12 is responsible for the optimization of the pointwise convolution.
The first part (of line 16) propagates the error of a specific pointwise convolution to the previous operation, while the second part calculates the error of that specific pointwise convolution operation.
The same process is repeated for the second and then for the first depthwise convolutions (lines 13 and 14, respectively).
Note that during the backpropagation, the depthwise pooling is not optimized since this operation has no parameters and only passes the gradients to the previous layer.
% (similar to the backpropagation employed in the max-pooling layers commonly explored in ConvNets).
The \textbf{last step} of the training process is the update of the weights and optimization of the network.
This process is comprised in the loop from line 17 to 21.
%Observe that, for simplicity, Algorithm~\ref{algo:training} uses SGD to optimize the network, however, any other optimization algorithm could be exploited.
For a specific layer, line 18 updates the weights of the pointwise convolution while lines 19 and 20 update the parameters of the first and second depthwise convolutions, respectively.
\begin{algorithm}\captionsetup{labelfont={sc,bf}, labelsep=newline}
	\caption{Training a Deep Morphological Network with L layers.}
	\label{algo:training}
	\begin{algorithmic}[1]
		\REQUIRE{a minibatch of inputs and targets $(y_0, y^{*})$, previous weights $W$, and previous learning rate $\alpha$.}
		\ENSURE{updated weights $W$.}
		\STATE{{1. Forward propagation:}}
			\FOR{k=1 to L}
				% first neuron
				\STATE{\{First Processing Unit Operation\}}
				\STATE{$W_k^{b^{(1)}} \leftarrow b(W_k^{(1)})$} %,$s_k^{(1)} \leftarrow y_{k-1} \ast W_k^{b^{(1)}}$,$y_k^{(1)} \leftarrow P(s_k^{(1)})$} %\hspace{1.37cm} \COMMENT{Binarization}
				\STATE{$s_k^{(1)} \leftarrow y_{k-1} \ast W_k^{b^{(1)}}$} %\hspace{1cm} \COMMENT{Depthwise Convolution}
				\STATE{$y_k^{(1)} \leftarrow P(s_k^{(1)})$} %\hspace{1.77cm} \COMMENT{Depthwise Pooling}
				%\FOR{i=1 to I} 
				%\FOR{j=1 to O}
				%\STATE{$W_{k,i,j}^b \leftarrow Max\_Binarize(W_{k,i,j})$}
				%\STATE{$s_{k,i,j} \leftarrow a_{k-1,i,j} \ast W_{k,i,j}^b$} \COMMENT{Depthwise Convolution}
				%\ENDFOR
				%\ENDFOR

				% second neuron
				\STATE{\{Second Processing Unit Operation\}}
				\STATE{$W_k^{b^{(2)}} \leftarrow b(W_k^{(2)})$}  %, $s_k^{(2)} \leftarrow y_k^{(1)} \ast W_k^{b^{(2)}}$, $y_k^{(2)} \leftarrow P(s_k^{(2)})$} %\hspace{1.37cm} \COMMENT{Binarization}
				\STATE{$s_k^{(2)} \leftarrow y_k^{(1)} \ast W_k^{b^{(2)}}$} %\hspace{1.17cm} \COMMENT{Depthwise Convolution}
				\STATE{$y_k^{(2)} \leftarrow P(s_k^{(2)})$} %\hspace{1.77cm} \COMMENT{Depthwise Pooling}
				\STATE{$y_k \leftarrow y_k^{(2)} \ast W_k^{(1\times1)}$} \hspace{1.07cm} \COMMENT{Pointwise Convolution}
		\ENDFOR
		%\STATE{$a_{k,i,j} \leftarrow max(s_{k,i,j})$} \COMMENT{Depthwise Max-pooling}
		\STATE{{2. Backpropagation:}} \COMMENT{Gradients are not binary.}
			\STATE{Compute $g_{y_L^{(1)}} = \frac{\partial \mathcal{L}}{\partial y_L}$ knowing $y_L$ and $y^{*}$}
			\FOR{k=L to 1}
				\STATE{$g_{y_{k-1}} \leftarrow g_{y_k^{(1)}} W_{k-1}^{(1\times1)}$, $g_{W_{k-1}^{(1\times1)}} \leftarrow g^{\intercal}_{y_k^{(1)}} y_{k-1}$}
				%\STATE{$g_{W_{k-1}^{(1\times1)}} \leftarrow g^{\intercal}_{y_k^{(1)}} y_{k-1}$}

				\STATE{$g_{y_{k-1}^{(2)}} \leftarrow g_{y_{k-1}} W_{k-1}^{b^{(2)}}$, $g_{W_{k-1}^{b^{(2)}}} \leftarrow g^{\intercal}_{y_{k-1}} y_{k-1}^{(2)}$}
				%\STATE{$g_{W_{k-1}^{b^{(2)}}} \leftarrow g^{\intercal}_{y_{k-1}} y_{k-1}^{(2)}$}
				
				\STATE{$g_{y_{k-1}^{(1)}} \leftarrow g_{y_{k-1}^{(2)}} W_{k-1}^{b^{(1)}}$, $g_{W_{k-1}^{b^{(1)}}} \leftarrow g^{\intercal}_{y_{k-1}^{(2)}} y_{k-1}^{(1)}$}
				% \STATE{$g_{W_{k-1}^{b^{(1)}}} \leftarrow g^{\intercal}_{y_{k-1}^{(2)}} y_{k-1}^{(1)}$}
			\ENDFOR
		\STATE{{3. Update the weights:}}
			\FOR{k=1 to L}
%				\IF{k == L}
%					\STATE{$W_k \leftarrow W_k - \alpha g_{W_k}$}
%				\ELSE
				\STATE{$W_k^{(1\times1)} \leftarrow W_k^{(1\times1)} - \alpha g_{W_k^{(1\times1)}}$}
				\STATE{$W_k^{(1)} \leftarrow W_k^{(1)} - \alpha g_{W_k^{b^{(1)}}}$}
				\STATE{$W_k^{(2)} \leftarrow W_k^{(2)} - \alpha g_{W_k^{b^{(2)}}}$}
		\ENDFOR
	\end{algorithmic}
\end{algorithm}

\subsection{DeepMorphNet Architectures} \label{sec:arch}

%With all the fundamentals defined, we can finally specify the DeepMorphNet architectures exploited in this work.
%Particularly, , presented in Figure~\ref{fig:cnn_arch},
Two networks, composed essentially of morphological and fully connected layers, were proposed for the image and pixel classification tasks.
Although such architectures have distinct designs, the pointwise convolutions exploited in the morphological layers have always the same configuration: kernel $1\times 1$, stride 1, and no padding.
Furthermore, all networks receive input images with $224 \times 224$ pixels, use cross-entropy as loss function, and SGD as optimization algorithm~\cite{goodfellow2016deep}.
For the pixel classification task, the proposed networks were exploited based on the pixelwise paradigm defined by~\cite{nogueira2015improving}, in which each pixel is classified independently through a context window.

The first network is the simplest one, having just a unique layer composed of one morphological opening $M^{\mathcal{\gamma}}$.
% (with kernel size of $11\times 11$, stride 1 and padding 5 for both depthwise convolutions).
This architecture was designed to be used with the proposed synthetic datasets (presented in Section~\ref{subsec:syn_datasets}).
Because of this, it is referenced hereafter as \textbf{DeepMorphSynNet}.
Note that this network was only conceived to validate the learning process of the proposed framework as explained in Section~\ref{subsec:synthetic_res}.

\begin{figure*}[t]
	\centering
	%\subfloat[Simple DeepMorphSynNet proposed for evaluating the learning process]{
	%	\includegraphics[width=0.3\textwidth]{network_synthetic_v2.pdf}
	%	\label{DeepMorphNet1}
	%}
	%\hspace{1mm}
	%\subfloat[DeepMorphLeNet version of the LeNet archicture~\cite{lecun1998gradient}]{
	%	\includegraphics[width=0.69\textwidth]{network_lenet_v2.pdf}
	%	\label{DeepMorphNet2}
	%}
	%\hspace{1mm}
	%\subfloat[]{
		\includegraphics[width=\textwidth]{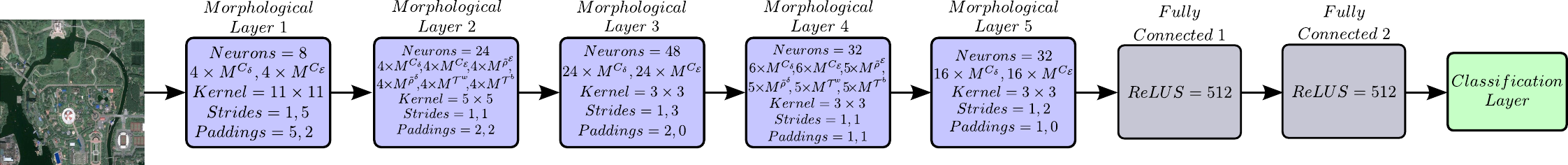}
	%	\label{DeepMorphNet3}
	%}
	\caption{The proposed morphological network DeepMorphNet conceived inspired by the AlexNet~\cite{krizhevsky2012imagenet}.
		% The three DeepMorphNet architectures proposed and exploited in this work.
		% Note that the number of morphological neurons of each type in each layer is codified using the Equations of Section~\ref{sec:neurons}.
		Since each layer is composed of distinct types of morphological neurons, the number of each type of neuron in each layer is presented as an integer times the symbol that represents that neuron (as presented in Section~\ref{sec:neurons}).
		Hence, the total number of processing units in a layer is the sum of all neurons independently of the type.
		Also, observe that the two depthwise convolutions of a same layer share the kernel size, differing only in the stride and padding.
		These both parameters are presented as follows: the value related to the first depthwise convolution is reported separated by a comma of the value related to the second depthwise convolution.
		Although not visually represented, the pointwise convolutions explored in the morphological layers always use the same configuration: kernel $1\times 1$, stride 1, and no padding.
}
	\label{fig:cnn_arch}
\end{figure*}

To analyze the effectiveness of the technique in a more complex scenario, we proposed a larger network inspired by the famous AlexNet~\cite{krizhevsky2012imagenet} architecture.
It is important to highlight that AlexNet~\cite{krizhevsky2012imagenet} was our inspiration, and not a more complex architecture, because if its simplicity, which allows a clear analysis of the benefits of the proposed technique, thus avoiding confusing them with other advantages of more complex deep learning approaches.
This proposed morphological version of the AlexNet~\cite{krizhevsky2012imagenet}, called \textbf{DeepMorphNet} and presented in Figure~\ref{fig:cnn_arch}, has the same number of layers of the original architecture but fewer neurons in each layer.
Specifically, this network has 5 morphological and 3 fully connected layers, responsible to learn high-level features and perform the final classification.
To further evaluated the potential of the proposed technique, in some experiments, a new version of the DeepMorphNet, using a modern component called Selective Kernels (SK)~\cite{li2019selective}, was developed and experimented.
This new network, referenced hereafter as \textbf{DeepMorphNet-SK}, uses such components to weigh the features maps, giving more attention to some maps than the others.

\section{Experimental Setup} \label{sec:experiments}

In this section, we present the experimental setup.
Section~\ref{subsec:datasets} presents the datasets.
Baselines are described in Section~\ref{subsec:baselines} while the protocol is introduced in Section~\ref{subsec:protocol}.

\subsection{Datasets} \label{subsec:datasets}

\textbf{Six} datasets were employed to validate the proposed DeepMorphNets.
Two image classification synthetic ones were exclusively designed to check the feature learning of the proposed technique.
%In order to first verify the DeepMorphNets, two synthetic datasets were designed.
%The first two are synthetic datasets designed exclusively to validate the feature learning process of the proposed technique.
Other two image classification datasets were selected to further verify the potential of DeepMorphNets.
Finally, to assess the performance of the proposed technique in distinct scenarios, two pixel classification datasets were exploited.
%: UCMerced Land-use~\cite{mercedlanduse} and WHU-RS19~\cite{xia2010structural}.
%It is important to highlight that the images of these last two datasets are resized in order to fit the requirements of the proposed architectures.

%In order to perform the first validations of the DeepMorphNet, we proposed two synthetic datasets, in which we could know, a priori, the structuring element that should be learned by the network as well as the desired outcome. Then, to really verify the potential of DeepMorphNets, we selected two image classification datasets: UCMerced Land-use Dataset~\cite{mercedlanduse} and WHU-RS19 Datasets~\cite{xia2010structural}.

%The motivation to use datasets of this domain comes from the fact that morphological operations are still considered state-of-the-art in this area.

%Two datasets have been employed to evaluate the proposed DeepMorphNets:
%(i) UCMerced Land-use Dataset~\cite{mercedlanduse}, which is composed of aerial high resolution scenes in the visible spectrum divided into 21 classes;
%(ii) Aerial Image Dataset (AID)~\cite{xia2017aid}, which is a large aerial dataset divided into 30 classes.

\subsubsection{Image Classification Datasets} \label{subsec:syn_datasets}

%In order to allow such validation, these synthetic datasets were created the optimal structuring element is known a priori

\noindent\textbf{Synthetic Datasets.} As introduced, two simple synthetic (image classification) datasets were designed in this work to validate the feature learning process of the proposed DeepMorphNets.
In order to allow such validation, these datasets were created so that it is possible to define, a priori, the optimal SE (i.e., the SE that would produce the best results) for a classification scenario.
Hence, in this case, the validation would be performed by comparing the learned SE with the optimal one, i.e., if both SEs are similar, then the proposed technique is able to perform well the feature learning step.

Specifically, both datasets are composed of \textbf{1,000 grayscale images} with a resolution of $224\times 224$ pixels (a common image size employed in famous architecture such as AlexNet~\cite{krizhevsky2012imagenet}) \textbf{equally} divided into two classes.
% In both datasets, the background is represented by 0 while the object in the scene is 255.

The \textbf{first dataset} has two classes.
The first one is composed of images with small ($5\times 5$ pixels) squares whereas the second consists of images with large ($9\times 9$ pixels) squares.
Each image of this dataset has only one square (of one of the above classes) positioned randomly.
In this case, an opening with a SE larger than $5\times 5$ but smaller than $9\times 9$ should erode the small squares while keeping the others, allowing the model to perfectly classify the dataset.

More difficult, the \textbf{second synthetic dataset} has two classes of rectangles.
The first class has shapes of $7\times 3$ pixels while the other one is composed of rectangles of $3\times 7$.
%The position of the rectangles in each image is defined randomly.
As in the previous dataset, each image of this dataset has only one rectangle (of one of the above classes) positioned randomly.
This case is a little more complicated because the network should learn a SE based on the orientation of one the rectangles.
Particularly, it is possible to perfectly classify this dataset using a single opening operation with one of the following types of SEs:
(i) a rectangle of at least 7 pixels of width and height larger than 3 but smaller than 7 pixels, which would erode the first class of rectangles and preserve the second one, or
(ii) a rectangle with a width larger than 3 but smaller than 7 pixels and height larger than 7 pixels, which would erode the second class of rectangle while keeping the first one.

%in the same direction of one of the rectangles to be used in an opening operation 

\noindent\textbf{UCMerced Land-use Dataset.} This publicly available dataset~\cite{mercedlanduse} is composed of 2,100 aerial images, each one with $256\times256$ pixels and 0.3-meter resolution per pixel.
These images, obtained from different US locations, were classified into 21 classes: agricultural, airplane, baseball diamond, beach, buildings, chaparral, dense residential, forest, freeway, golf course, harbor, intersection, medium density residential, mobile home park, overpass, parking lot, river, runway, sparse residential, storage tanks, and tennis courts.
As can be noticed, this dataset has highly overlapping classes such as the dense, medium, and sparse residential classes which mainly differs in the density of structures.
Samples of these and other classes are shown in Figure~\ref{fig:merced_dataset}.

\newcommand{\exFigSize}{0.07}

\begin{figure}[h!]
	\centering
	\scriptsize
%	\subfloat[Agricultural]{
%		\includegraphics[width=\exFigSize\textwidth, keepaspectratio=true]{ucmerced/c_agricultural04.jpg}
%		\includegraphics[width=\exFigSize\textwidth, keepaspectratio=true]{ucmerced/c_agricultural07.jpg}
%	}
%	\hspace{1mm}
	\subfloat[Airplane]{
		\includegraphics[width=\exFigSize\textwidth, keepaspectratio=true]{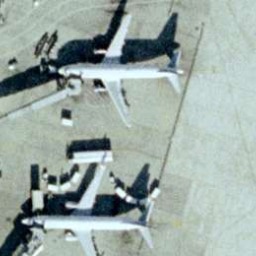}
		\includegraphics[width=\exFigSize\textwidth, keepaspectratio=true]{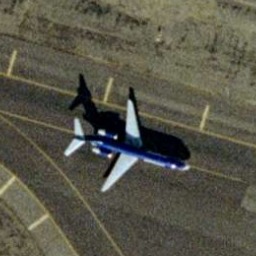}
	}
	\hspace{1mm}
	\subfloat[Baseball Diamond]{
		\includegraphics[width=\exFigSize\textwidth, keepaspectratio=true]{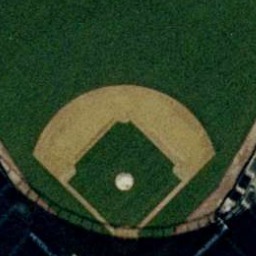}
		\includegraphics[width=\exFigSize\textwidth, keepaspectratio=true]{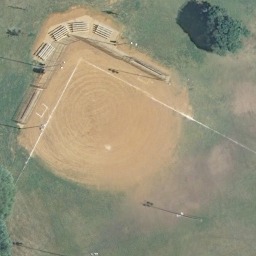}
	}
%	\hspace{1mm}
%	\subfloat[Beach]{
%		\includegraphics[width=\exFigSize\textwidth, keepaspectratio=true]{ucmerced/c_beach04.jpg}
%		\includegraphics[width=\exFigSize\textwidth, keepaspectratio=true]{ucmerced/c_beach13.jpg}
%	}
	\hspace{1mm}
	\subfloat[Buildings]{
		\includegraphics[width=\exFigSize\textwidth, keepaspectratio=true]{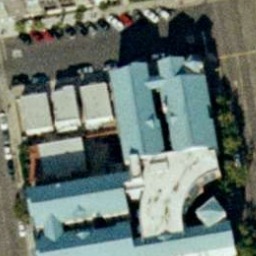}
		\includegraphics[width=\exFigSize\textwidth, keepaspectratio=true]{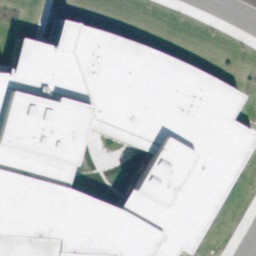}
	}
%	\hspace{1mm}
%	\subfloat[Chaparral]{
%		\includegraphics[width=\exFigSize\textwidth, keepaspectratio=true]{ucmerced/chaparral06.jpg}
%		\includegraphics[width=\exFigSize\textwidth, keepaspectratio=true]{ucmerced/chaparral13.jpg}
%	}
	\hspace{1mm}
	\subfloat[Dense Residential]{
		\includegraphics[width=\exFigSize\textwidth, keepaspectratio=true]{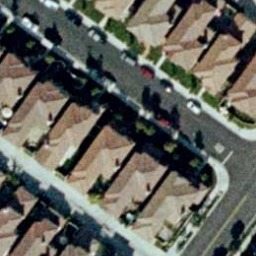}
		\includegraphics[width=\exFigSize\textwidth, keepaspectratio=true]{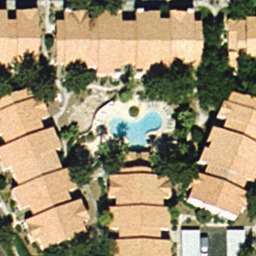}
	}
%	\hspace{1mm}
%	\subfloat[Forest]{
%		\includegraphics[width=\exFigSize\textwidth, keepaspectratio=true]{ucmerced/c_forest04.jpg}
%		\includegraphics[width=\exFigSize\textwidth, keepaspectratio=true]{ucmerced/c_forest28.jpg}
%	}
	\hspace{1mm}
	\subfloat[Freeway]{
		\includegraphics[width=\exFigSize\textwidth, keepaspectratio=true]{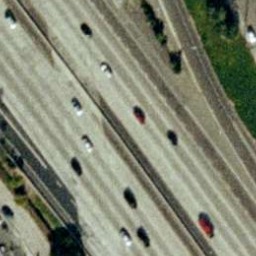}
		\includegraphics[width=\exFigSize\textwidth, keepaspectratio=true]{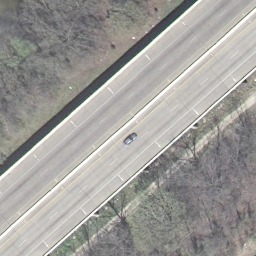}
	}
%	\hspace{1mm}
%	\subfloat[Golf Course]{
%		\includegraphics[width=\exFigSize\textwidth, keepaspectratio=true]{ucmerced/c_golfcourse20.jpg}
%		\includegraphics[width=\exFigSize\textwidth, keepaspectratio=true]{ucmerced/c_golfcourse31.jpg}
%	}
%	\hspace{1mm}
%	\subfloat[Harbor]{
%		\includegraphics[width=\exFigSize\textwidth, keepaspectratio=true]{ucmerced/harbor00.jpg}
%		\includegraphics[width=\exFigSize\textwidth, keepaspectratio=true]{ucmerced/harbor28.jpg}
%	}
	\hspace{1mm}
	\subfloat[Intersection]{
		\includegraphics[width=\exFigSize\textwidth, keepaspectratio=true]{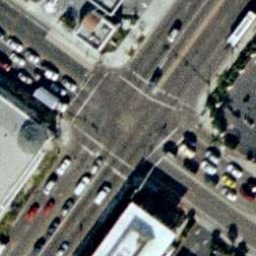}
		\includegraphics[width=\exFigSize\textwidth, keepaspectratio=true]{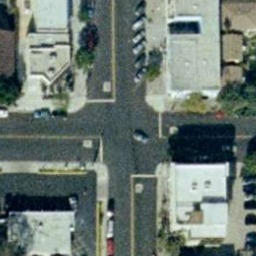}
	}
	\hspace{1mm}
	\subfloat[Medium Residential]{
		\includegraphics[width=\exFigSize\textwidth, keepaspectratio=true]{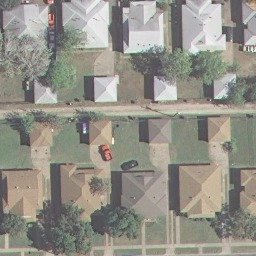}
		\includegraphics[width=\exFigSize\textwidth, keepaspectratio=true]{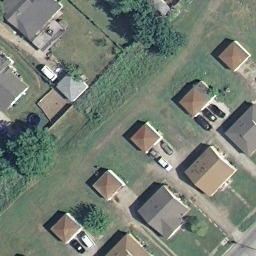}
	}
	\hspace{1mm}
	\subfloat[Mobile Park]{
		\includegraphics[width=\exFigSize\textwidth, keepaspectratio=true]{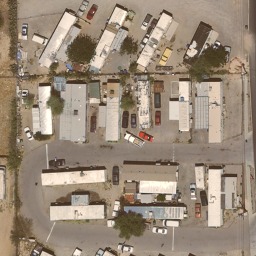}
		\includegraphics[width=\exFigSize\textwidth, keepaspectratio=true]{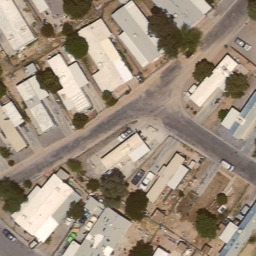}
	}
	\hspace{1mm}
	\subfloat[Overpass]{
		\includegraphics[width=\exFigSize\textwidth, keepaspectratio=true]{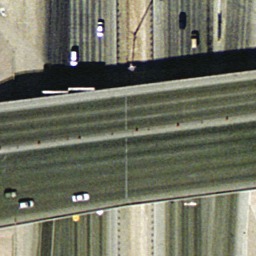}
		\includegraphics[width=\exFigSize\textwidth, keepaspectratio=true]{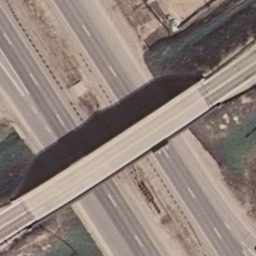}
	}
%	\hspace{1mm}
%	\subfloat[Parking Lot]{
%		\includegraphics[width=\exFigSize\textwidth, keepaspectratio=true]{ucmerced/parkinglot00.jpg}
%		\includegraphics[width=\exFigSize\textwidth, keepaspectratio=true]{ucmerced/parkinglot04.jpg}
%	}
%	\hspace{1mm}
%	\subfloat[River]{
%		\includegraphics[width=\exFigSize\textwidth, keepaspectratio=true]{ucmerced/c_river02.jpg}
%		\includegraphics[width=\exFigSize\textwidth, keepaspectratio=true]{ucmerced/c_river07.jpg}
%	}
%	\hspace{1mm}
%	\subfloat[Runway]{
%		\includegraphics[width=\exFigSize\textwidth, keepaspectratio=true]{ucmerced/c_runway01.jpg}
%		\includegraphics[width=\exFigSize\textwidth, keepaspectratio=true]{ucmerced/c_runway10.jpg}
%	}
	\hspace{1mm}
	\subfloat[Sparse Residential]{
		\includegraphics[width=\exFigSize\textwidth, keepaspectratio=true]{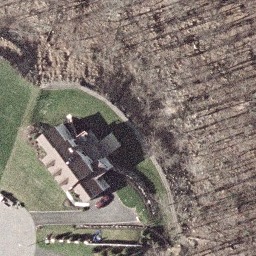}
		\includegraphics[width=\exFigSize\textwidth, keepaspectratio=true]{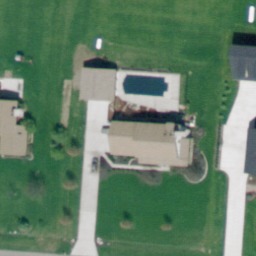}
	}
	\hspace{1mm}
	\subfloat[Storage Tanks]{
		\includegraphics[width=\exFigSize\textwidth, keepaspectratio=true]{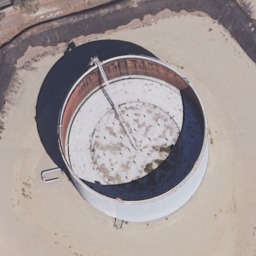}
		\includegraphics[width=\exFigSize\textwidth, keepaspectratio=true]{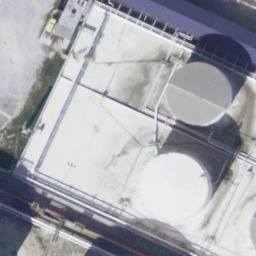}
	}
	\hspace{1mm}
	\subfloat[Tennis Court]{
		\includegraphics[width=\exFigSize\textwidth, keepaspectratio=true]{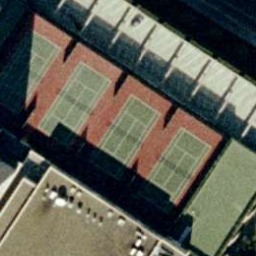}
		\includegraphics[width=\exFigSize\textwidth, keepaspectratio=true]{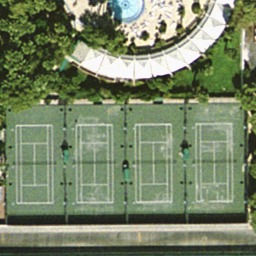}
	}
	\caption{Examples of the UCMerced Land-Use Dataset.}
	\label{fig:merced_dataset}
\end{figure}

\noindent\textbf{WHU-RS19 Dataset.} This public dataset~\cite{xia2010structural} contains 1,005 high-resolution images with $600\times600$ pixels divided into 19 classes (approximately 50 images per class), including: airport, beach, bridge, river, forest, meadow, pond, parking, port, viaduct, residential area, industrial area, commercial area, desert, farmland, football field, mountain, park and railway station.
Exported from Google Earth, that provides high-resolution satellite images up to half a meter, this dataset has samples collected from different regions all around the world, which increases its diversity but creates challenges due to the changes in resolution, scale, and orientation of the images.
Figure~\ref{fig:rs19_dataset} presents examples of some classes.

\begin{figure}[t!]
	\centering
	\scriptsize
	\subfloat[Airport]{
		\includegraphics[width=\exFigSize\textwidth, keepaspectratio=true]{airport_05.jpg}
		\includegraphics[width=\exFigSize\textwidth, keepaspectratio=true]{airport_31.jpg}
	}
	\hspace{1mm}
	\subfloat[Bridge]{
		\includegraphics[width=\exFigSize\textwidth, keepaspectratio=true]{bridge_33.jpg}
		\includegraphics[width=\exFigSize\textwidth, keepaspectratio=true]{bridge_40.jpg}
	}
	\hspace{1mm}
	\subfloat[Commercial]{
		\includegraphics[width=\exFigSize\textwidth, keepaspectratio=true]{commercial_17.jpg}
		\includegraphics[width=\exFigSize\textwidth, keepaspectratio=true]{commercial_42.jpg}
	}
	\hspace{1mm}
	\subfloat[Football Field]{
		\includegraphics[width=\exFigSize\textwidth, keepaspectratio=true]{footballField_05.jpg}
		\includegraphics[width=\exFigSize\textwidth, keepaspectratio=true]{footballField_35.jpg}
	}
	\hspace{1mm}
	\subfloat[Industrial]{
		\includegraphics[width=\exFigSize\textwidth, keepaspectratio=true]{industrial_05.jpg}
		\includegraphics[width=\exFigSize\textwidth, keepaspectratio=true]{industrial_25.jpg}
	}
	\hspace{1mm}
	\subfloat[Park]{
		\includegraphics[width=\exFigSize\textwidth, keepaspectratio=true]{Park_03.jpg}
		\includegraphics[width=\exFigSize\textwidth, keepaspectratio=true]{Park_11.jpg}
	}
	\hspace{1mm}
	\subfloat[Parking]{
		\includegraphics[width=\exFigSize\textwidth, keepaspectratio=true]{parking_35.jpg}
		\includegraphics[width=\exFigSize\textwidth, keepaspectratio=true]{parking_45.jpg}
	}
	\hspace{1mm}
	\subfloat[Pond]{
		\includegraphics[width=\exFigSize\textwidth, keepaspectratio=true]{pond_33.jpg}
		\includegraphics[width=\exFigSize\textwidth, keepaspectratio=true]{pond_49.jpg}
	}
	\hspace{1mm}
	\subfloat[Port]{
		\includegraphics[width=\exFigSize\textwidth, keepaspectratio=true]{port_12.jpg}
		\includegraphics[width=\exFigSize\textwidth, keepaspectratio=true]{port_16.jpg}
	}
	\hspace{1mm}
	\subfloat[Railway Station]{
		\includegraphics[width=\exFigSize\textwidth, keepaspectratio=true]{railwayStation_01.jpg}
		\includegraphics[width=\exFigSize\textwidth, keepaspectratio=true]{railwayStation_14.jpg}
	}
	\hspace{1mm}
	\subfloat[Residential]{
		\includegraphics[width=\exFigSize\textwidth, keepaspectratio=true]{residential_09.jpg}
		\includegraphics[width=\exFigSize\textwidth, keepaspectratio=true]{residential_33.jpg}
	}
	\hspace{1mm}
	\subfloat[Viaduct]{
		\includegraphics[width=\exFigSize\textwidth, keepaspectratio=true]{viaduct_04.jpg}
		\includegraphics[width=\exFigSize\textwidth, keepaspectratio=true]{viaduct_29.jpg}
	}
	\caption{Examples of the WHU-RS19 Dataset.}
	\label{fig:rs19_dataset}
\end{figure}

\subsubsection{Pixel Classification Datasets} \label{subsec:hyper_datasets}

\noindent\textbf{Pavia Centre.} This publicly available dataset~\cite{plaza2009recent} is composed of one image, acquired by the Reflective Optics System Imaging Spectrometer (ROSIS), covering the city of Pavia, southern Italy. This hyperspectral image has 715$\times$1,096 pixels, spatial resolution of 1.3m per pixel, and 102 spectral bands. Pixels of this image are categorized into 9 classes. The false-color and ground-truth images, as well as the number of pixels in each class, are presented in Figure~\ref{fig:hyperspectral_datasets}.

\noindent\textbf{Pavia University.} This public dataset~\cite{plaza2009recent} was also acquired by the ROSIS sensor during a flight campaign over Pavia. This hyperspectral image has 610$\times$340 pixels, spatial resolution of 1.3m per pixel, and 103 spectral bands. Pixels of this image are also categorized into 9 classes. The false-color and ground-truth images, as well as the number of pixels in each class, are presented in Figure~\ref{fig:hyperspectral_datasets}.

For both datasets, in order to reduce the computational complexity, Principal Component Analysis (PCA)~\cite{sonka2014image} was used as a pre-processing method to reduce the dimensionality.
Specifically, following~\cite{jiao2017deep,hu2020spatial}, for both datasets, we selected the first 3 principal components, which explain, approximately, 99\% of the total variance of the data.

\begin{table*}[]
	\begin{center}
		\begin{tabular}{cc||ccc||ccc}
		 \multicolumn{2}{c||}{\textbf{Dataset}} & \multicolumn{3}{c||}{\textbf{Pavia Centre}} & \multicolumn{3}{c}{\textbf{Pavia University}} \\ \hline \hline
		 \multicolumn{2}{c||}{\multirow{-14}{*}{\textbf{Images}}} & \multicolumn{3}{c||}{ \cincludegraphics[width=0.2\textwidth]{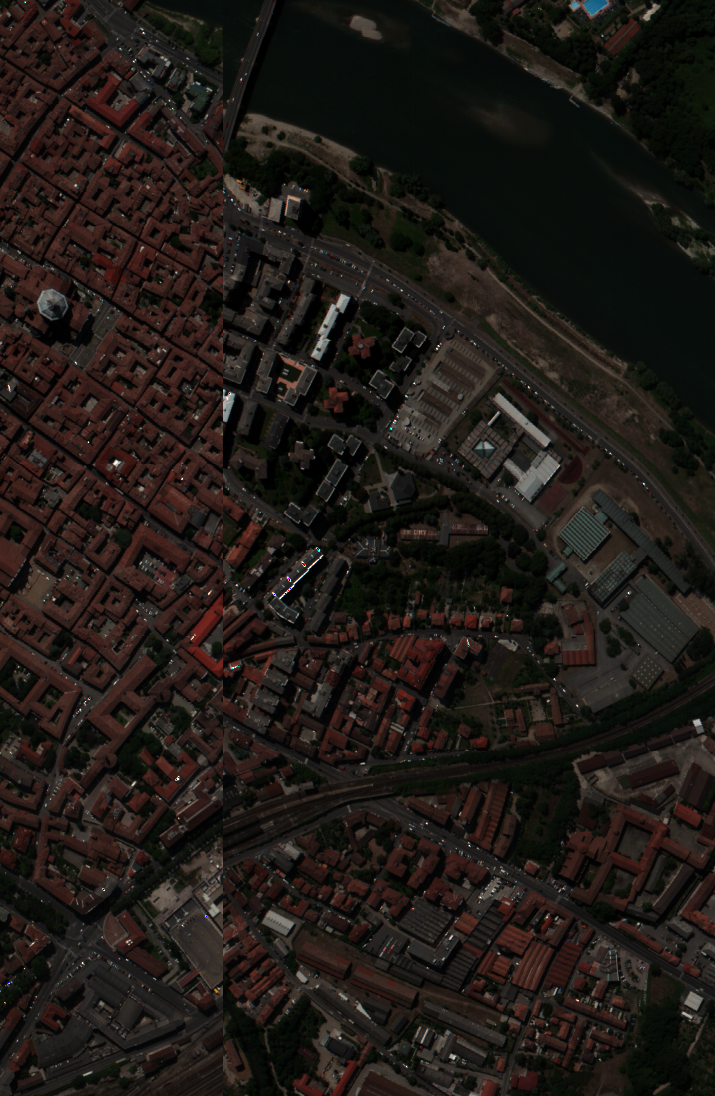} \cincludegraphics[width=0.2005\textwidth]{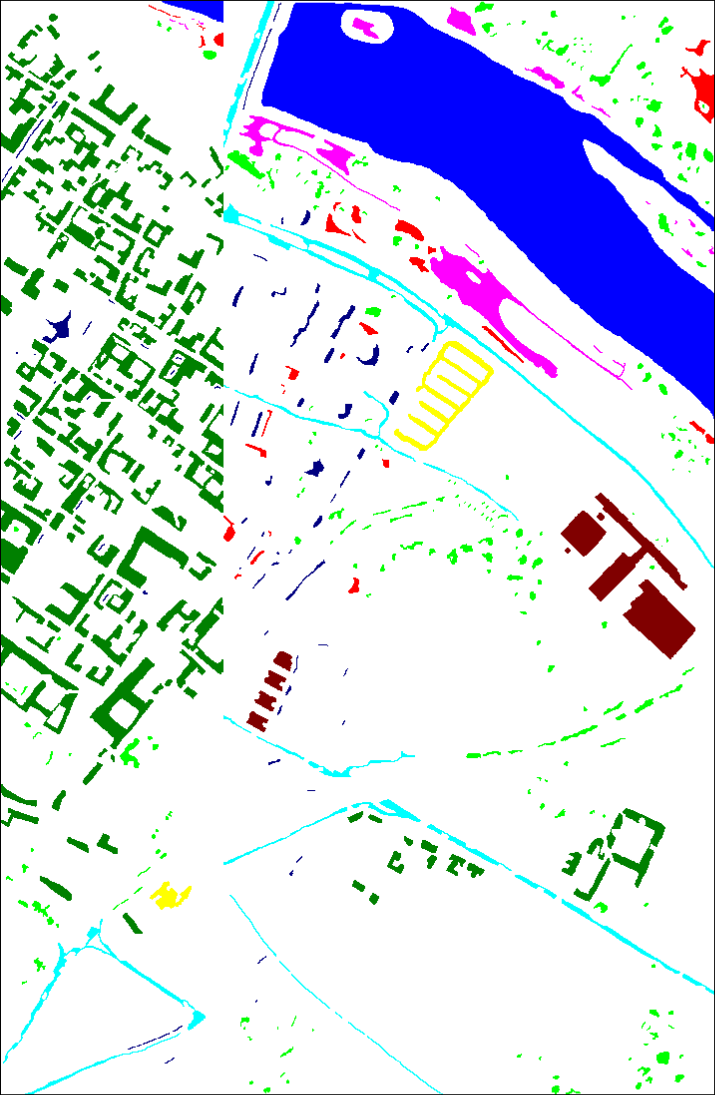} }  & \multicolumn{3}{c}{ \cincludegraphics[width=0.171\textwidth]{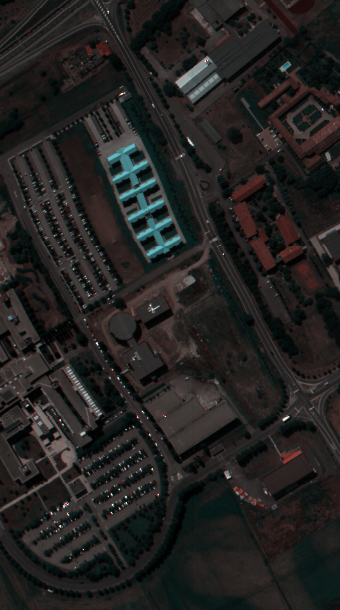} \cincludegraphics[width=0.1718\textwidth]{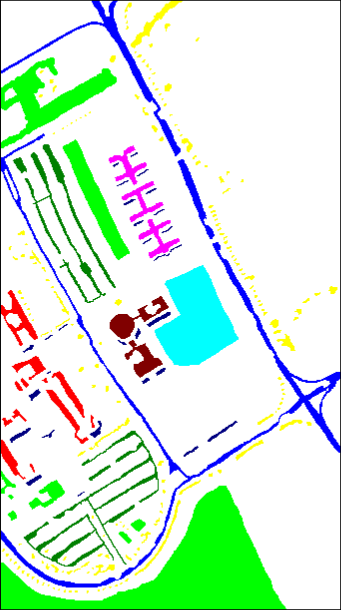} }    \\ \hline \hline
		
		\textbf{No.}         & \textbf{Color} & \textbf{Class} & \multicolumn{1}{c}{\textbf{Train}} & \multicolumn{1}{c||}{\textbf{Total}} & \textbf{Description} & \multicolumn{1}{c}{\textbf{Train}} & \multicolumn{1}{c}{\textbf{Total}} \\ \hline
		1                    & \begin{tabular}[c]{>{\centering\arraybackslash}m{0.18cm}}\cellcolor[HTML]{0000FF}\end{tabular} & Water                & 466                                & 65,505                             & Asphalt              & 141                                & 6,631                             \\
		2                    & \begin{tabular}[c]{>{\centering\arraybackslash}m{0.18cm}}\cellcolor[HTML]{00FF00}\end{tabular} & Trees                & 50                                 & 7,548                              & Meadows              & 447                                & 18,649                            \\
		3                    & \begin{tabular}[c]{>{\centering\arraybackslash}m{0.18cm}}\cellcolor[HTML]{FF0000}\end{tabular} & Asphalt              & 23                                 & 3,067                              & Gravel               & 48                                 & 2,099                             \\
		4                    & \begin{tabular}[c]{>{\centering\arraybackslash}m{0.18cm}}\cellcolor[HTML]{FFFF00}\end{tabular} & Self-Blocking Bricks & 18                                 & 2,667                              & Trees                & 72                                 & 3,064                             \\
		5                    & \begin{tabular}[c]{>{\centering\arraybackslash}m{0.18cm}}\cellcolor[HTML]{FF00FF}\end{tabular} & Bitumen              & 39                                 & 6,545                              & Painted metal sheets & 32                                 & 1,345                             \\
		6                    & \begin{tabular}[c]{>{\centering\arraybackslash}m{0.18cm}}\cellcolor[HTML]{00FFFF}\end{tabular} & Tiles                & 57                                 & 9,191                              & Bare Soil            & 119                                & 5,029                             \\
		7                    & \begin{tabular}[c]{>{\centering\arraybackslash}m{0.18cm}}\cellcolor[HTML]{800000}\end{tabular} & Shadows              & 41                                 & 7,246                              & Bitumen              & 36                                 & 1,330                             \\
		8                    & \begin{tabular}[c]{>{\centering\arraybackslash}m{0.18cm}}\cellcolor[HTML]{008000}\end{tabular} & Meadows              & 284                                & 42,542                             & Self-Blocking Bricks & 85                                 & 3,682                             \\
		9                    & \begin{tabular}[c]{>{\centering\arraybackslash}m{0.18cm}}\cellcolor[HTML]{000080}\end{tabular} & Bare Soil            & 22                                 & 2,841                              & Shadows              & 20                                 & 947                               \\ \hline
		\multicolumn{2}{c||}{\textbf{Total}}                                 &  & 1,000                              & 147,152                            & \multicolumn{1}{l}{} & 1,000                              & 42,776                            \\ \hline
	\end{tabular}
	\end{center}
	\captionof{figure}{In the first (top) part, false-color and ground-truth images of the explored hyperspectral datasets.
		In the second (bottom) part, the classes, the number of pixels used for training the models, and the total amount of labeled pixels.
	}
	\label{fig:hyperspectral_datasets}
\end{table*}

\subsection{Baselines} \label{subsec:baselines}

% For all datasets, two baselines were used.
Several baselines were employed in this work.
%, referenced hereafter as ConvNet,
The first one, exploited in both tasks and all aforementioned datasets, is a standard convolutional version of the corresponding DeepMorphNet.
This baseline, referenced as ConvNet, recreates the exact morphological architecture using the traditional convolutional layer
%(instead of depthwise and pointwise convolutions)
but preserving all remaining configurations (such as filter sizes, padding, stride, etc).
% Furthermore, differently from the morphological networks, 
Moreover, this baseline makes use of max-pooling layers between the convolutions, which makes it very similar to the traditional architectures of the literature~\cite{krizhevsky2012imagenet}.
An extension of this first baseline, referenced as PreMorph-ConvNet and exploited only for the (non-synthetic) image classification datasets, uses pre-defined morphological operations as pre-processing to extract the first set of discriminative features~\cite{aptoula2016deep,wang2018lidar}.
Such data is then used as input to a ConvNet (in this case, the previously described one), which is responsible for performing the final classification.
The third baseline, used only for the (non-synthetic) image classification datasets and referenced hereafter Depth-ConvNet, is exactly the DeepMorphNet architecture but without using binary weights and depthwise pooling.
This baseline reproduces the same DeepMorphNet architecture (also using depthwise and pointwise convolutions) and, consequently, \textbf{has the same number of parameters}, except for the binary weights.
%(i.e., depthwise separable convolution~\cite{chollet2017xception})
\changes{
	Other baselines, used only for the (non-synthetic) image classification datasets, are the deep morphological frameworks proposed by~\cite{mondal2020image,franchi2020deep}.
	In these cases, the DeepMorphNet architecture was recreated, as far as possible, using both techniques.
}
Finally, the last baseline, referenced with the suffix ``SK'' and explored only for the (non-synthetic) image classification datasets, is the convolutional and morphological networks, but with Selective Kernels (SK)~\cite{li2019selective}, which allows the network to give more importance to certain feature maps than to others.
% Note that those two last baselines were not employed for the synthetic datasets, being used only for the other ones.

Differently from the morphological networks, all baselines use batch normalization~\cite{goodfellow2016deep} (after each convolution) and Rectified Linear Units (ReLUs)~\cite{nair2010rectified} as activation functions.
It is important to note that, despite the differences, all baselines have the exact same number of layers and feature maps of the base DeepMorphNet.
We believe that this conservation allows a fair comparison between the models given that the potential representation of the networks is somehow the same.

%A first baseline used for this dataset, is the convolutional version of the proposed DeepMorphNet.
%Specifically, the network has 5 layers, each one composed of convolution operation (with the same configuration of the morphological layers), batch normalization~\cite{ioffe2015batch}, and Rectified Linear Unit (ReLU)~\cite{nair2010rectified}.
%After the first, the second and the fifth layers, there are max-pooling operations~\cite{goodfellow2016deep}.
%In the end, three fully connected (FC) layers, with the exact same configuration of the DeepMorphNets, are responsible to perform the final classification.
%Both architectures were inspired in the famous AlexNet~\cite{krizhevsky2012imagenet} one and were trained using the following hyper-parameters: learning rate, weight decay, momentum, and number of epochs of 0.01, 0.0005, 0.9, and 100, respectively.

%In order to make the comparison fair, all networks have the same number of layers and feature maps.

%Although the DeepMorphNet can have more neurons because of its natural design (as explained in Section~\ref{sec:arch}), the number of generated feature maps is the same.
%Thus, we believe this makes the comparison fair since the representational power provides from one layer to another is based on the number of feature maps generated by the former.

%%%%%%%%%%%%%%%%%%%%%%%%%%%%%%%%%%%%
\subsection{Experimental Protocol} \label{subsec:protocol}

For the synthetic datasets, a train/validation/test protocol was employed.
In this case, 60\% of the instances were used as training, 20\% as validation, and another 20\% as test.
%For synthetic datasets, a simple protocol was employed.
%Particularly, in this case, the whole dataset is randomly divided into three sets: training (composed of 60\% of the instances), validation and test (each one composed of 20\% of the dataset samples).
%%Once determined, these sets are used throughout the experiments for all networks and baselines.
Results of this protocol are reported in terms of the average accuracy of the test set.
%%%%%%For the all datasets, a validation was performed dividing the samples into three fixed-size sets: training (with 80\% of the samples), validation, and test (each one with 10\% of the dataset size).
For the other image classification datasets, five-fold cross-validation was conducted to assess the accuracy of the proposed algorithm.
%In this protocol, the dataset is split into five folds with almost the same size, where each one is balanced according to the number of images per class, giving diversity to each set.
%Those folds are processed in five different runs where, at each run, three distinct folds are used as the training set, one as validation (used to tune the parameters of the network) and the remaining one is used as the test set.
The final results are the mean of the average accuracy (for the test set) of the five runs followed by its corresponding standard deviation.
Finally, for the pixel classification datasets, following previous works~\cite{ji2019semisupervised,feng2019cnn,hu2020spatial}, we performed a random sampling to select 1,000 training samples/pixels from all classes, as presented in (the bottom part of) Figure~\ref{fig:hyperspectral_datasets}.
% 0.02\% of samples in the Pavia Centre and approximately 2\% of the Pavia University.
All remaining pixels are used to assess the tested architectures.
The final results are reported in terms of the average accuracy of those remaining pixels.

All networks proposed in this work were implemented using PyTorch.
%\footnote{Torch is a scientific computing framework with wide support for machine learning algorithms available at~\url{http://torch.ch/} (as of May 2019).}.
%This framework is more suitable due to its support to parallel programming using CUDA, an NVidia parallel programming based on Graphics Processing Units.
All experiments were performed on a 64 bit Intel i7 5930K machine with 3.5GHz of clock, Ubuntu 18.04.1 LTS, 64GB of RAM memory, and a GeForce GTX Titan X Pascal with 12GB of memory under a 10.0 CUDA version.
% Ubuntu version 18.04.1 LTS was used as operating system.

\section{Results and Discussion} \label{sec:results}

In this section, we present and discuss the outcomes.
Section~\ref{subsec:synthetic_res} presents the results of the synthetic datasets while Section~\ref{subsec:image_class} discusses the results of the other datasets.
Finally, Section~\ref{subsec:pixel_classification} presents the results of the pixel classification datasets.
% chanussot
Please remember that we present a proof of concept of a new paradigm: using non-linear morphological operations instead of standard linear convolutions for a deep network.
To demonstrate the benefits of this proposal, similar architectures must be considered, including in terms of the number of neurons, layers, etc.
In order not to confuse the benefits of proposal with the advantages of complex networks, as introduced, the proposed models were inspired by the simple AlexNet~\cite{krizhevsky2012imagenet} architecture.

\subsection{Synthetic Image Classification Datasets} \label{subsec:synthetic_res}

As explained in Section~\ref{subsec:syn_datasets}, two synthetic datasets were proposed in this work to validate the feature learning of the deep morphological networks.
Furthermore, as introduced, both datasets can be perfectly classified using one opening with specific SEs.
%Based on this, we designed a basic DeepMorphNet composed of two layers: the first one has a unique neuron performing opening (using structuring element of $11\times11$) while the second one is a fully connected layer (with two neurons, one for each class) responsible for the final classification.
Supported by this, the proposed DeepMorphSynNet, composed of one opening neuron, can be used to validate the feature learning process of the proposed technique, given that this network has the capacity of perfectly classifying the datasets as long as it successfully learns the SE.
%given that this network should be able to learn the expected SE and then perfectly classified the datasets.

Given the simplicity of these datasets, aside from the first method describe in Section~\ref{subsec:baselines}, we also employed as baseline a basic architecture composed uniquely of a classification layer.
Specifically, this network has one layer that receives a linearized version of the input data and outputs the classification.
The proposed morphological network and baselines were tested for both synthetic datasets using the same configuration, i.e., learning rate, weight decay, momentum, and number of epochs of 0.01, 0.0005, 0.9, and 10, respectively.

%A convolutional version of this network was implemented using exactly the same configuration but with a regular convolution operation instead of an opening.
%In order to show that the problem is not so easy, we also run experiments using only the final classification layer (that ), as in a multi-layer perceptron network (MLP)~\cite{goodfellow2016deep}.

\begin{figure}[h!]
	\hspace{-1cm}
	\begin{minipage}{.5\linewidth}
		\centering
		\subfloat[Learned SE]{
			\label{fig:se_square}
			\includegraphics[width=0.45\textwidth, keepaspectratio=true]{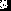}
		}
	\end{minipage}%
	\hspace{-1.2cm}
	\begin{minipage}{.5\linewidth}
		\centering
		\subfloat[Class 1: Square $5\times 5$]{
			\label{fig:sq_class1}
			\includegraphics[width=0.45\textwidth, keepaspectratio=true]{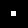}
			\includegraphics[width=0.45\textwidth, keepaspectratio=true]{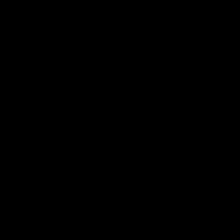}
			\includegraphics[width=0.45\textwidth, keepaspectratio=true]{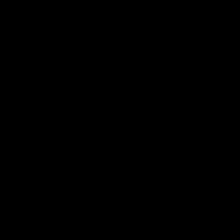}
		}
		\hspace{1mm}
		\subfloat[Class 2: Square $9\times 9$]{
			\label{fig:sq_class2}
			\includegraphics[width=0.45\textwidth, keepaspectratio=true]{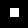}
			\includegraphics[width=0.45\textwidth, keepaspectratio=true]{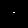}
			\includegraphics[width=0.45\textwidth, keepaspectratio=true]{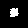}
		}
	\end{minipage}
	
	\caption{Visual results for the square synthetic dataset.
		(a) The learned SE. (b)-(c) Examples of the output of the opening neuron for two classes of the square synthetic dataset.
		The first column represents the input image, the second one is the output of the erosion, and the last one is the output of the dilation.
		Since erosion and dilation have tied weights (i.e., the same SE), they implement an opening.}
	\label{fig:square_syn_image}
\end{figure}

Results for the synthetic \textbf{square} dataset are presented in Table~\ref{tab:synthetic_results}.
Among the baselines, the worst results were generated by the ConvNets while the best outcome was produced by the network composed of a single classification layer (86.50\%).
%Both baselines that have convolution layers generated worse results.
%Precisely, the ConvNet and the DepthConvNet (composed of depthwise separable convolutions~\cite{chollet2017xception}) yielded 58 and 64\% of average accuracy, respectively.
A reason for this is that the proposed dataset does not have much visual information to be extracted by the convolution layer.
Hence, in this case, the raw pixels themselves are able to provide relevant information for the classification.
% (such as the total amount of pixels of the square) for the classification.
%However, the result yielded by the best baseline (the network composed of the classification layer) was worse than the result generated by the proposed morphological network.
However, the proposed morphological network was able to outperform all baselines.
Precisely, the DeepMorphSynNet yielded a 100\% of average accuracy, perfectly classifying the entire test set of this synthetic dataset.
As introduced in Section~\ref{subsec:syn_datasets}, in order to achieve this perfect classification, the opening would require a square SE larger than $5\times 5$ but smaller than $9\times 9$ pixels.
As presented in Figure~\ref{fig:se_square}, this was the SE learned by the network.
Moreover, as introduced, with this SE, the opening would erode the small $5\times 5$ squares while keeping the larger $9\times 9$ ones, the exact outcome of the morphological network, as presented in Figures~\ref{fig:sq_class1} and~\ref{fig:sq_class2}.
% Figure~\ref{fig:square_syn_image}.

\begin{table}[]
	\caption{Results, in terms of average accuracy, of the proposed method and the baselines for the synthetic datasets.}
	\label{tab:synthetic_results}
	\centering
	\begin{tabular}{@{}clr@{}}
		\toprule
		\multicolumn{1}{c}{\textbf{Dataset}} & \multicolumn{1}{c}{\textbf{Method}} & \multicolumn{1}{c}{\textbf{Average Accuracy (\%)}} \\ \midrule
		\multirow{3}{*}{\textbf{Square}}           
		& \textbf{Classification Layer}           & 86.50                                     \\
		& \textbf{ConvNet}                        & 58.00                                     \\
		%& \textbf{Depth-ConvNet}                   & 60.00                                     \\
		& \textbf{DeepMorphSynNet}                & \textbf{100.00}                           \\ 
		\midrule
		\multirow{3}{*}{\textbf{Rectangle}}           
		& \textbf{Classification Layer}           & 52.00                                     \\
		& \textbf{ConvNet}                        & \textbf{100.00}                            \\
		%& \textbf{Depth-ConvNet}                   & \textbf{100.00}                           \\
		& \textbf{DeepMorphSynNet}                & \textbf{100.00}                           \\ 
		\bottomrule
	\end{tabular}
\end{table}

\begin{figure}[h!]
	\hspace{-1cm}
	\begin{minipage}{.5\linewidth}
		\centering
		\subfloat[Learned SE]{
			\label{fig:se_rec}
			\includegraphics[width=0.45\textwidth, keepaspectratio=true]{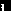}
		}
	\end{minipage}%
	\hspace{-1.2cm}
	\begin{minipage}{.5\linewidth}
		\centering
		\subfloat[Class 1: Rectangle $7\times 3$]{
			\label{fig:rec_class1}
			\includegraphics[width=0.45\textwidth, keepaspectratio=true]{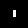}
			\includegraphics[width=0.45\textwidth, keepaspectratio=true]{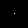}
			\includegraphics[width=0.45\textwidth, keepaspectratio=true]{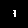}
		}
		\hspace{1mm}
		\subfloat[Class 2: Rectangle $3\times 7$]{
			\label{fig:rec_class2}
			\includegraphics[width=0.45\textwidth, keepaspectratio=true]{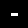}
			\includegraphics[width=0.45\textwidth, keepaspectratio=true]{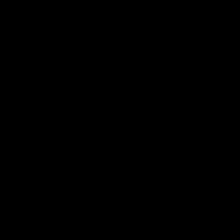}
			\includegraphics[width=0.45\textwidth, keepaspectratio=true]{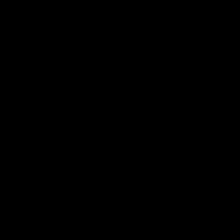}
		}
	\end{minipage}
	
	\caption{Visual results for the rectangle synthetic dataset.
		(a) The learned SE. (b)-(c) Examples of the output of the opening neuron for two classes of the square synthetic dataset.
		The first column represents the input image, the second one is the output of the erosion, and the last one is the output of the dilation.
		Since erosion and dilation have tied weights (i.e., the same SE), they implement an opening.}
	\label{fig:square_rec_image}
\end{figure}

Results for the synthetic \textbf{rectangle} dataset are presented in Table~\ref{tab:synthetic_results}.
%Differently from the synthetic square dataset, the network composed of a single classification layer produced the worst outcome while the convolutional architectures yielded perfect results (100\%).
%the best results among the baselines were generated by the convolutional architectures with the network composed uniquely of a classification layer producing the worst result.
%Precisely, the best outcome was generated by the DepthConvNet (94\%) followed closely by the ConvNet (93\%).
%This difference may be justified by the fact that this is a more complex dataset that has two classes with equal shapes but differing in other relevant properties (such as the orientation) that may be extracted by the convolution layers.
In this case, the proposed DeepMorphSynNet and the ConvNet baseline produced perfect results, with the classification layer producing the worst result.
Such results may be justified based on properties of this dataset: the distinct orientation of the rectangles may help both DeepMorphSynNet and ConvNet, while the exact same number of pixels for both rectangle classes may hinder the classification layer.
% be justified by the this is a more complex dataset and a single classification layer is not capable of per
%Also different from previous outcomes, in this case, the proposed DeepMorphSynNet and best baselines produced the same results (100\% of average accuracy), perfectly classifying this synthetic dataset.
%However, similar to the previous outcome, the result yielded by the best baseline (DepthConvNet) was worse than the result generated by the proposed morphological network.
%In fact, the DeepMorphSynNet yielded a 100\% of average accuracy, also perfectly classifying this synthetic dataset.
%Different from the previous results, in this scenario, the gap between ConvNet and DeepMorphNet was smaller.
As introduced in Section~\ref{subsec:syn_datasets}, to perform this perfect classification, the opening operation (of the DeepMorphSynNet) would require a specific SE that should have the same orientation of one of the rectangles.
As presented in Figure~\ref{fig:se_rec}, this is the SE learned by the morphological network.
With such filter, the opening operation would erode the one type of rectangles while keeping the other, the exact outcome presented in Figures~\ref{fig:rec_class1} and~\ref{fig:rec_class2}. 

%Results obtained with the synthetic datasets show that the proposed morphological networks are able to optimize and learn interesting SEs.
%Furthermore, in some scenarios, such as those simulated by the synthetic datasets, the DeepMorphNets have achieved promising results.
%A better analysis of the potential of the proposed technique is performed in the next section using real (not synthetic) datasets.

\subsection{Image Classification Datasets} \label{subsec:image_class}

For the image classification datasets, all networks were tested using essentially the same configuration, i.e., batch size, learning rate, weight decay, momentum, and number of epochs of 16, 0.01, 0.0005, 0.9, and 2,000 respectively.
Aside from those approaches, we also employed as baseline an approach, called hereafter \textbf{Static SEs}, that reproduces the exactly morphological architecture but with static (non-optimized) SEs.
In this case, each neuron has the configuration based on the most common SEs (mentioned in Section~\ref{sec:background}).
The features extracted by these static neurons are classified by a Support Vector Machine (SVM).
The idea behind this baseline is to have a lower bound for the morphological network, given that this proposed approach should be able to learn better SEs and produce superior results.
%Aside from the ConvNet, another baseline was created using the same DeepMorphNet architecture, but without optimizing the parameters related to the morphological layers, i.e., only parameters related to the FC layers were trained.

% The only difference in the experiments is related to the number of epochs.
% Precisely, the LeNet-based architectures were trained using 500 epochs while for the AlexNet-based networks, epochs were employed.

\subsubsection{UCMerced Land-use Dataset} \label{subsec:ucmerced_res}

Results for the UCMerced Land-use dataset are reported in Table~\ref{tab:ucmerced_results}.
In this case, all networks outperformed the lower bound result, generated by the Static SEs.
%, an expected outcome given the feature learning step performed by the deep learning-based approaches.
%Among the baselines that do not use selective kernels, the best outcome was produced by the Depth-AlexNet-based architecture.
\changes{
	Additionally, the MorphoN~\cite{mondal2020image} and DMNN~\cite{franchi2020deep} networks were outperformed by all other approaches. 
%	 which, in turn, yielded very similar results.
	In fact, all other experimented techniques, including the DeepMorphNets and the networks with Selective Kernels~\cite{li2019selective}, yielded very similar results.
}
% However, the DeepMorphNets yielded similar results (when compared to the respective baseline), 
This indicates that the proposed method is capable of optimizing the morphological filters to extract salient and relevant features, generating satisfactory outcomes.
% generating comparable results.
In order to grasp the difference between the information captured by DeepMorphNets and ConvNets, we performed a comparison, presented in Figure~\ref{fig:ucmerced_feature_maps}, between the feature maps of such networks.
As can be observed, there is a clear difference between the characteristics learned by the distinct networks.
In general, the DeepMorphNet is able to preserve different features when compared to the ConvNets, which corroborates with our initial analysis.

%Considering the LeNet-based networks, the best result, among the baselines, was produced by the Depth-LeNet.
%%architecture based on depthwise separable convolutions~\cite{chollet2017xception}, i.e., the Depth-LeNet.
%The proposed DeepMorphLeNet yielded similar results when compared to this baseline, which shows the potential of the proposed technique that optimizes the morphological filters to extract salient and relevant features.

%The exact same conclusions can be drawn from the AlexNet-based networks.
%Specifically, the best baseline was the Depth-AlexNet-based, that produces 73.14$\pm$1.43\% of average accuracy.
%However, again, the proposed DeepMorphAlexNet yielded competitive results when compared to this baseline (76.86$\pm$1.97\% of average accuracy).
%These results show that morphological operations are able to learn useful features.
%Some of these feature maps of the AlexNet-based architectures are presented in Figure~\ref{fig:ucmerced_feature_maps}.
%Note the difference between the characteristics learned by the distinct networks.
%In general, the morphological network is able to preserve different features when compared to the ConvNets.

\begin{table}
	\caption{Results, in terms of accuracy, of the proposed method and the baselines for the image classification datasets.}
	\label{tab:ucmerced_results}
	\centering
	\resizebox{\columnwidth}{!}{
	\begin{tabular}{@{}clrrr@{}}
		\toprule
		\multicolumn{1}{c}{\textbf{Dataset}} & \multicolumn{1}{c}{\textbf{Method}} & 
		\begin{tabular}[c]{@{}c@{}}\textbf{Average}\\ \textbf{Accuracy (\%)}\end{tabular} &
		\begin{tabular}[c]{@{}c@{}}\textbf{Number of}\\ \textbf{Parameters} \\ \textbf{(millions)}\end{tabular} & 
		\begin{tabular}[c]{@{}c@{}}\textbf{Training}\\ \textbf{Time}\\ \textbf{(hours per fold)}\end{tabular} \\
		\midrule
		\multirow{8}{*}{\textbf{UCMerced}}
		%& \textbf{Static SEs}   			         	  		      & 19.46$\pm$2.06 &  -     &   -   \\
		%& \textbf{LeNet}~\cite{lecun1998gradient}          		  & 53.29$\pm$0.86 &  4.42  &  1.2   \\
		%& \textbf{Depth-LeNet}                               		  & 54.81$\pm$1.25 &  5.04  &  6.2   \\
		%& \textbf{DeepMorphLeNet (ours)}                            & 56.52$\pm$1.74 &  6.04  &  7.8   \\
		%\cmidrule{2-5}
		& \textbf{Static SEs}   			         	  		    		& 28.21$\pm$2.64 &  -     &   -   \\

		& \textbf{ConvNet}~\cite{krizhevsky2012imagenet}      				& 74.10$\pm$2.06 &  6.50  &    8.5   \\
		& \textbf{PreMorph-ConvNet}~\cite{wang2018lidar}    				& 74.31$\pm$1.36 &  6.50  &    8.5   \\

		& \textbf{Depth-ConvNet}                              				& 75.24$\pm$2.17 &  7.47  &  101.1   \\ % 7.47
		
		& \textbf{MorphoN}~\cite{mondal2020image}                       	& 34.52$\pm$4.93 &  0.31  &  1.0   \\
		& \textbf{DMNN}~\cite{franchi2020deep}                              & 54.81$\pm$6.08 &  0.50  &  4.5   \\
		
		& \textbf{DeepMorphNet (ours)}   				       				& 76.68$\pm$2.41 & 10.50  &  209.9   \\ 
		\cmidrule{2-5}
		& \textbf{ConvNet-SK}~\cite{li2019selective}      					& 75.05$\pm$2.26 &  9.36  &   12.1  \\ % 2.867.200
		& \textbf{DeepMorph-SK (ours)}				      					& 76.96$\pm$2.70 & 13.37  &  251.6   \\

		\midrule
		\midrule

		\multirow{8}{*}{\textbf{WHU-RS19}} 
		%& \textbf{Static SEs}   			         	  		  	  & 15.17$\pm$2.83 &  -   &   -   \\
		%& \textbf{LeNet}~\cite{lecun1998gradient}          		  & 48.26$\pm$2.01 &  4.42  &  0.6   \\
		%& \textbf{Depth-LeNet}                               		  & 47.19$\pm$2.43 &  5.04  &  3.1   \\
		%& \textbf{DeepMorphLeNet (ours)}                            & 52.91$\pm$2.60 &  6.04  &  3.7   \\
		%\cmidrule{2-5}
		& \textbf{Static SEs}   			         	  		  	& 25.33$\pm$2.95 &  -   &   -   \\

		& \textbf{ConvNet}~\cite{krizhevsky2012imagenet}			& 64.38$\pm$2.93 &  6.50  &   4.7   \\
		& \textbf{PreMorph-ConvNet}~\cite{wang2018lidar}    		& 66.35$\pm$2.91 &  6.50  &    8.5   \\

		& \textbf{Depth-ConvNet}                             		& 65.25$\pm$2.12 &   7.47  &  44.7   \\ % 7.47
		
		& \textbf{MorphoN}~\cite{mondal2020image}                	& 40.88$\pm$2.60 &  0.31  &  0.5   \\
		& \textbf{DMNN}~\cite{franchi2020deep}                      & 56.31$\pm$1.26 &  0.50  &  3.7   \\
		
		& \textbf{DeepMorphNet (ours)}   				        	& 67.20$\pm$2.75 &  10.50  &  99.8   \\
		\cmidrule{2-5}
		& \textbf{ConvNet-SK}~\cite{li2019selective}      			& 72.06$\pm$2.27 &  9.36  &    6.6   \\
		& \textbf{DeepMorphNet-SK (ours)}					 		& 74.80$\pm$2.36 & 13.37  &  123.1   \\
		\bottomrule
	\end{tabular}
	}
\end{table}

%Some of these optmizable structuring elements are presented in Figure~\ref{fig:ses_ucmerced}.
%
%\begin{figure}[h!]
%	\centering
%	\includegraphics[width=0.12\textwidth, keepaspectratio=true]{ucmerced/layer1_weight1_dilation.png}
%	\includegraphics[width=0.12\textwidth, keepaspectratio=true]{ucmerced/layer1_weight1_erosion.png}
%	\caption{Some interesting learned structuring elements for the UCMerced dataset.}
%	\label{fig:ses_ucmerced}
%\end{figure}

\newcommand{\exFeatureMap}{0.15}
\begin{table*}[t]
	\begin{center}
		%%\begin{tabular}{>{\centering\arraybackslash} m{0.6cm} | >{\centering\arraybackslash}m{2.4cm} >{\centering\arraybackslash}m{2.4cm} >{\centering\arraybackslash}m{2.4cm} >{\centering\arraybackslash}m{2.4cm} >{\centering\arraybackslash}m{2.4cm} >{\centering\arraybackslash}m{2.4cm} @{}m{0pt}@{} } 
		\begin{tabular}{>{\centering\arraybackslash}m{2.4cm} >{\centering\arraybackslash}m{2.4cm} >{\centering\arraybackslash}m{2.4cm} >{\centering\arraybackslash}m{2.4cm} >{\centering\arraybackslash}m{2.4cm} >{\centering\arraybackslash}m{2.4cm} @{}m{0pt}@{} } 
			%%\textbf{Image} & 
			\textbf{Input Image} & \textbf{Layer 1} & \textbf{Layer 2} & \textbf{Layer 3} & \textbf{Layer 4} & \textbf{Layer 5} & \\
			%\hline
			%%\textbf{1} &
			\cincludegraphics[width=\exFeatureMap\textwidth]{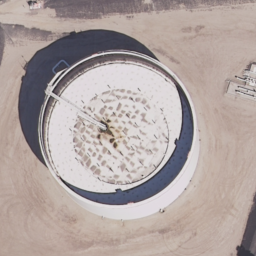} &
			\cincludegraphics[width=\exFeatureMap\textwidth]{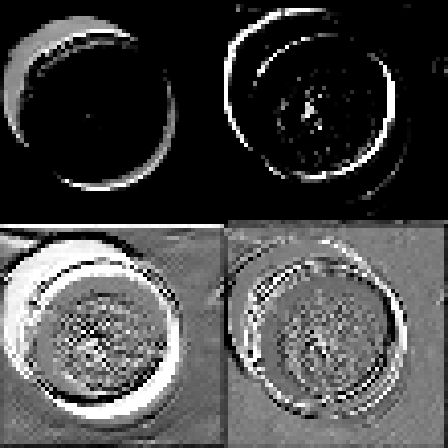} &
			\cincludegraphics[width=\exFeatureMap\textwidth]{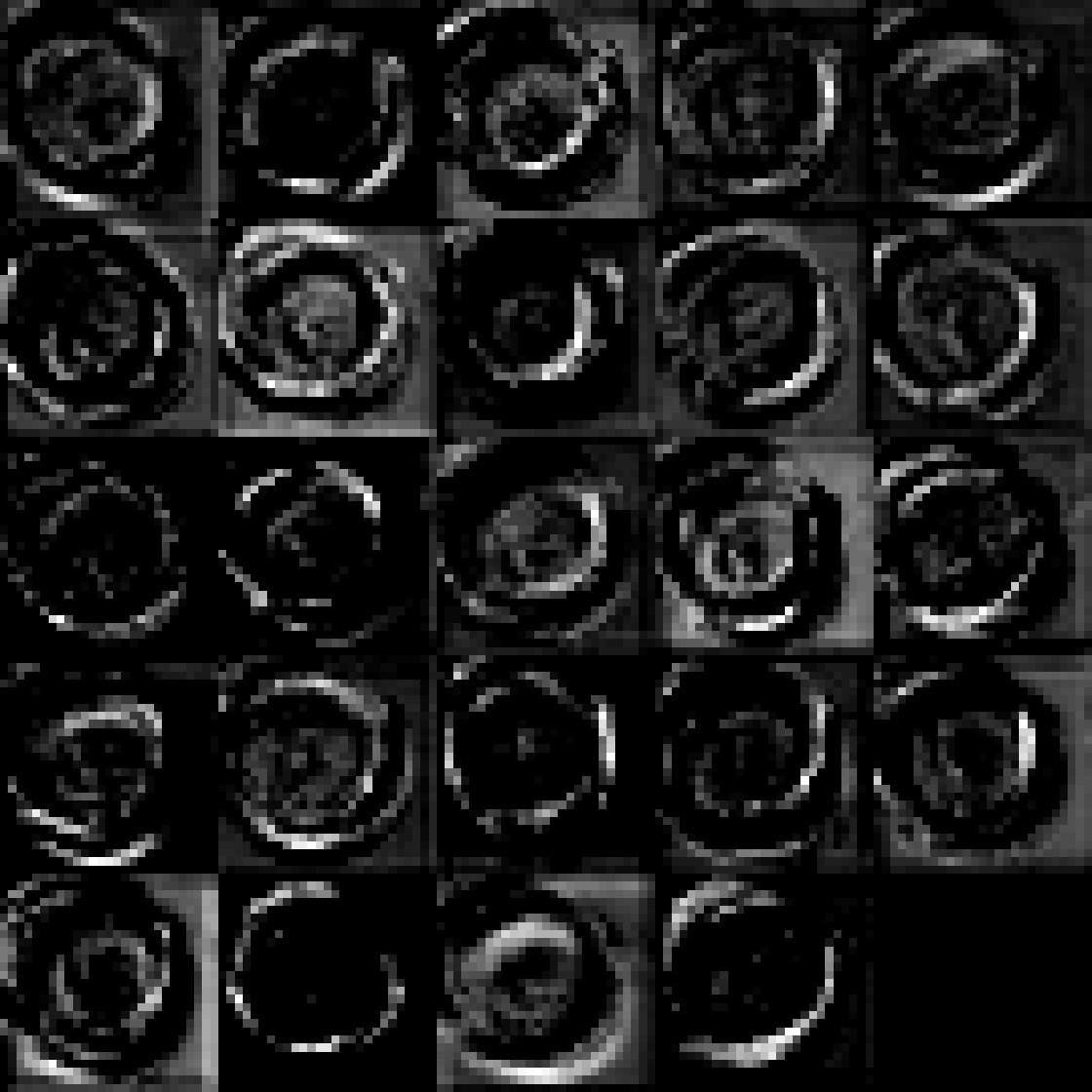} &
			\cincludegraphics[width=\exFeatureMap\textwidth]{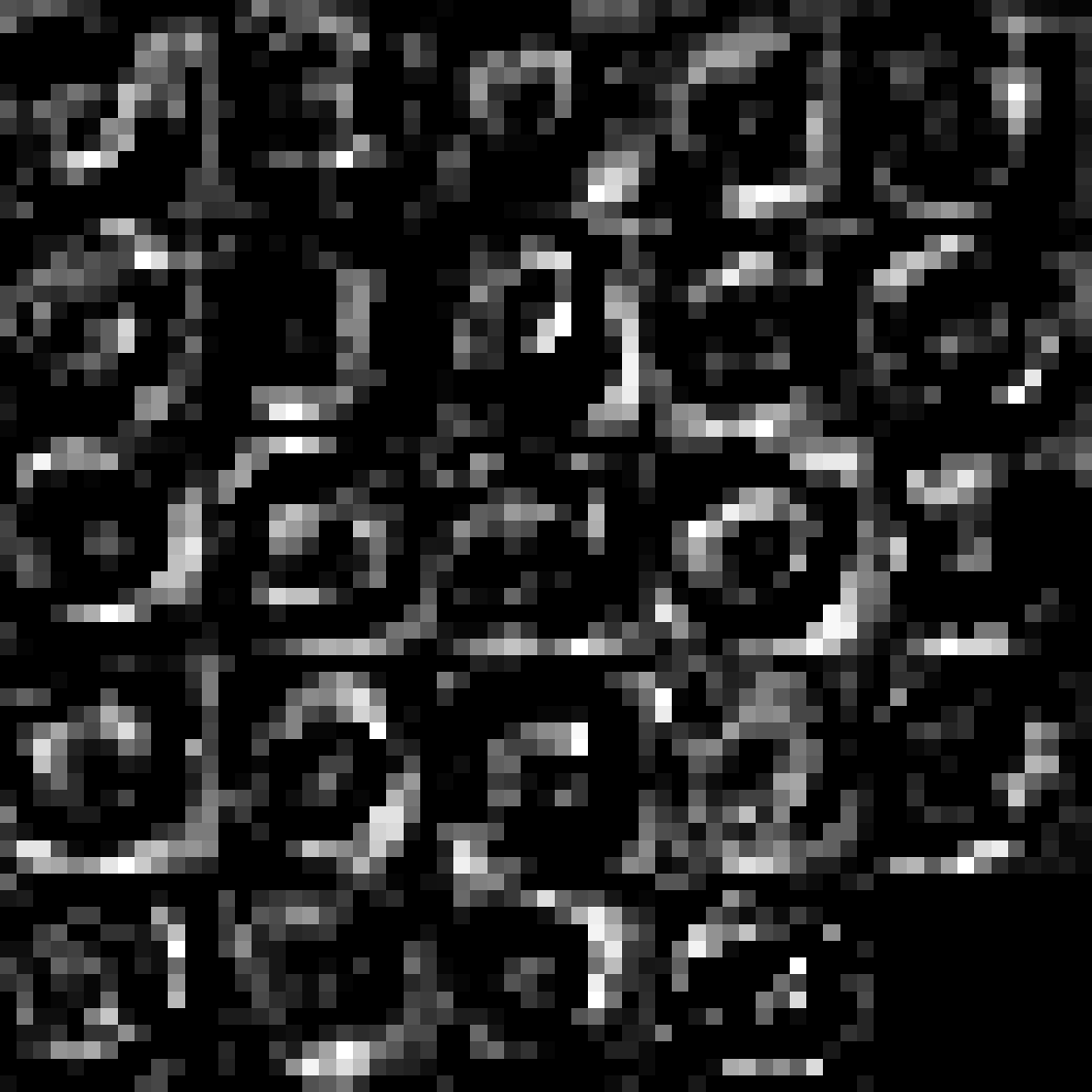} &
			\cincludegraphics[width=\exFeatureMap\textwidth]{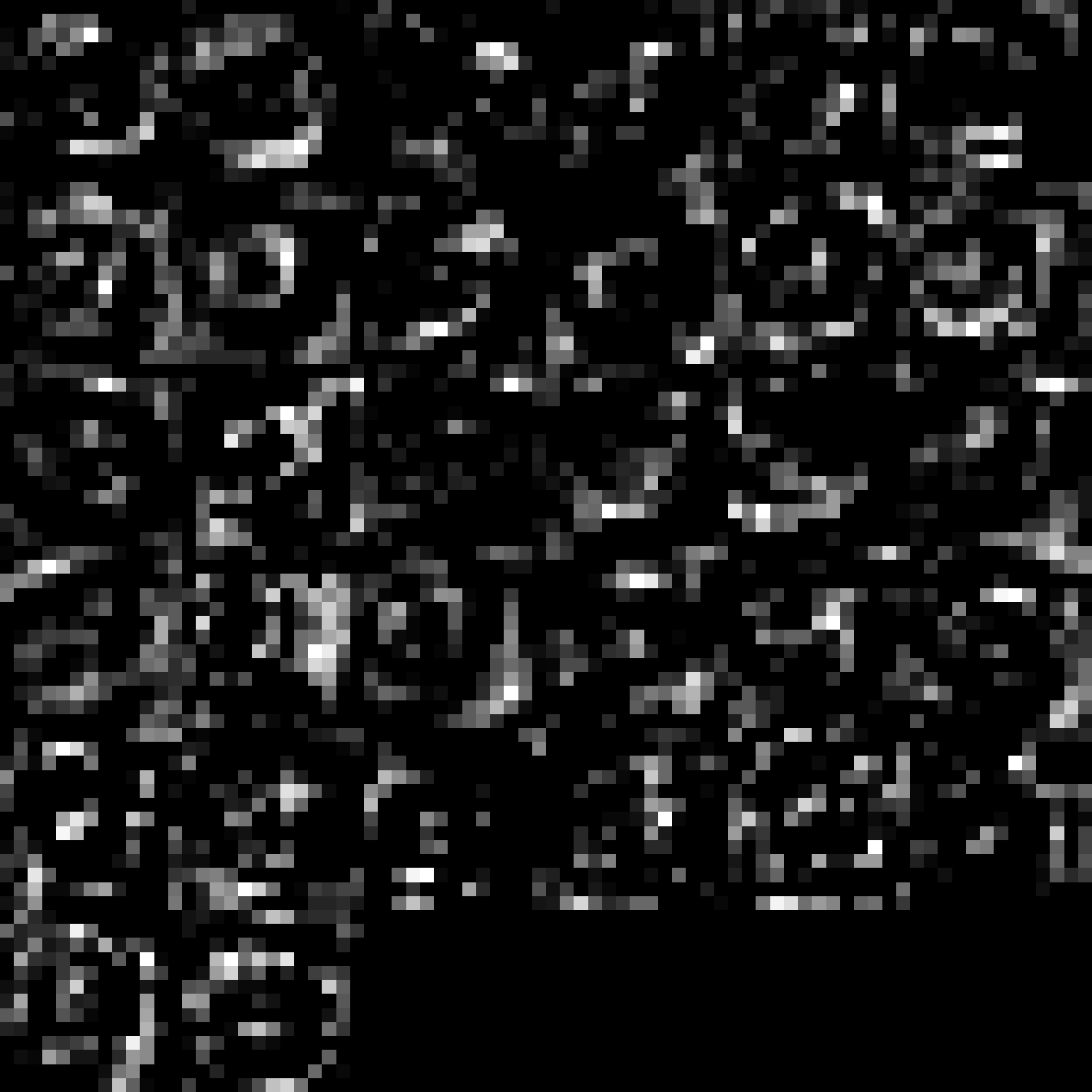} &
			\cincludegraphics[width=\exFeatureMap\textwidth]{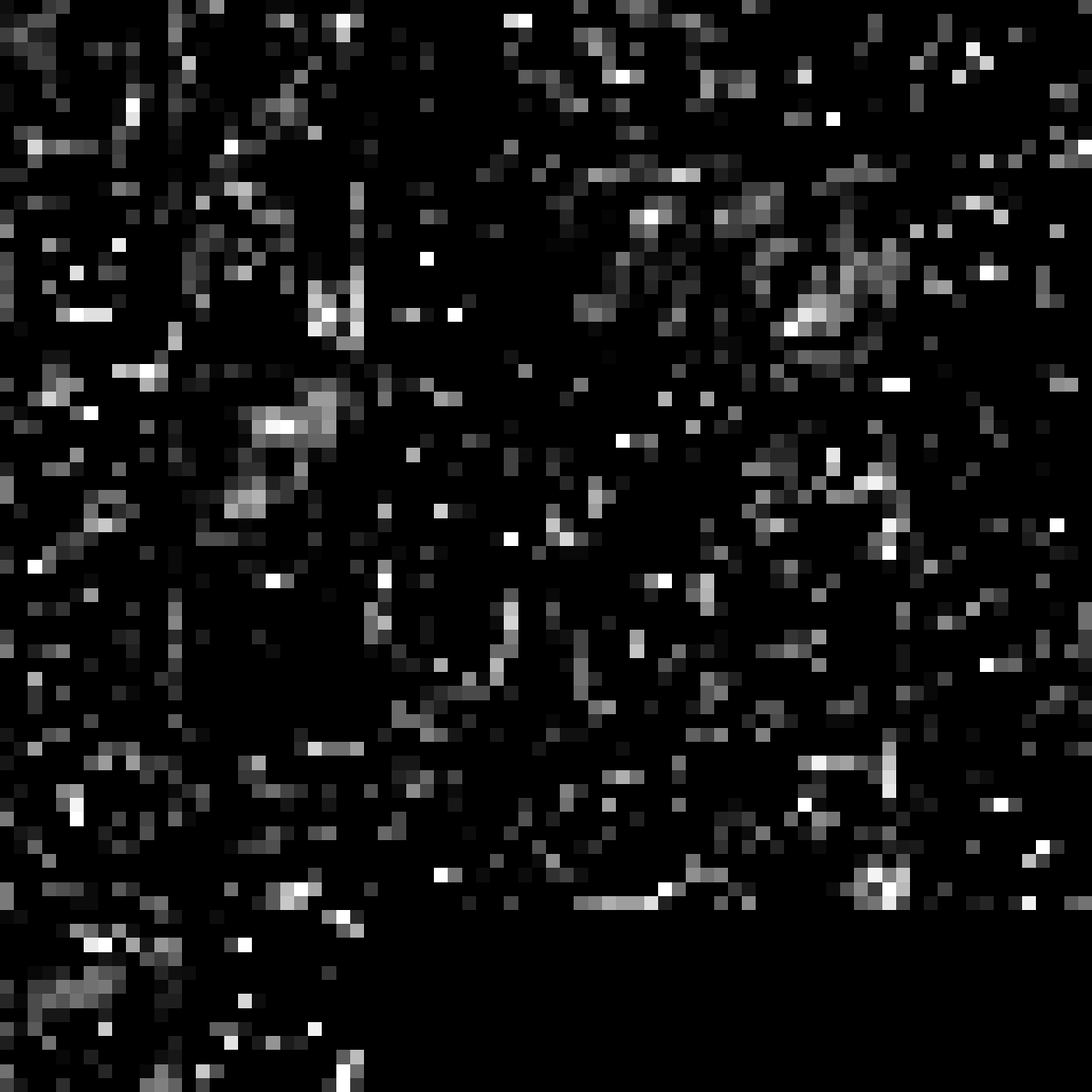} & \\[1.35cm] % & linux = 2.5cm
			%\cincludegraphics[width=\exFeatureMap\textwidth]{coffee/pixelwise_icpr/map1.jpg} & \\[3cm]
			%\hline
			%%\textbf{2} &
			&
			\cincludegraphics[width=\exFeatureMap\textwidth]{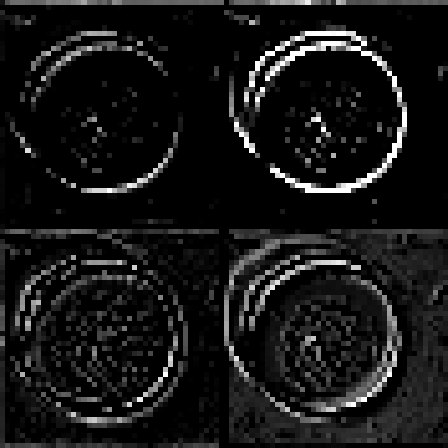} &
			\cincludegraphics[width=\exFeatureMap\textwidth]{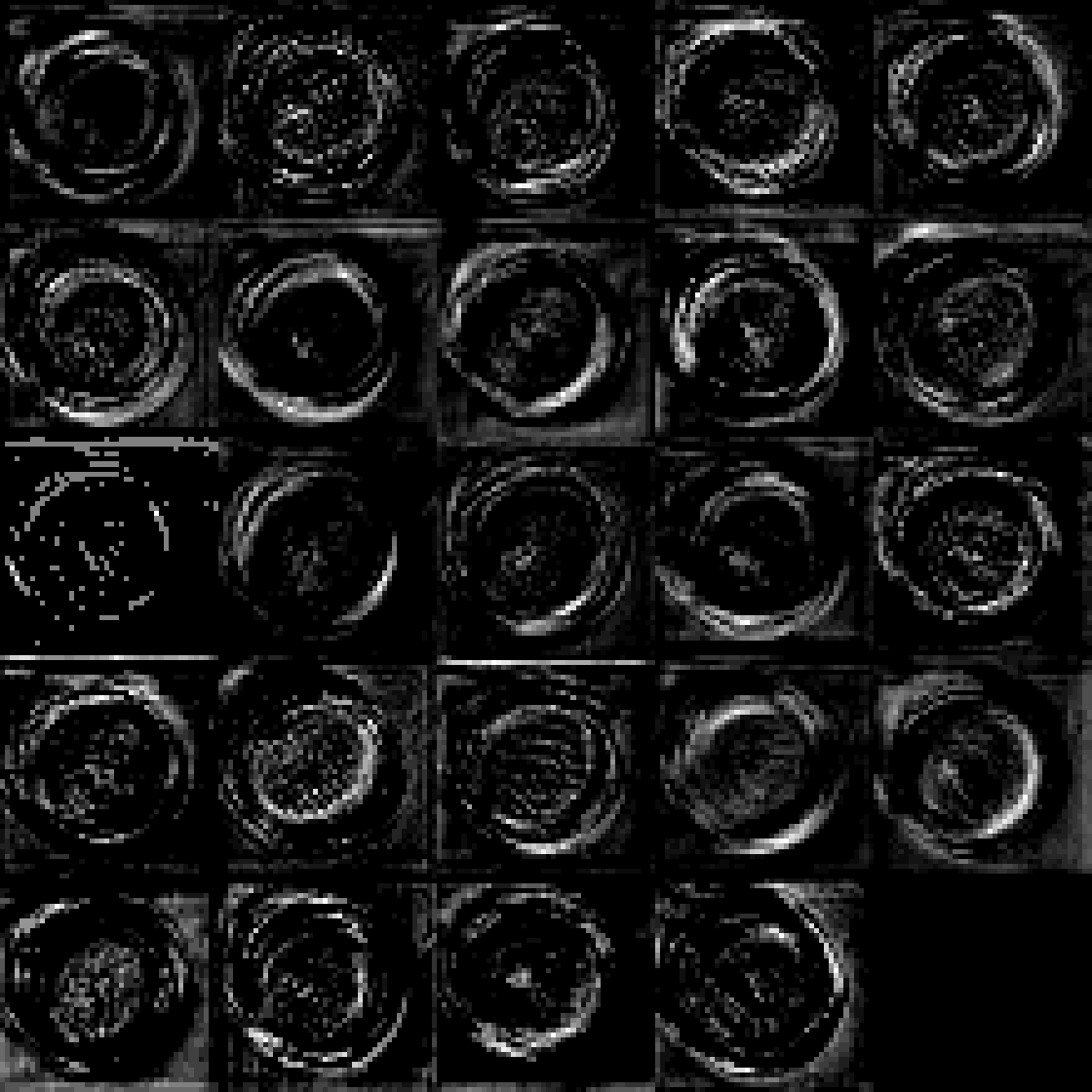} & 
			\cincludegraphics[width=\exFeatureMap\textwidth]{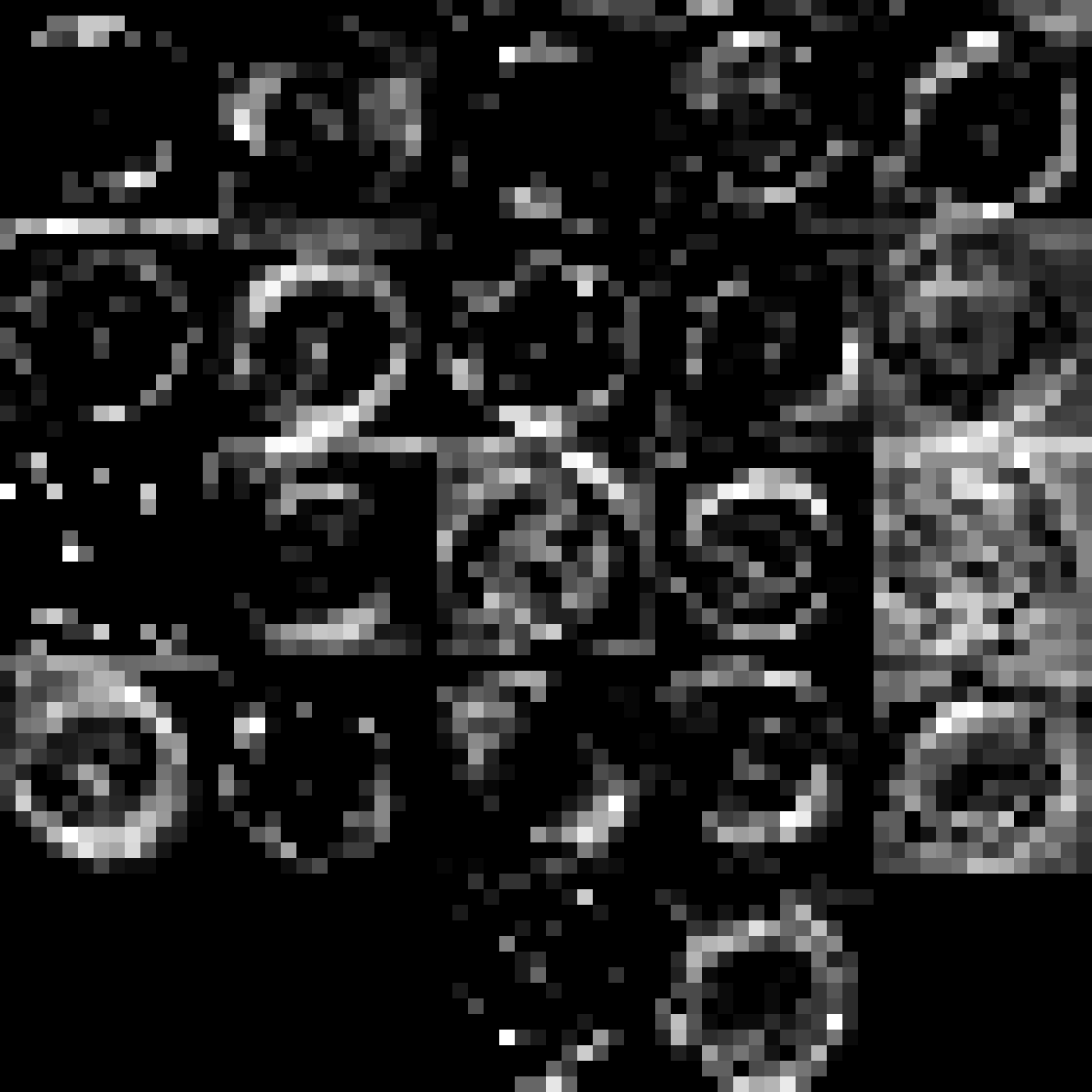} &
			\cincludegraphics[width=\exFeatureMap\textwidth]{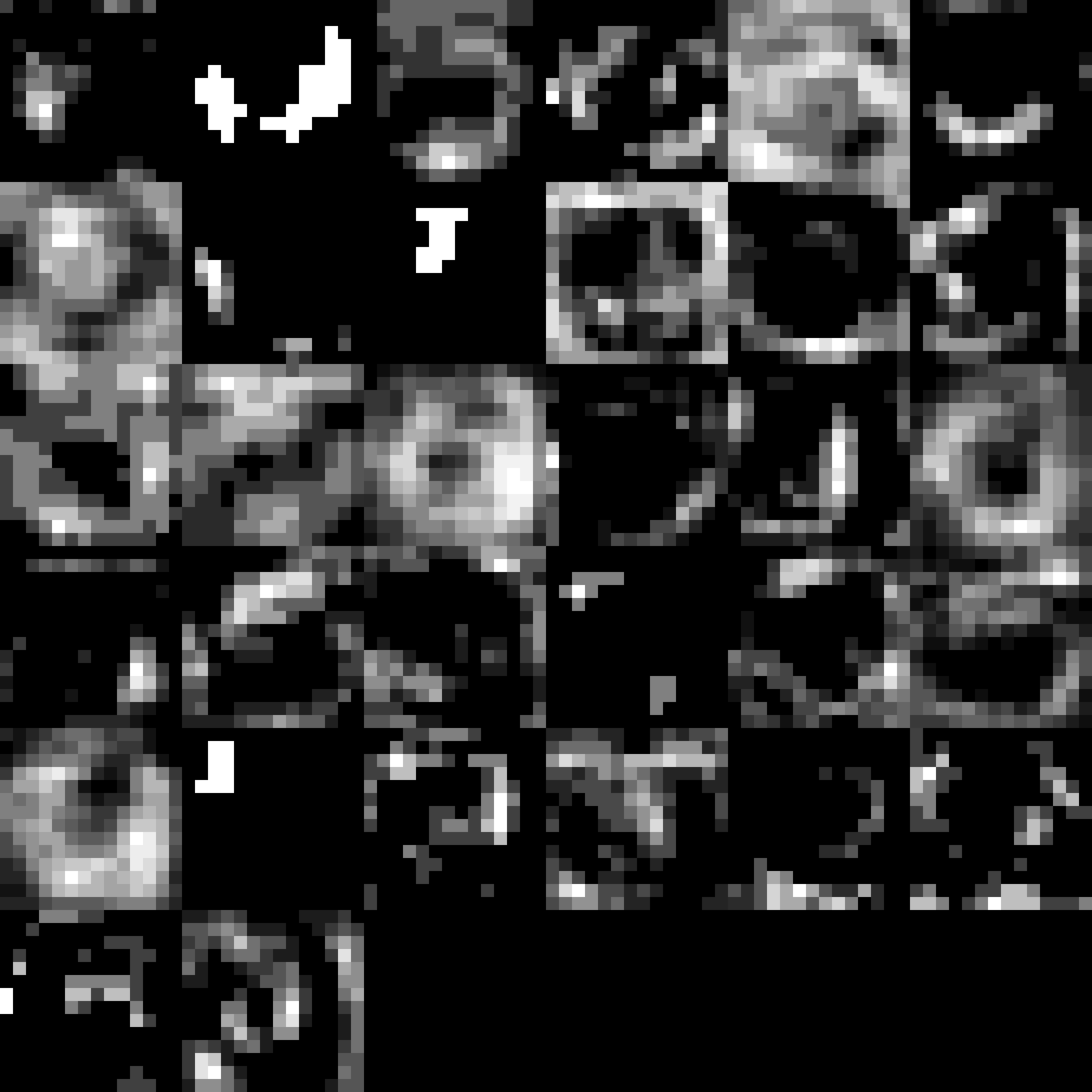} &
			\cincludegraphics[width=\exFeatureMap\textwidth]{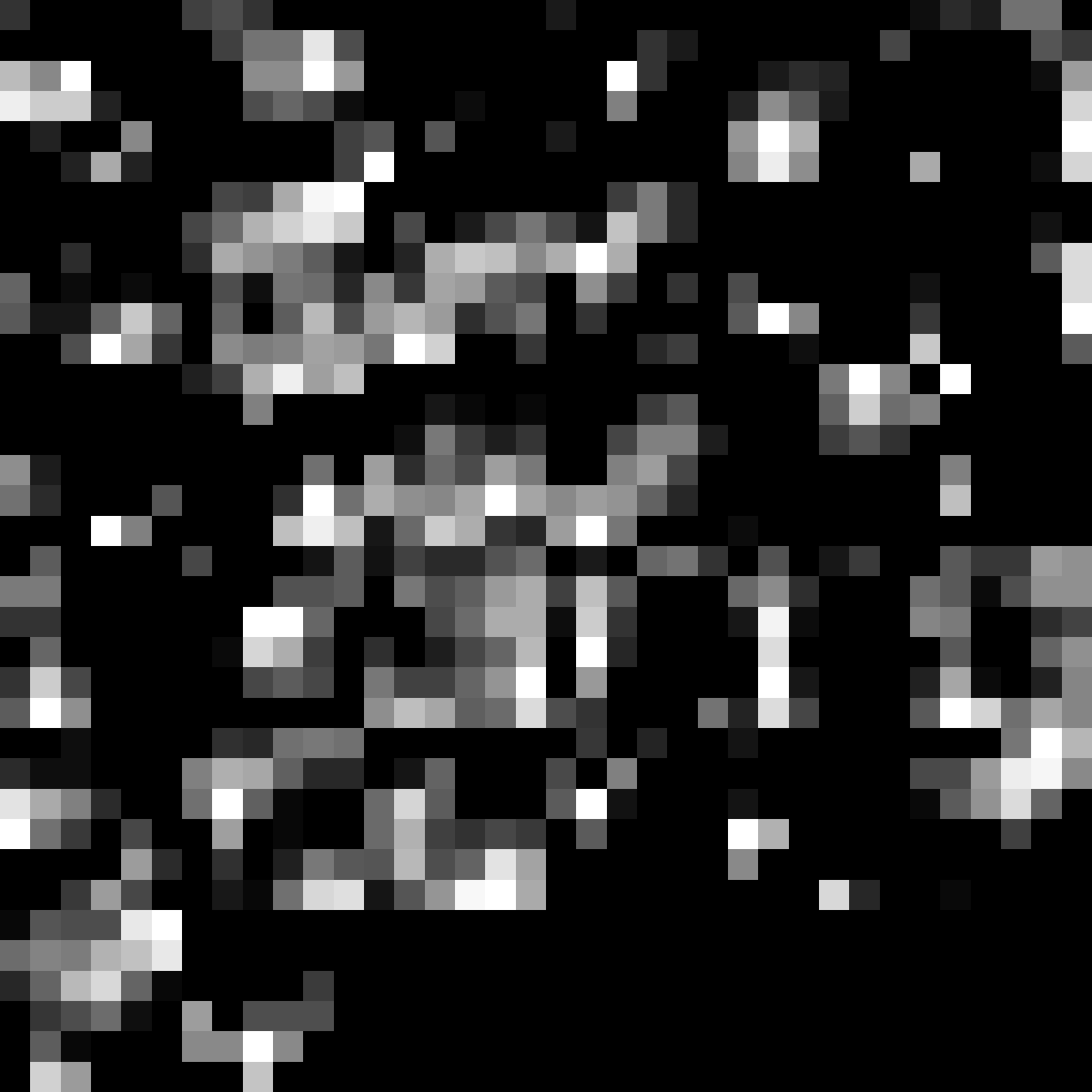} & \\[1.35cm] % &
			%\cincludegraphics[width=\exFeatureMap\textwidth]{coffee/pixelwise_icpr/map2.jpg} & \\[3cm]
			%\hline
			%%\textbf{3} &
			&
			\cincludegraphics[width=\exFeatureMap\textwidth]{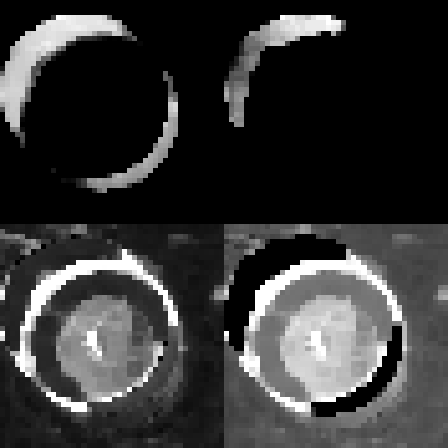} &
			\cincludegraphics[width=\exFeatureMap\textwidth]{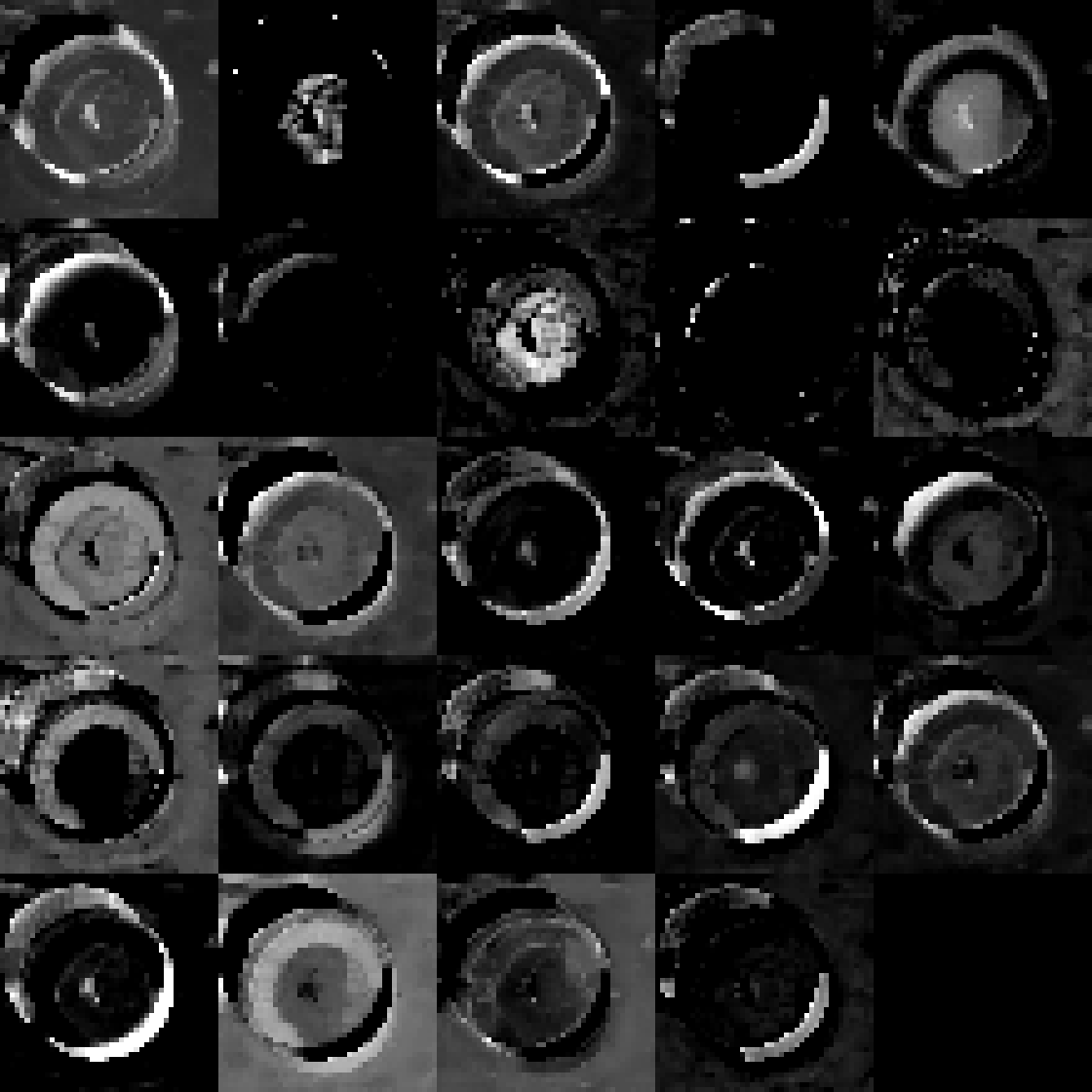} & 
			\cincludegraphics[width=\exFeatureMap\textwidth]{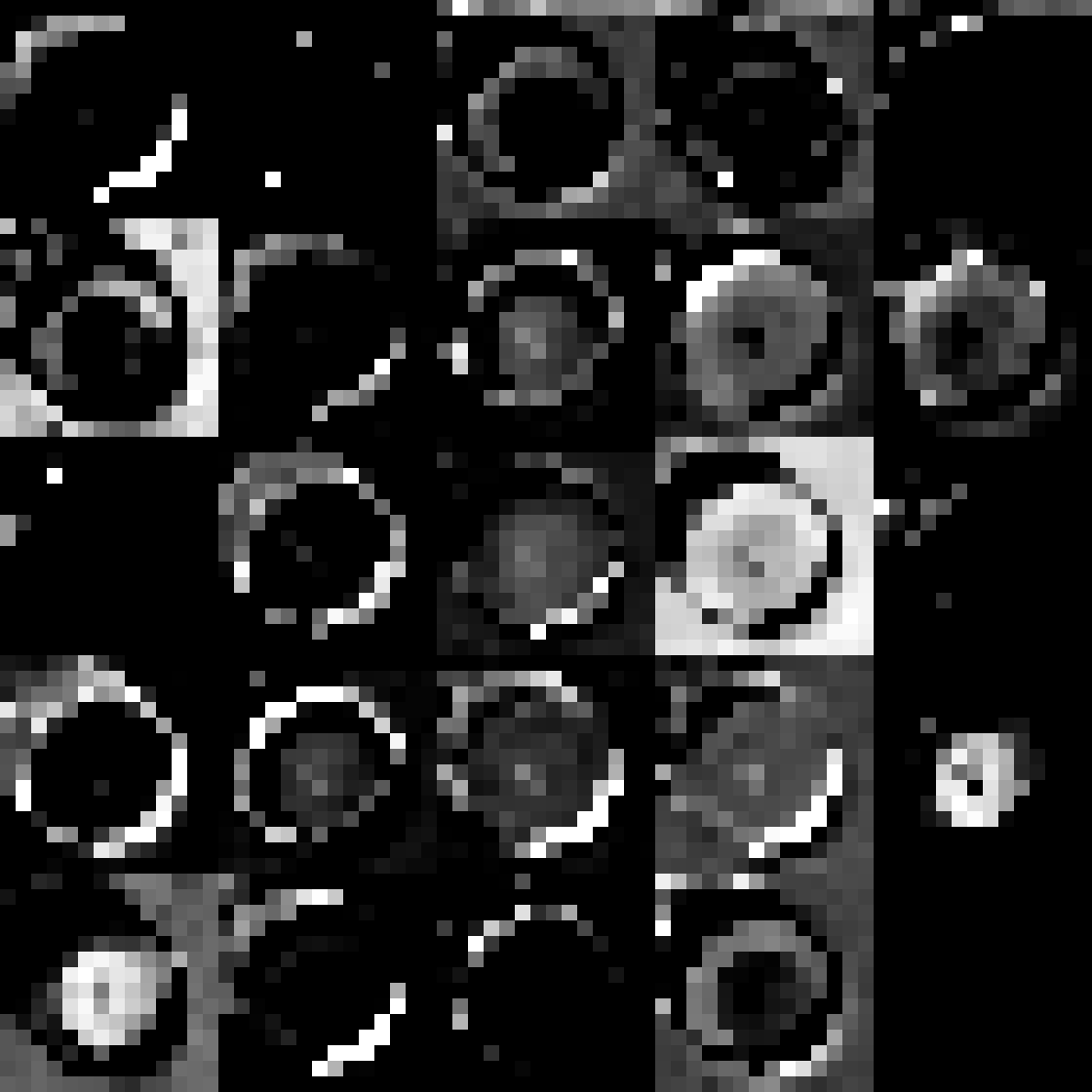} &
			\cincludegraphics[width=\exFeatureMap\textwidth]{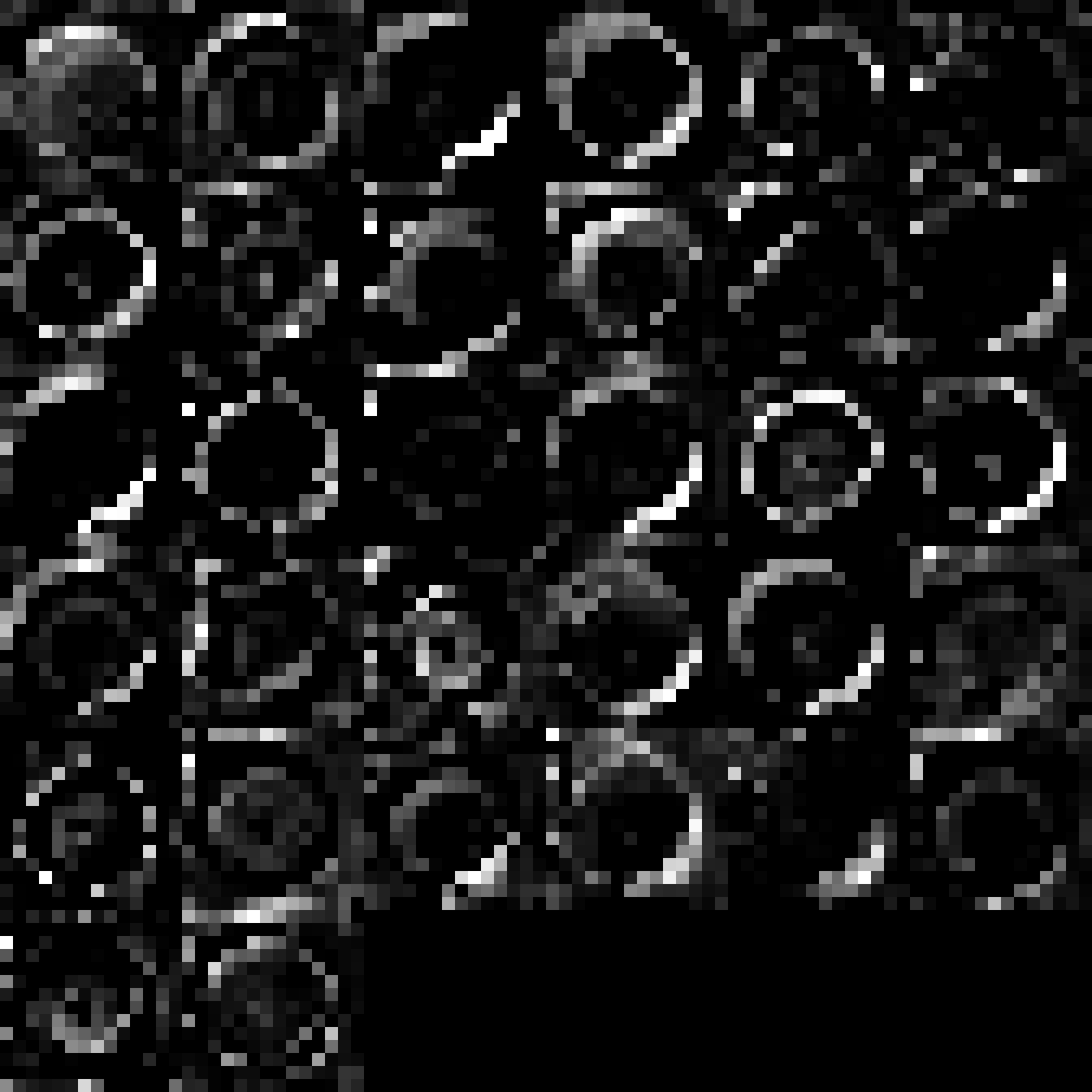} &
			\cincludegraphics[width=\exFeatureMap\textwidth]{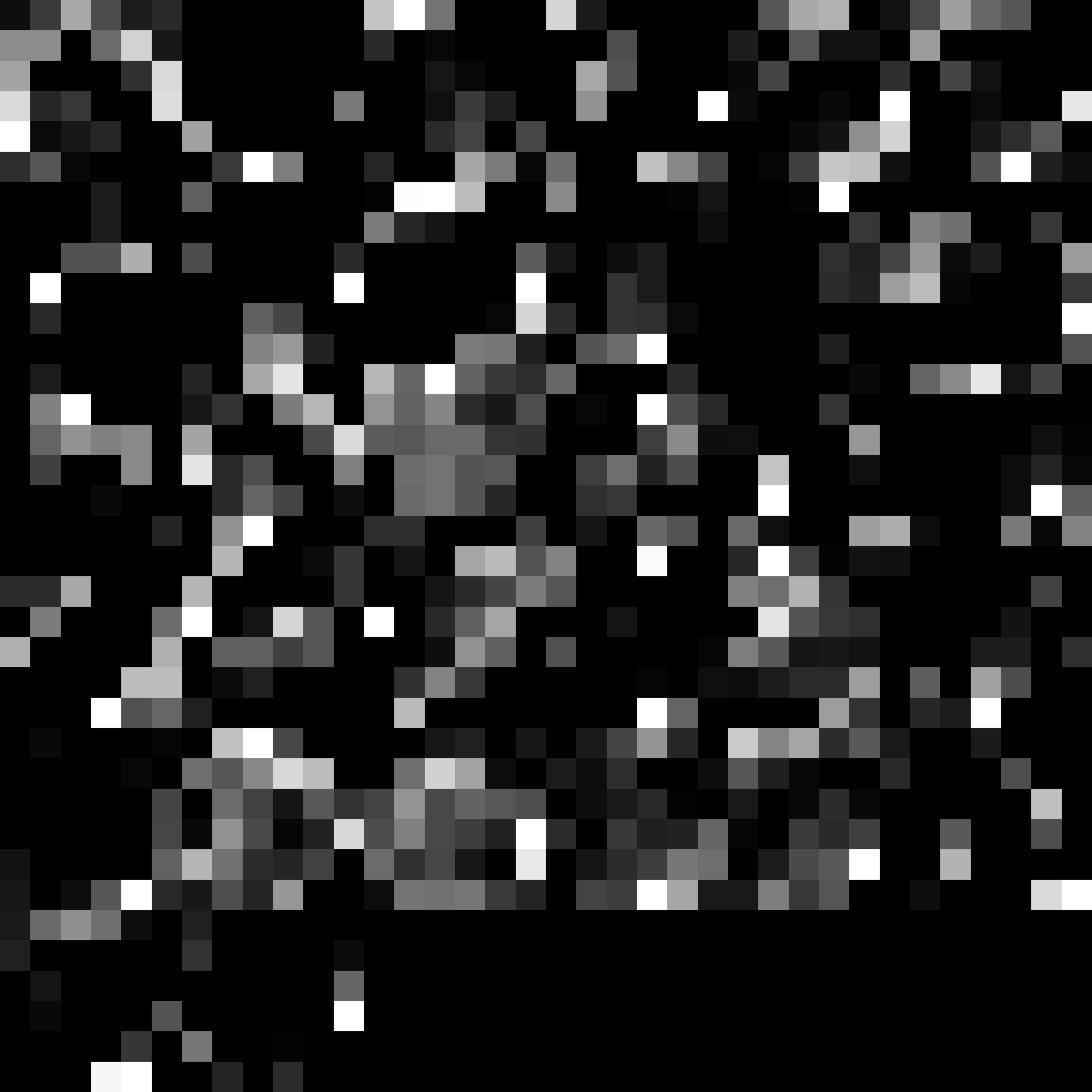} & \\[1.35cm] % &
			\cincludegraphics[width=\exFeatureMap\textwidth]{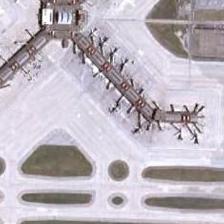} &
			\cincludegraphics[width=\exFeatureMap\textwidth]{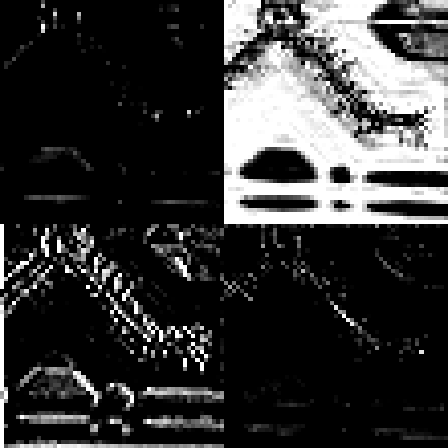} &
			\cincludegraphics[width=\exFeatureMap\textwidth]{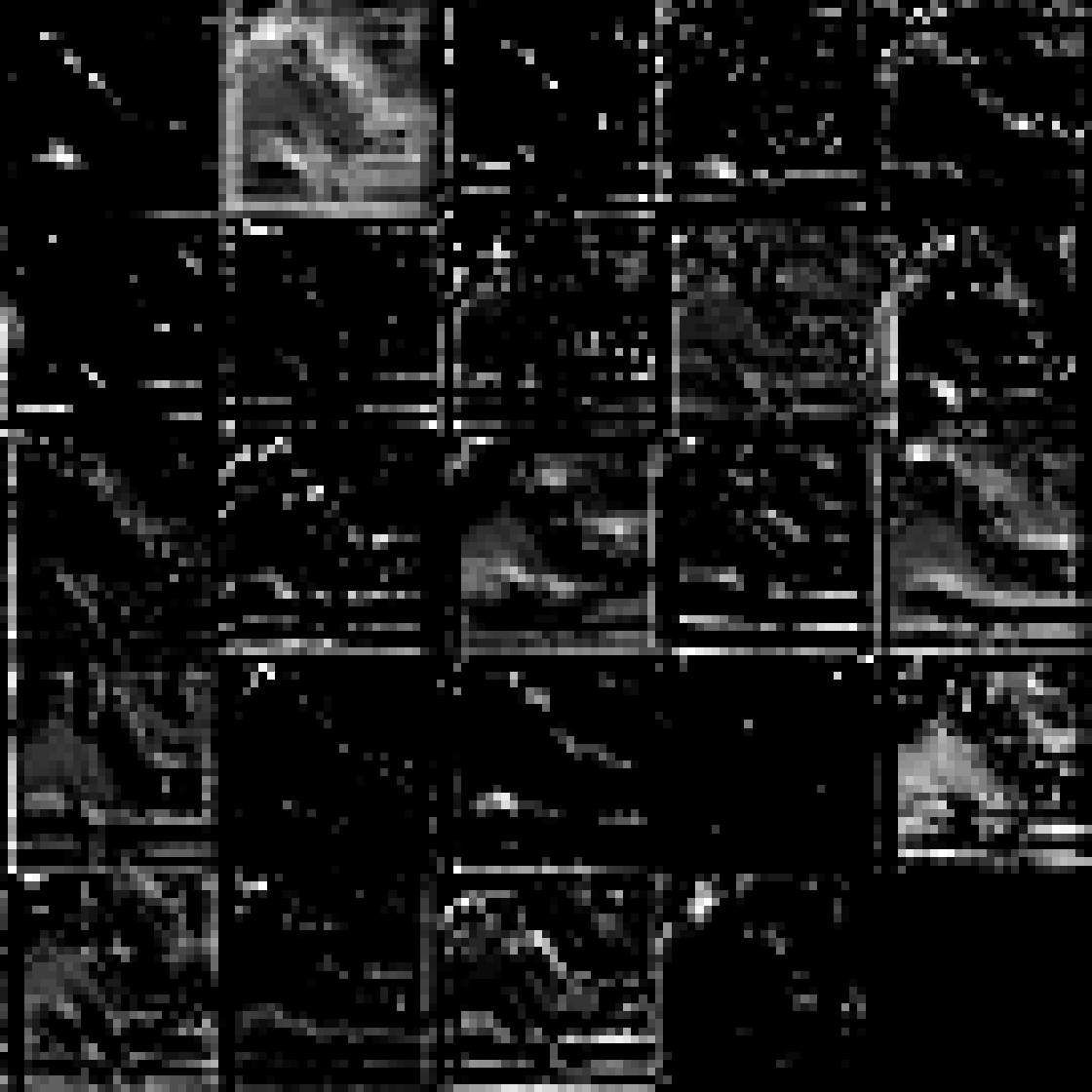} &
			\cincludegraphics[width=\exFeatureMap\textwidth]{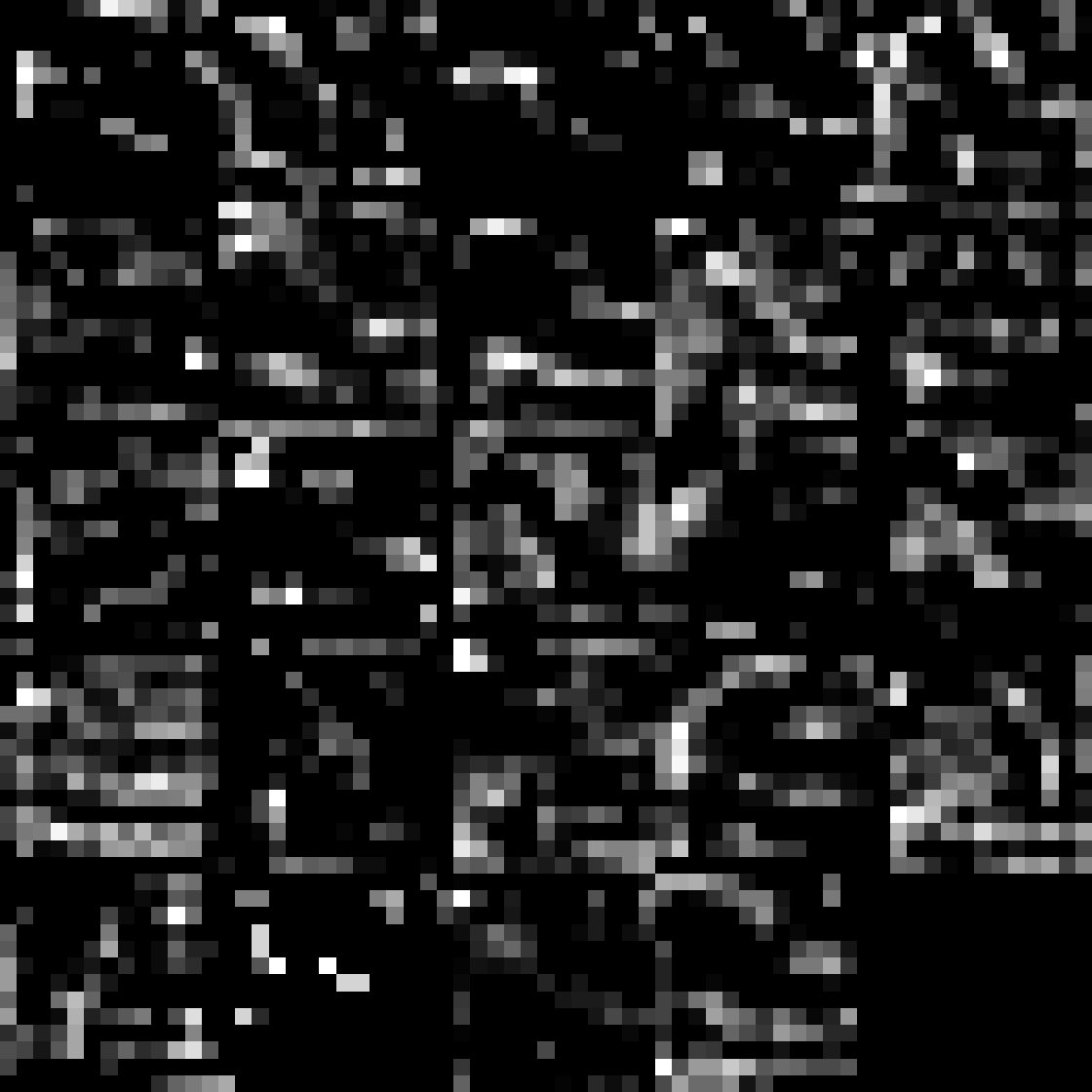} &
			\cincludegraphics[width=\exFeatureMap\textwidth]{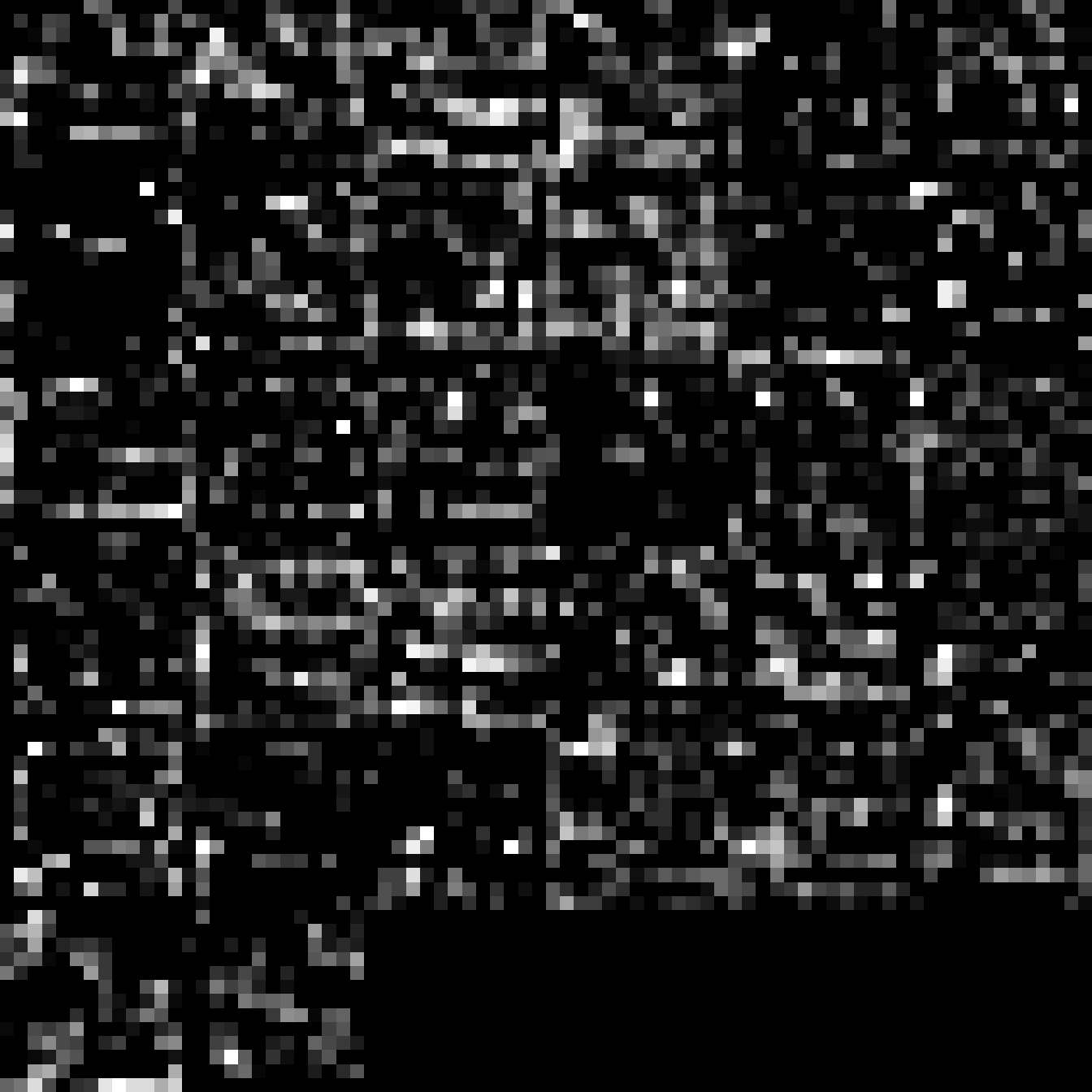} &
			\cincludegraphics[width=\exFeatureMap\textwidth]{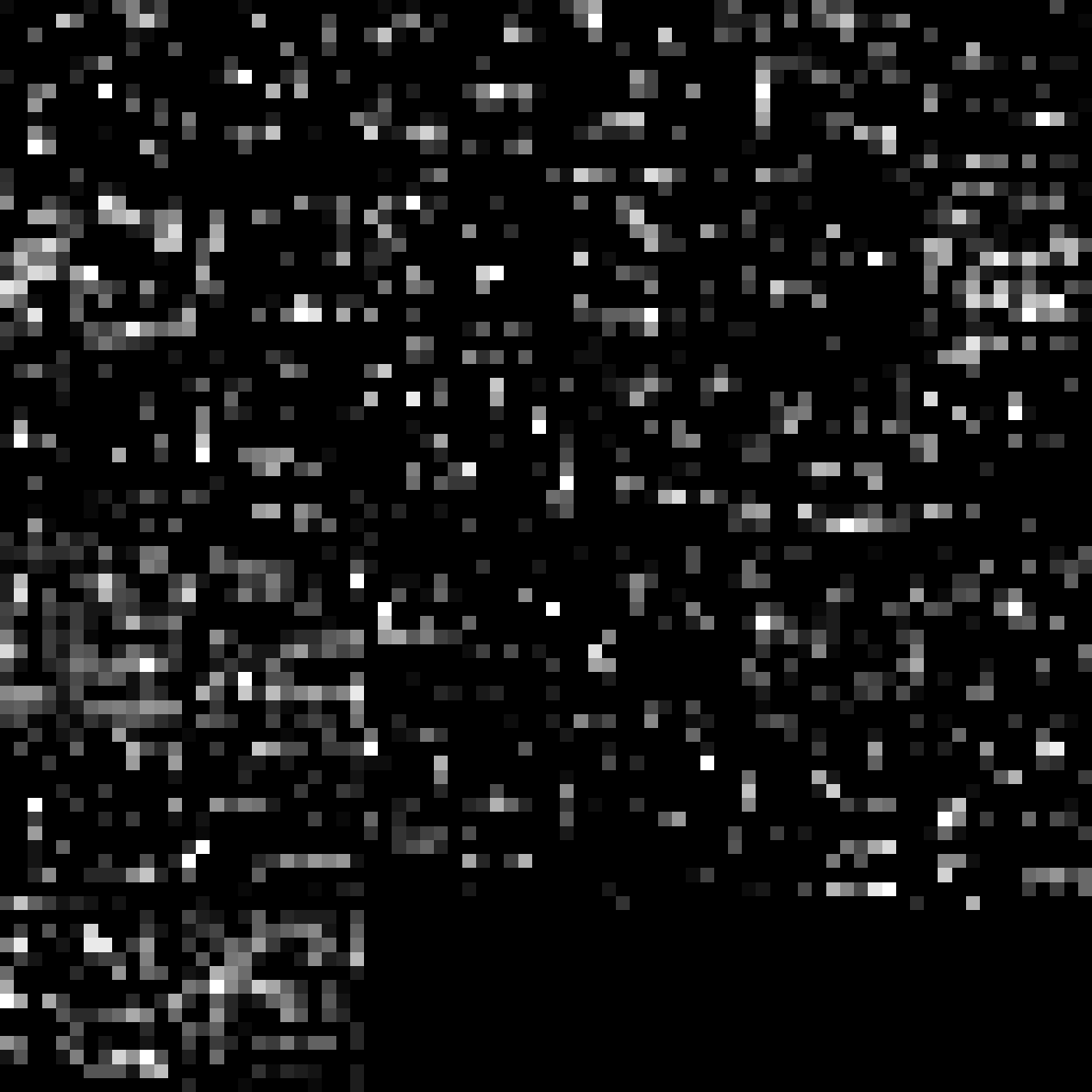} & \\[1.35cm] % &
			%\cincludegraphics[width=\exFeatureMap\textwidth]{coffee/pixelwise_icpr/map1.jpg} & \\[3cm]
			%\hline
			%%\textbf{2} &
			&
			\cincludegraphics[width=\exFeatureMap\textwidth]{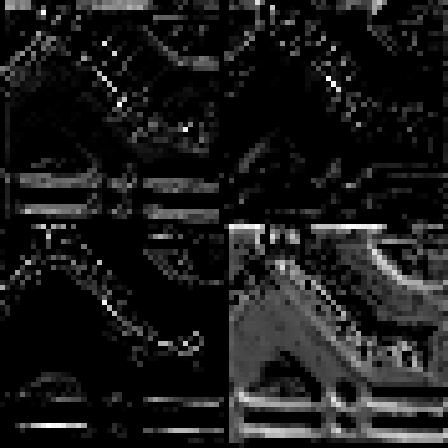} &
			\cincludegraphics[width=\exFeatureMap\textwidth]{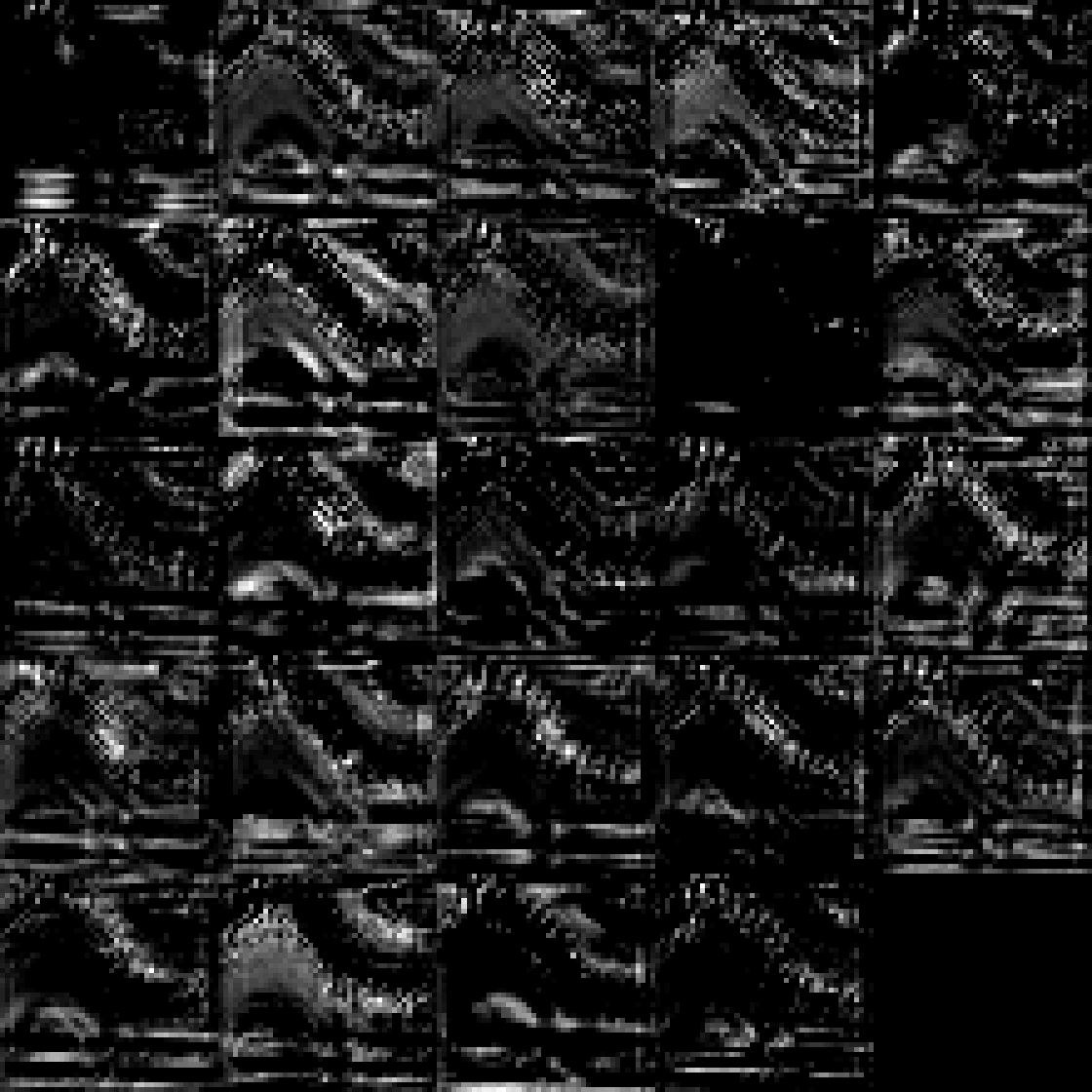} & 
			\cincludegraphics[width=\exFeatureMap\textwidth]{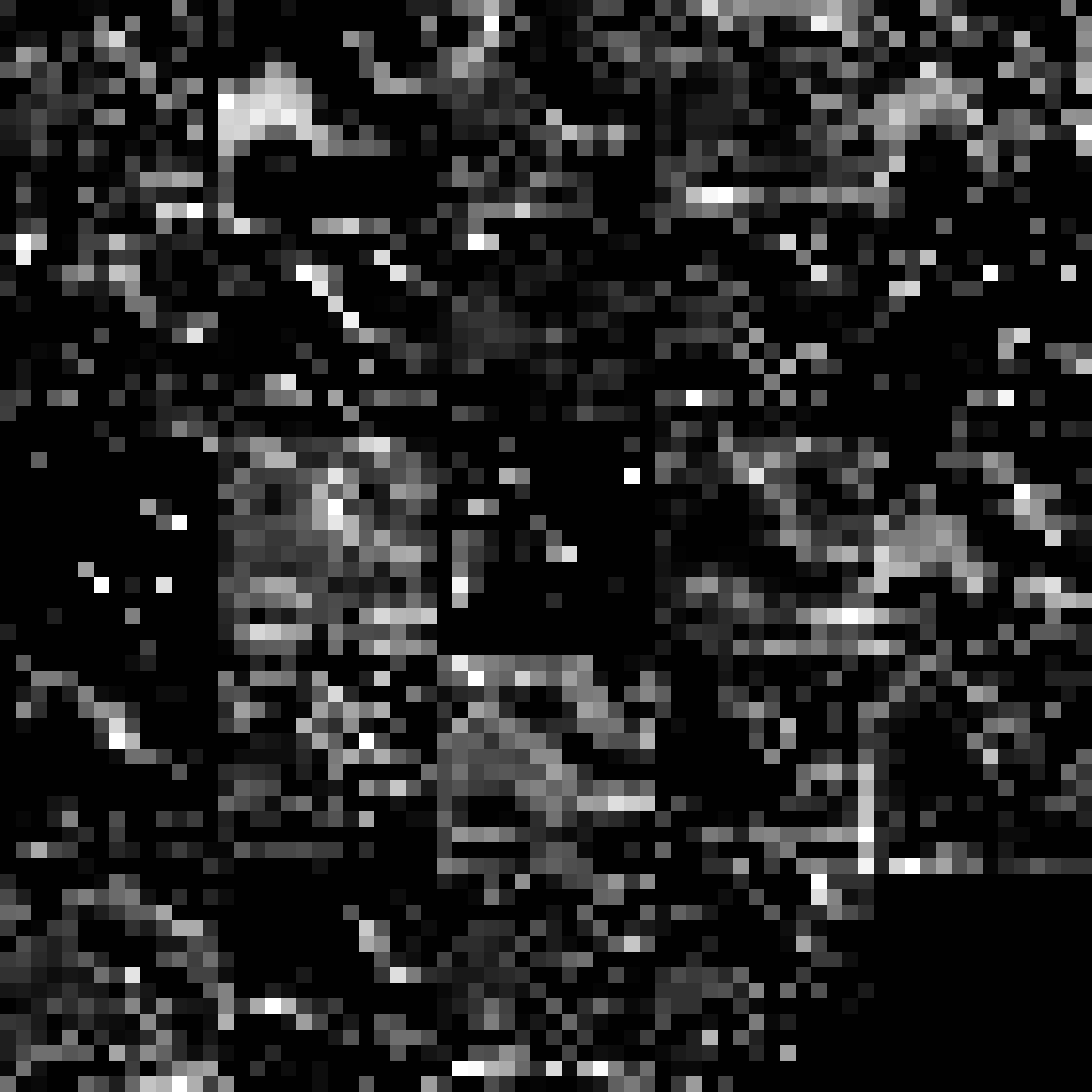} &
			\cincludegraphics[width=\exFeatureMap\textwidth]{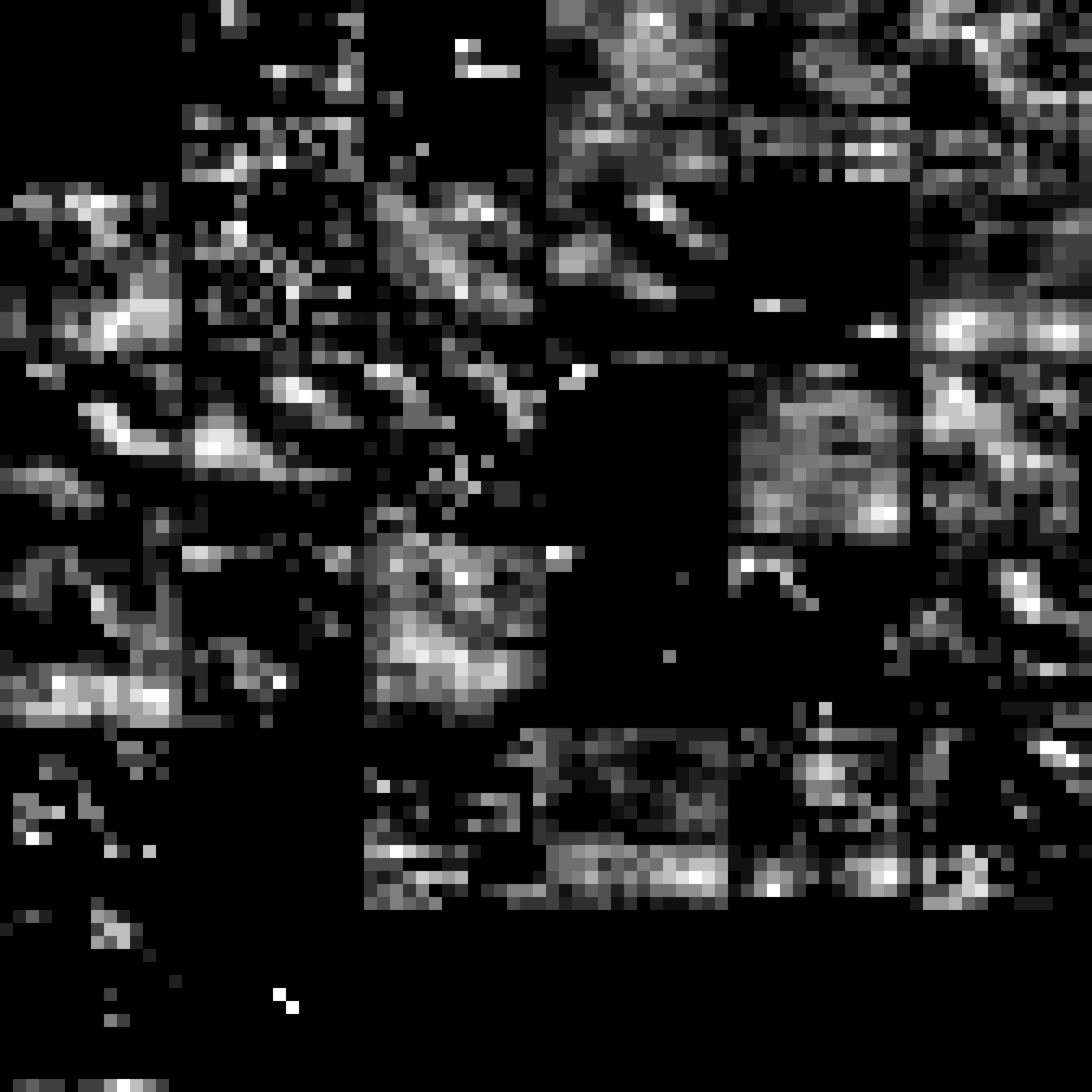} &
			\cincludegraphics[width=\exFeatureMap\textwidth]{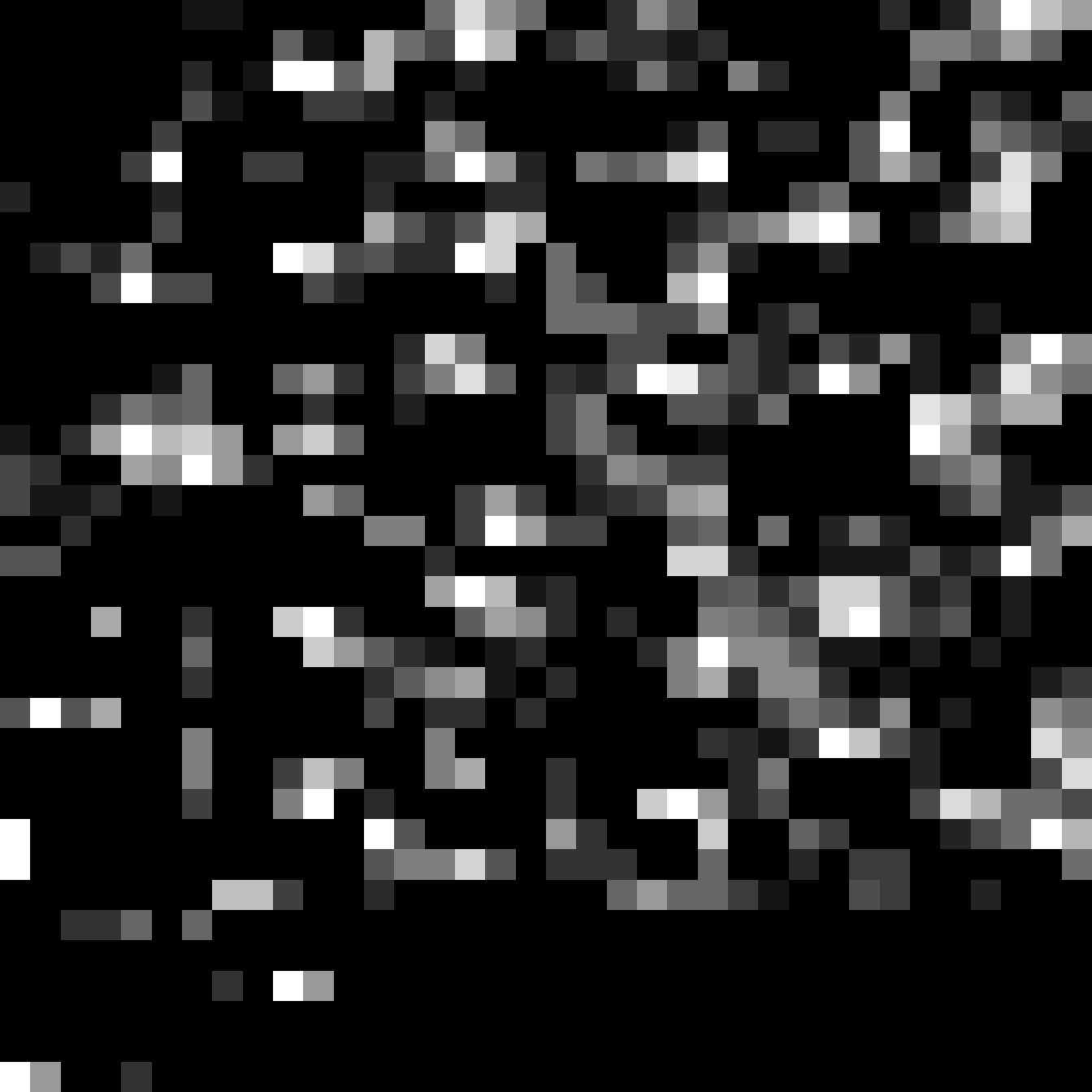} & \\[1.35cm] % &
			%\cincludegraphics[width=\exFeatureMap\textwidth]{coffee/pixelwise_icpr/map2.jpg} & \\[3cm]
			%\hline
			%%\textbf{3} &
			&
			\cincludegraphics[width=\exFeatureMap\textwidth]{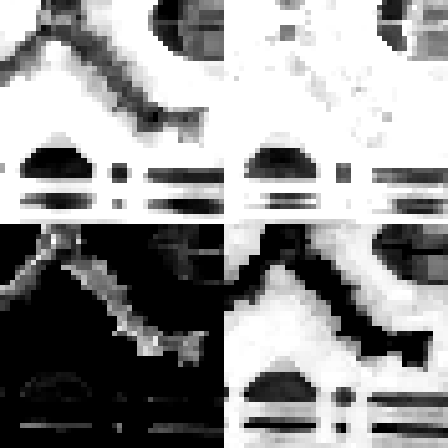} &
			\cincludegraphics[width=\exFeatureMap\textwidth]{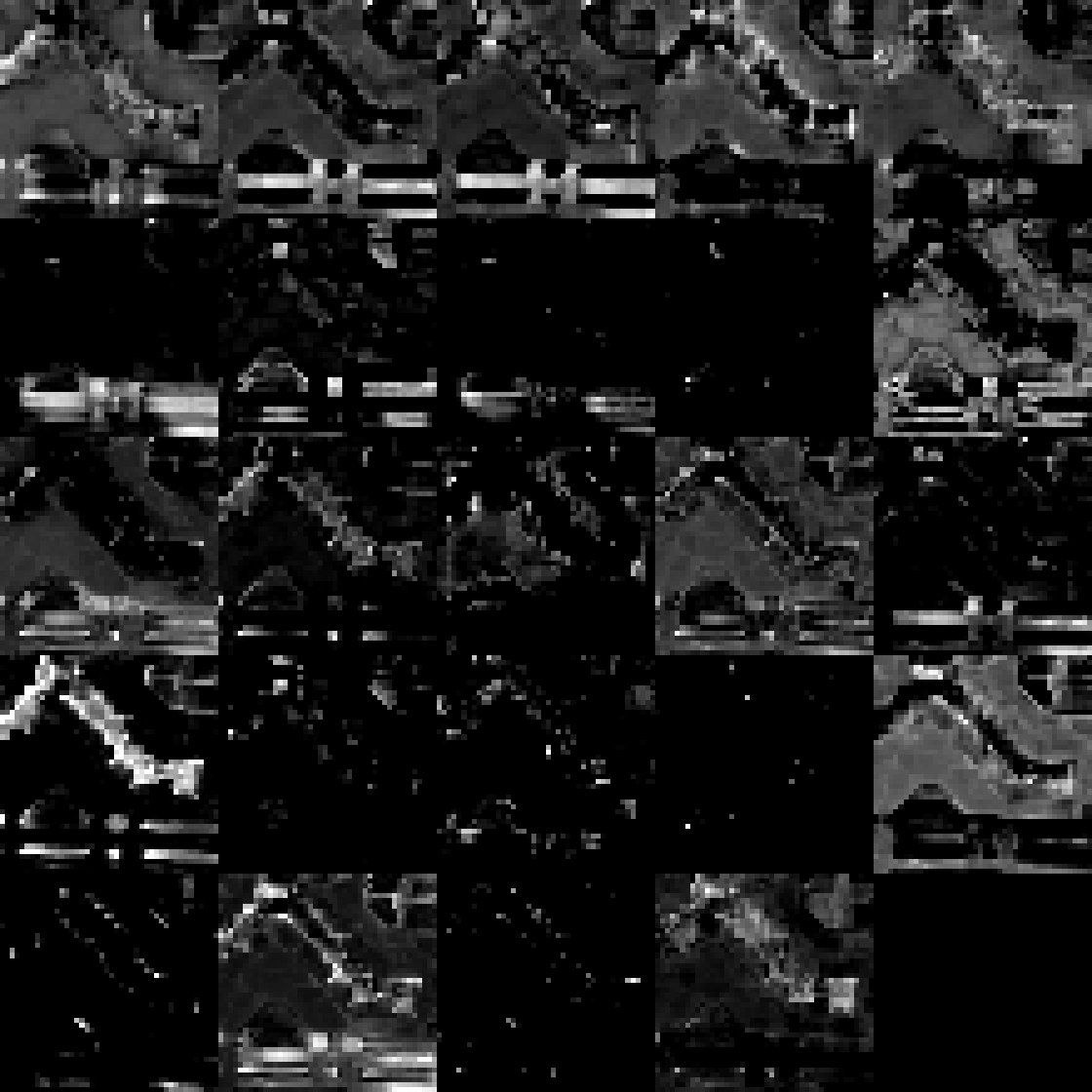} & 
			\cincludegraphics[width=\exFeatureMap\textwidth]{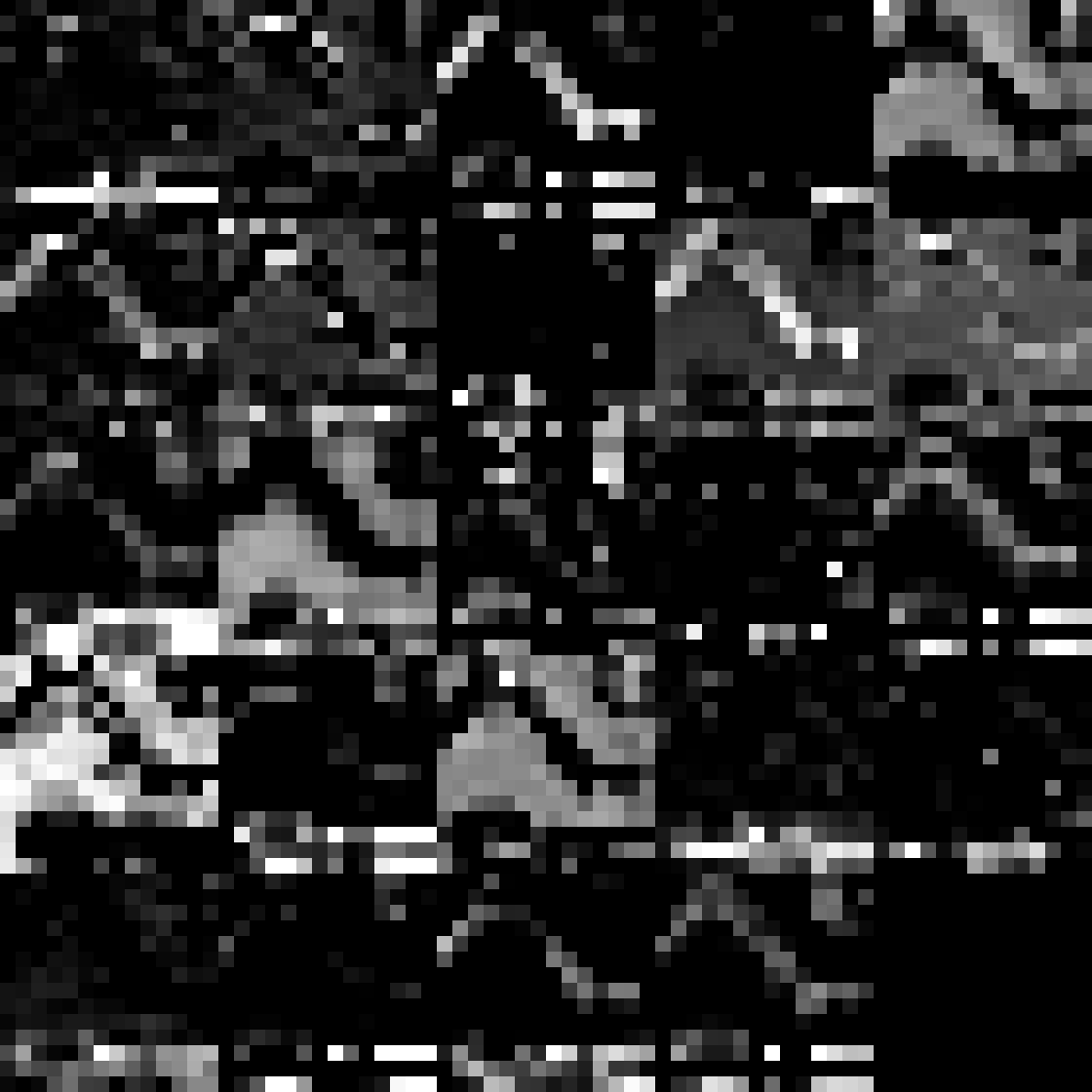} &
			\cincludegraphics[width=\exFeatureMap\textwidth]{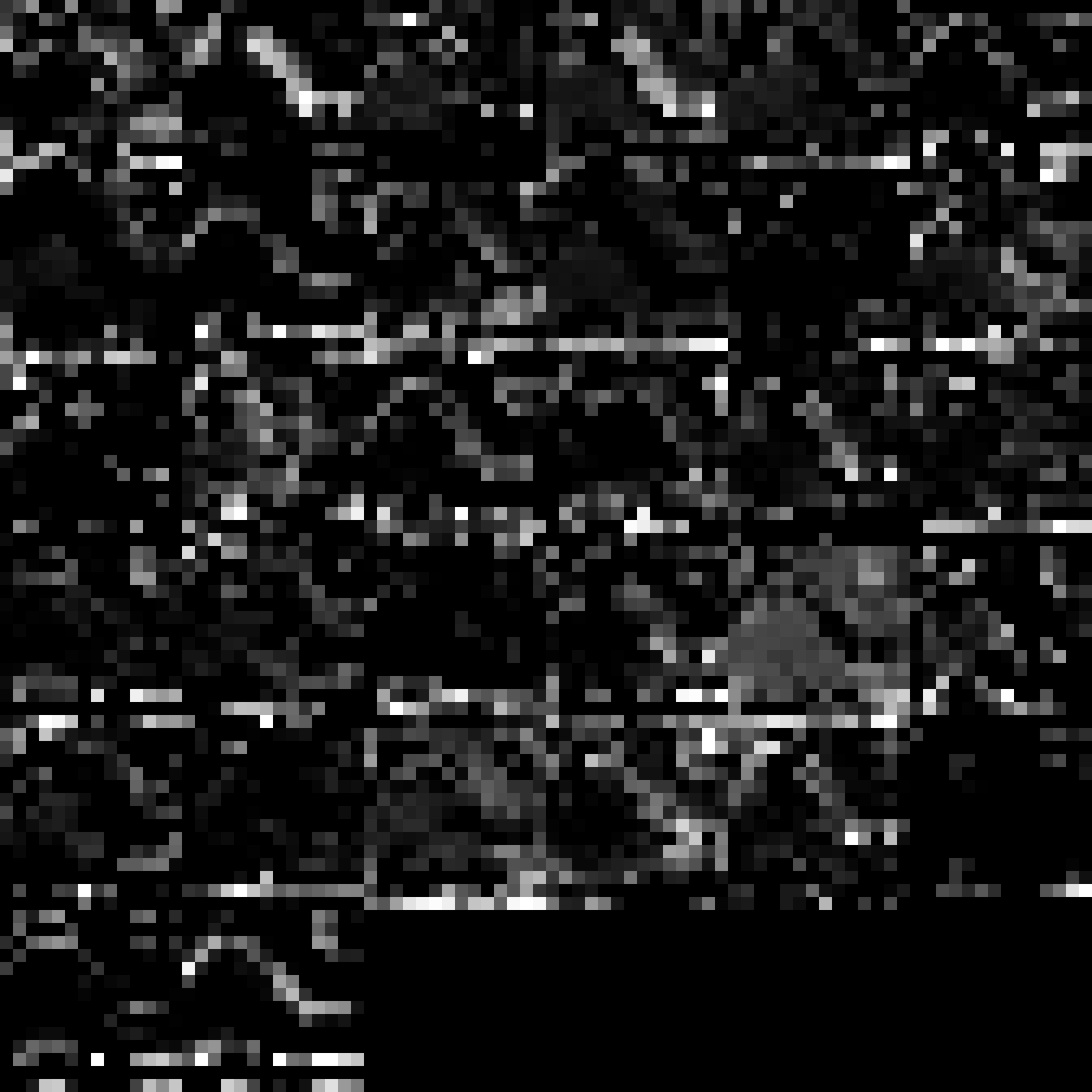} &
			\cincludegraphics[width=\exFeatureMap\textwidth]{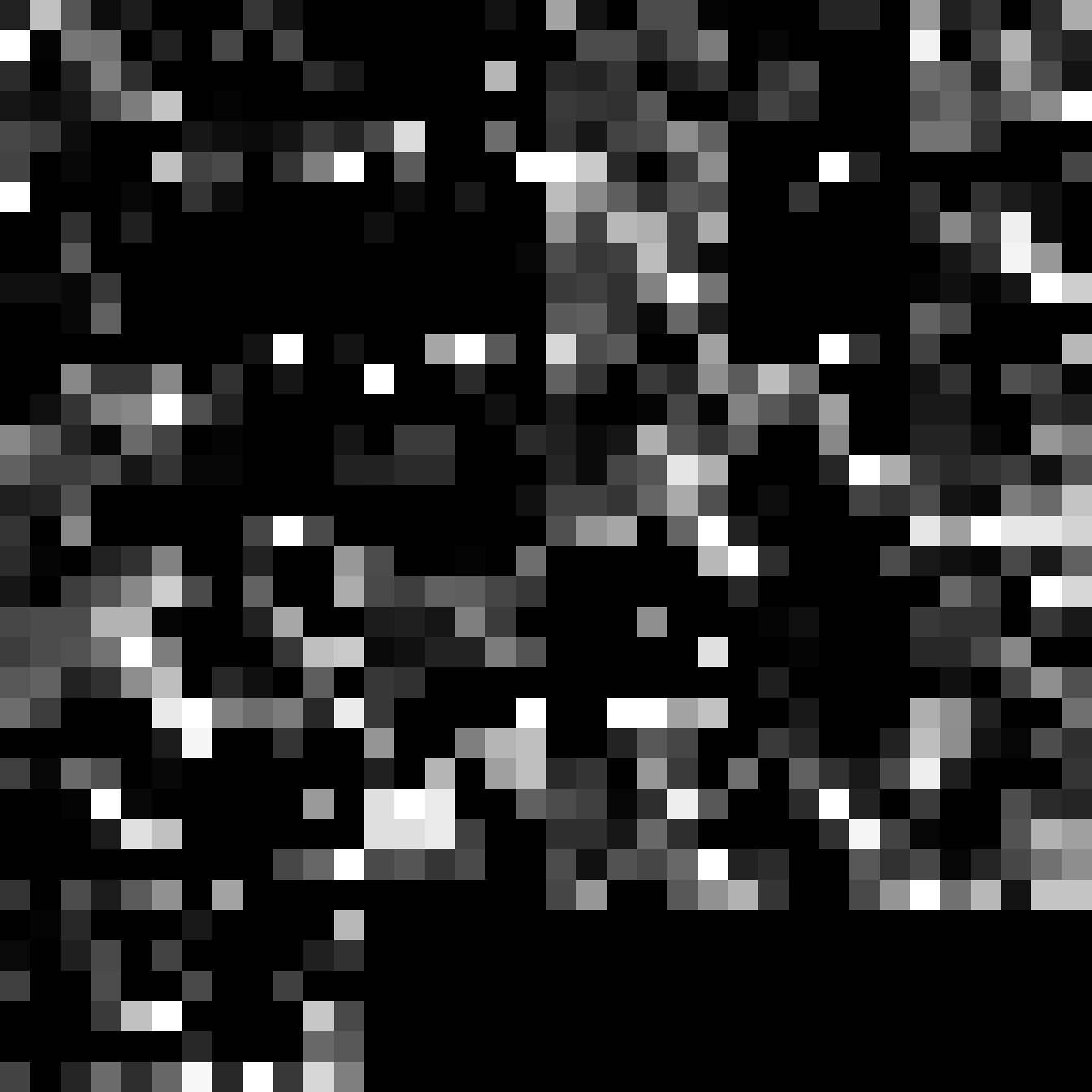}
		\end{tabular}
	\end{center}
	\captionof{figure}{Input images and some produced (upsampled) feature maps extracted from all layers of the networks for the UCMerced Land-use (top example) and RS19 Datasets (bottom example).
		Again, for simplicity, only the feature maps of the ConvNet (first row), Depth-ConvNet (second row), and DeepMorphNet (last row) models, without Selective Kernels~\cite{li2019selective}, were reported.
		% For each input image: the first row presents features from the ConvNet network, the second row presents the feature maps learned by the Depth-ConvNet architecture, and the last row presents the features of the proposed morphological network.
	}
	\label{fig:ucmerced_feature_maps}
\end{table*}

To better evaluate the proposed morphological network, a convergence analysis of the architectures in the UCMerced Land-use dataset is presented in Figure~\ref{convergence_ucmerced}.
For simplicity, only the networks that do not use selective kernels~\cite{li2019selective} were reported.
%Figure~\ref{convergence_ucmerced} presents the convergence of the morphological network as well as the baselines in the UCMerced Land-use dataset.
Note that the proposed model is slower to converge when compared to the other networks.
A reason for that is the large number of trainable parameters of DeepMorphNets, as presented in Table~\ref{tab:ucmerced_results}.
% As presented in Table~\ref{tab:ucmerced_results}, the DeepMorphNets have more parameters and, therefore, are more complex to train.
However, given enough training time, all networks converge very similarly, which confirms that the proposed DeepMorphNets are able to effectively learn interesting SEs and converge to a suitable solution.

\newcommand{\convergenceFigSize}{0.24}
\begin{figure}[h]
	\centering
	\subfloat[UCMerced Land-use Dataset]{
		\includegraphics[width=\convergenceFigSize\textwidth]{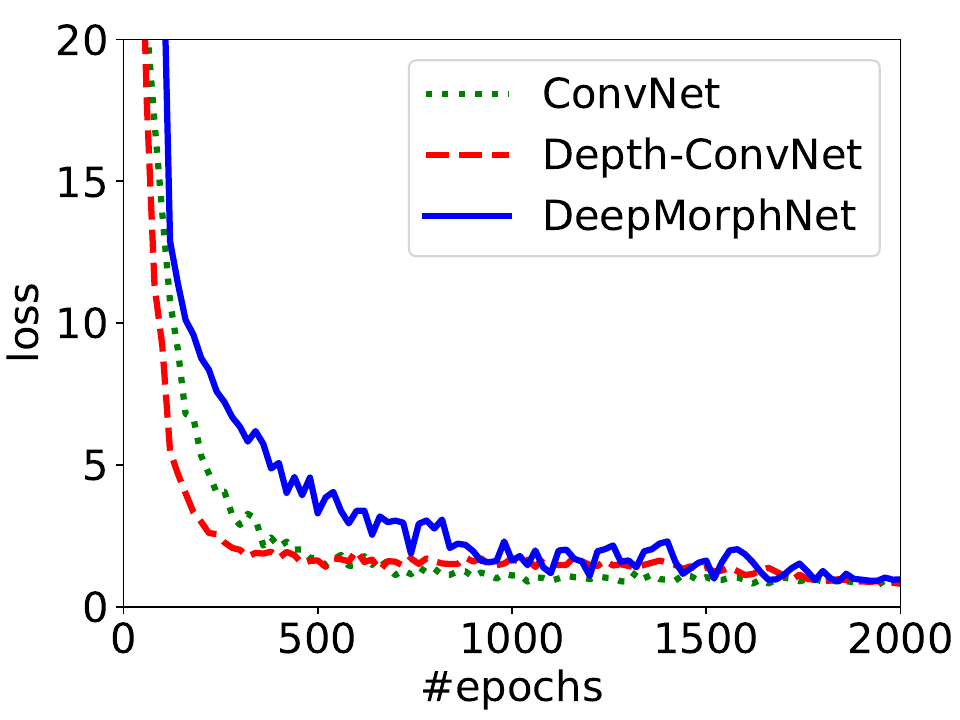}
		\label{convergence_ucmerced}
	}
	%\hspace{1mm}
	\subfloat[WHU-RS19 Dataset]{
		\includegraphics[width=\convergenceFigSize\textwidth]{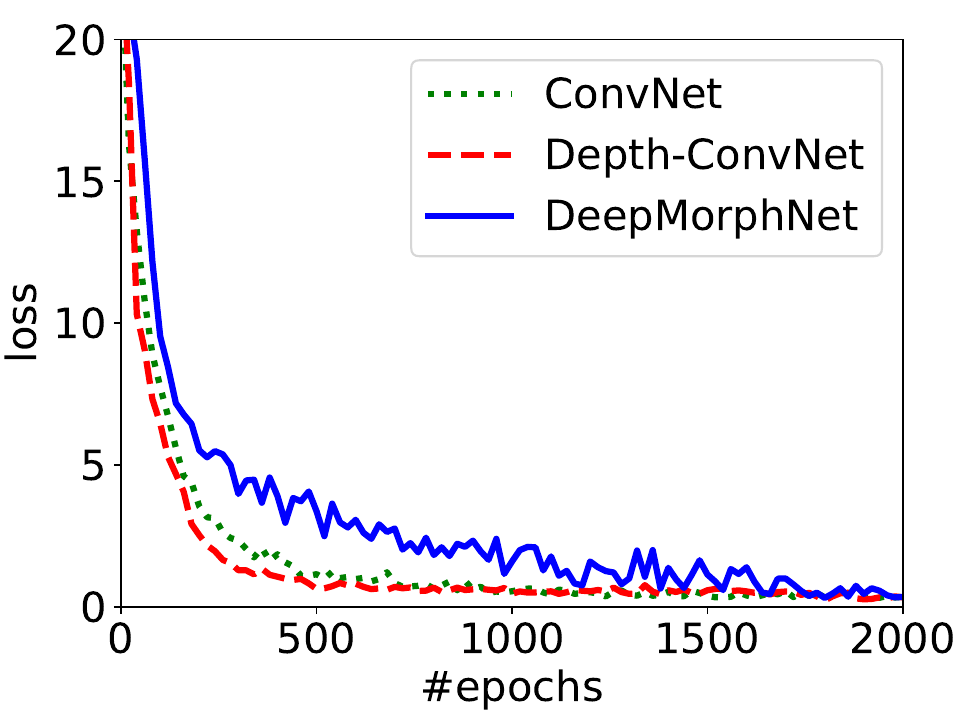}
		\label{convergence_rs19}
	}
	\caption{
		Convergence (fold 1) of the architectures for both datasets.
		%Convergence of the proposed morphological networks and the baselines for both datasets.
		Note that, for simplicity, only the networks without Selective Kernels~\cite{li2019selective} were reported. 
		% Furthermore, only fold 1 is reported.
	}
	\label{fig:convergence}
\end{figure}

\subsubsection{WHU-RS19 Dataset} \label{subsubsec:rs19_res}

The second part of Table~\ref{tab:ucmerced_results} presents the results related to the WHU-RS19 dataset.
Again, as expected, all architectures outperformed the lower bound result, generated by the Static SEs.
%Considering the LeNet-based networks, the best result, among the baselines, was produced by the LeNet architecture~\cite{lecun1998gradient}.
%The best result, among the baselines, was produced by the ConvNet-based architectures.
%However, as for the UCMerced Land-use dataset, the DeepMorphNets yielded similar results (when compared to their respective baseline).
\changes{
	As for the previous dataset, all experimented techniques, including the DeepMorphNets and the networks with selective kernels~\cite{li2019selective}, achieved very similar outcomes, outperforming the MorphoN~\cite{mondal2020image} and DMNN~\cite{franchi2020deep} approaches.
}
%The proposed DeepMorphLeNet generated competitive results when compared to this baseline, corroborating with previous conclusions about the ability of the proposed technique to optimize the morphological filters and extract important features.
%
%As for the UCMerced Land-use dataset, the exact same conclusions can be drawn from the AlexNet-based networks.
%In this case, the best baseline was the AlexNet-based network, that produces 64.38$\pm$2.93\% of average accuracy.
%However, the proposed DeepMorphAlexNet produced similar results when compared to this baseline (68.20$\pm$2.75\% of average accuracy).
These results reaffirm the previous conclusions related to the ability of the morphological networks to capture interesting features.
Figure~\ref{fig:ucmerced_feature_maps} presents the same comparison, as before, between the feature maps extracted from DeepMorphNets and ConvNets.
Again, it is remarkable the difference between the features extracted by the distinct architectures, which corroborates with previous analysis and results.
%Overall, the DeepMorphNet is capable of learning specific and distinct features when compared to the ConvNets.

%In order to grasp the difference between (the learning step of) DeepMorphNets and ConvNets, we performed a comparison between the feature maps of such networks, which is presented in Figure~\ref{fig:ucmerced_feature_maps}.
%As can be observed, there is a clear difference between the characteristics learned by the distinct networks.
%In general, the DeepMorphNet is able to preserve different features when compared to the ConvNets, which corroborates with our initial analysis.

As for the previous dataset, a convergence analysis of the architectures in the WHU-RS19 dataset is presented in Figure~\ref{convergence_rs19}.
Again, only the networks that do not exploit selective kernels~\cite{li2019selective} were reported.
As before, the proposed model is slower (due to the number of trainable parameters), but able to converge if enough time is provided for the training.

\subsection{Pixel Classification Datasets} \label{subsec:pixel_classification}

For the pixel classification datasets, all networks and baselines were assessed using the same configuration, i.e., batch size, learning rate, weight decay, momentum, and number of epochs of 10, 0.01, 0.0005, 0.9, and 1,000 respectively.

\subsubsection{Pavia Centre Dataset} \label{subsubsec:pavia_centre}

Results for the Pavia Centre dataset are reported in (the first part of) Table~\ref{tab:pixelclass_results}.
%In order to allow a visual comparison, prediction maps for this dataset are presented in Figure~\ref{fig:pavia_results}.
As can be seen through this table, the proposed DeepMorphNet outperformed the ConvNet model by almost 5 percentage points of average accuracy (88.28\% versus 83.73\%, respectively).
This outcome shows, once again, that the proposed technique is capable of effectively learning relevant characteristics of the data. 
% with previous analysis on the feature learning ability of the proposed technique
% Furthermore, these results show that, in some scenarios (such as this one), the features learned by the DeepMorphNet can make this networks to.

\begin{table}[H]
	\caption{Results, in terms of accuracy, of the proposed method and the baselines for the pixel classification datasets.}
	\label{tab:pixelclass_results}
	\centering
	\resizebox{\columnwidth}{!}{
		\begin{tabular}{@{}clrrr@{}}
			\toprule
			\multicolumn{1}{c}{\textbf{Dataset}} & \multicolumn{1}{c}{\textbf{Method}} & 
			\begin{tabular}[c]{@{}c@{}}\textbf{Average}\\ \textbf{Accuracy (\%)}\end{tabular} &
			\begin{tabular}[c]{@{}c@{}}\textbf{Number of}\\ \textbf{Parameters} \\ \textbf{(millions)}\end{tabular} & 
			\begin{tabular}[c]{@{}c@{}}\textbf{Training}\\ \textbf{Time}\\ \textbf{(hours per fold)}\end{tabular} \\
			\midrule
			\multirow{2}{*}{\begin{tabular}[c]{@{}c@{}}\textbf{Pavia}\\ \textbf{Centre}\end{tabular}}
			& \textbf{ConvNet}~\cite{krizhevsky2012imagenet}      				& 83.73 &  6.50  & 0.5   \\
			& \textbf{DeepMorphNet (ours)}   				       				& 88.28 & 10.50  & 3.5   \\ 
			\midrule
			\midrule
			\multirow{2}{*}{\begin{tabular}[c]{@{}c@{}}\textbf{Pavia}\\ \textbf{University}\end{tabular}} 
			& \textbf{ConvNet}~\cite{krizhevsky2012imagenet}					& 82.44 &   6.50  & 0.5   \\
			& \textbf{DeepMorphNet (ours)}   				        			& 86.52 &  10.50  & 3.5   \\
			\bottomrule
		\end{tabular}
	}
\end{table}

%This gain can be seen qualitatively in Figure~\ref{fig:pavia_results}, which presents the false-color image, the ground-truth, and the prediction maps generated by the proposed approach and the baseline.
In order to allow a visual comparison, Figure~\ref{fig:pavia_results} presents the false-color image, the ground-truth, and the prediction maps generated by the proposed approach and the baseline.
From this image, it is possible to observe that the DeepMorphNet produced a more consistent prediction map than the ConvNet baseline.
We argue that this is mainly due to the fact that the proposed method can learn distinct and relevant patterns of specific classes, such as Tiles and Bare Soil (cyan and dark blue colors, respectively), an outcome that can be better perceived in the normalized confusion matrices presented in Figure~\ref{fig:cms}.
% better than the ConvNet
%	Overall, this outcome allows us to conclude that the DeepMorphNet is able to learn some specific patterns that the ConvNet can not.

\newcommand{\exFigPaviaU}{0.2}
\begin{figure*}[h!]
	\centering
	\subfloat[False-color Image]{
		\includegraphics[height=\exFigPaviaU\textwidth]{c_false_color.png}
	}
	\subfloat[Ground-Truth]{
		\includegraphics[height=\exFigPaviaU\textwidth]{c_gt.png}
	}
	\subfloat[ConvNet Prediction]{
		\includegraphics[height=\exFigPaviaU\textwidth]{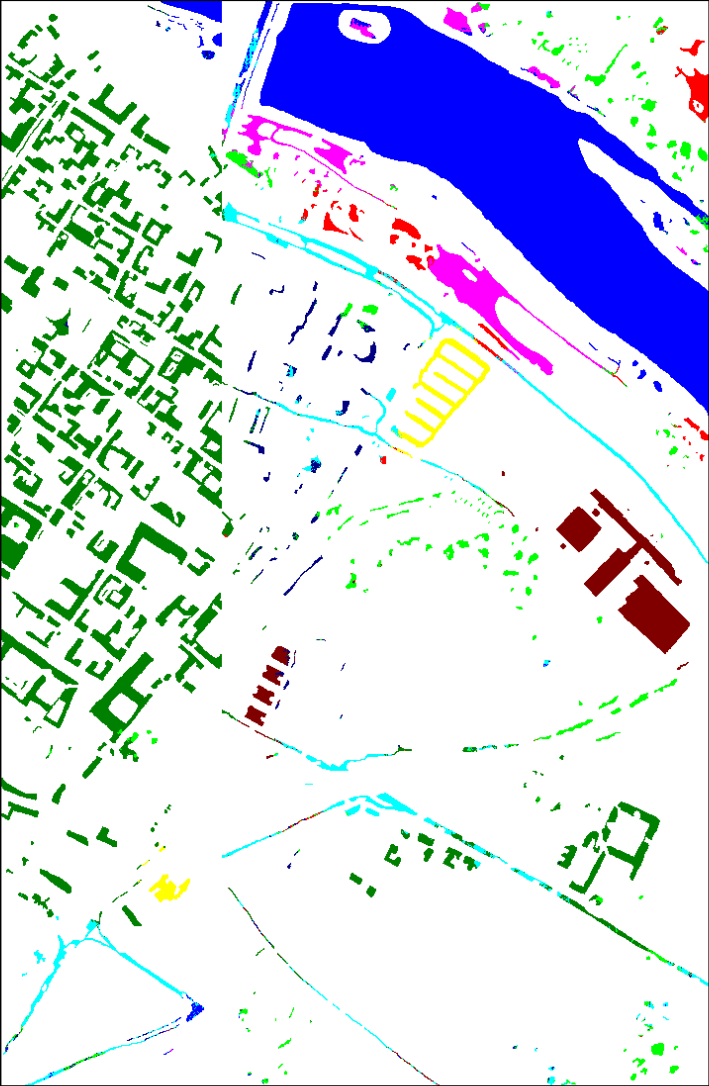}
	}
	\subfloat[DeepMorphNet Prediction]{
		\includegraphics[height=\exFigPaviaU\textwidth]{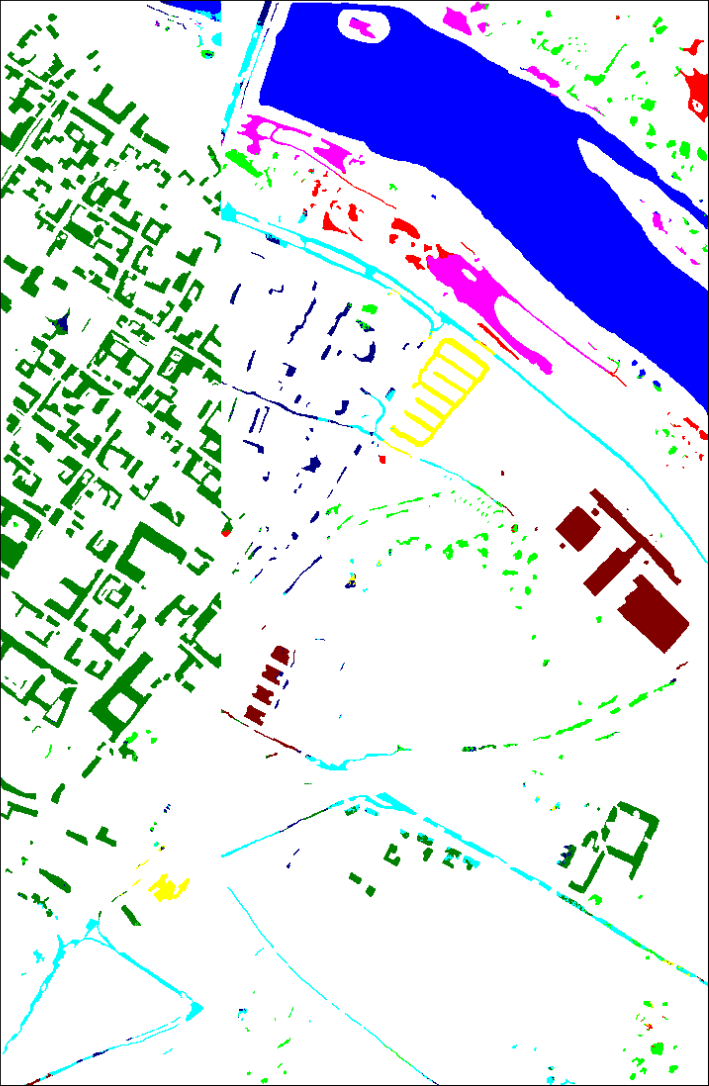}
	}
	\subfloat[False-color Image]{
		\includegraphics[height=\exFigPaviaU\textwidth]{u_false_color.png}
	}
	\subfloat[Ground-Truth]{
		\includegraphics[height=\exFigPaviaU\textwidth]{u_gt.png}
	}
	\subfloat[ConvNet Prediction]{
		\includegraphics[height=\exFigPaviaU\textwidth]{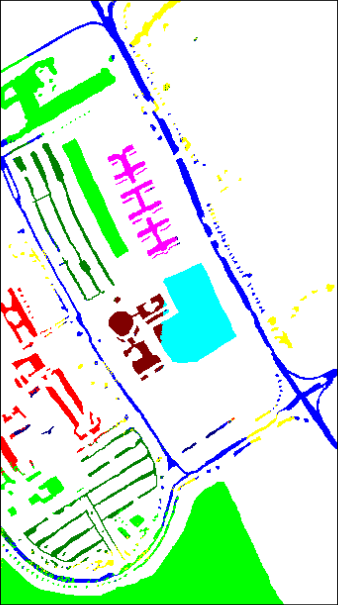}
	}
	\subfloat[DeepMorphNet Prediction]{
		\includegraphics[height=\exFigPaviaU\textwidth]{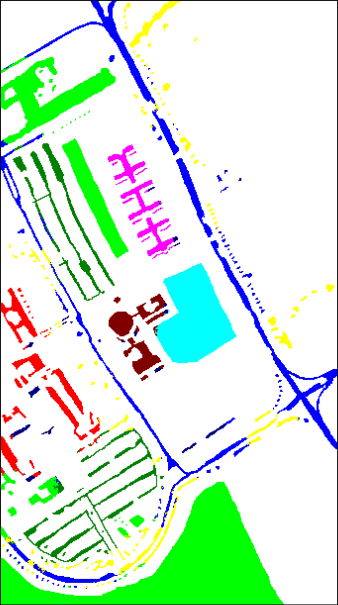}
	}
	\caption{The pixel classification datasets and the prediction maps generated by the proposed algorithm and baselines. Images (a)-(d) are related to the Pavia Centre dataset whereas images (e)-(h) are associated to the Pavia University dataset. The color encoding of the classes is presented in Figure~\ref{fig:hyperspectral_datasets}.
	}
	\label{fig:pavia_results}
\end{figure*}

%\newcommand{\exFigPavia}{0.17}
%\begin{figure}[h!]
%	\centering
%	\subfloat[False-color Image]{
%		\includegraphics[height=\exFigPavia\textwidth]{pavia_centre/false_color.png}
%	}
%	\subfloat[Ground-Truth]{
%		\includegraphics[height=\exFigPavia\textwidth]{pavia_centre/gt.png}
%	}
%	\subfloat[ConvNet Prediction]{
%		\includegraphics[height=\exFigPavia\textwidth]{pavia_centre/pred_convnet_border.png}
%	}
%	\subfloat[DeepMorphNet Prediction]{
%		\includegraphics[height=\exFigPavia\textwidth]{pavia_centre/pred_morphnet_border.png}
%	}
%	\caption{The GRSS Data Fusion training and test images, their respective ground-truths and the prediction maps generated by the proposed algorithm. Legend -- Black: unclassified. Light purple: road. Light green: trees. Red: red roof. Cyan: gray roof. Dark purple: concrete roof. Dark green: vegetation. Yellow: bare soil.
%	}
%	\label{fig:pavia_results}
%\end{figure}

\subsubsection{Pavia University Dataset} \label{subsubsec:pavia_university}

Results for the Pavia University dataset are reported in (the second part of) Table~\ref{tab:pixelclass_results}.
Again, the proposed DeepMorphNet outperformed the ConvNet model by approximately 5 percentage points of average accuracy (86.52\% versus 82.44\%, respectively).
% This outcome corroborates with previous analysis on the feature learning ability of the proposed technique.
% Furthermore, these results show that, in some scenarios (such as this one), the features learned by the DeepMorphNet can make this networks to.
This gain can be also seen in the prediction maps (Figure~\ref{fig:pavia_results}).
Overall, the proposed method was capable of producing better results for some specific classes, such as Trees and Shadows (yellow and dark blue colors, respectively), an outcome that can be better observed in Figure~\ref{fig:cms}.

\newcommand{\exCM}{0.23}
\begin{figure*}[h!]
	\centering
	% trim={<left> <lower> <right> <upper>}
	\subfloat[ConvNet]{
		\includegraphics[trim={6em 1.7em 9em 3em},clip,height=\exCM\textwidth]{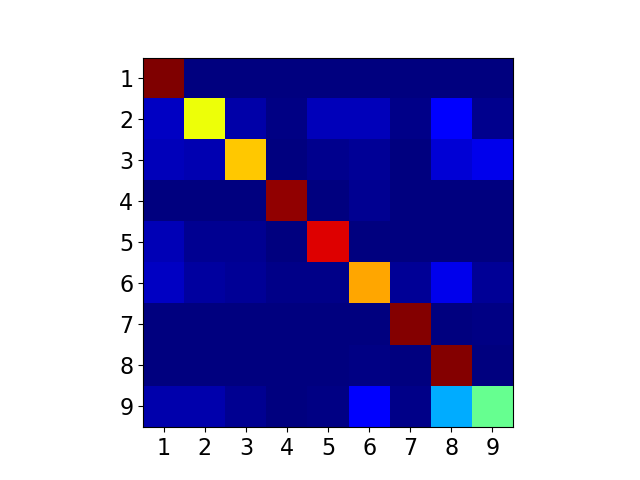}
	}
	\subfloat[DeepMorphNet]{
		\includegraphics[trim={6em 1.7em 9em 3em},clip,height=\exCM\textwidth]{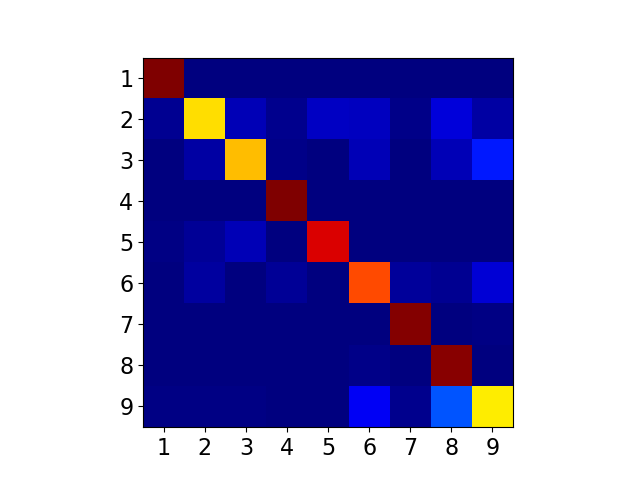}
	}
	\subfloat[ConvNet]{
		\includegraphics[trim={6em 1.7em 9em 3em},clip,height=\exCM\textwidth]{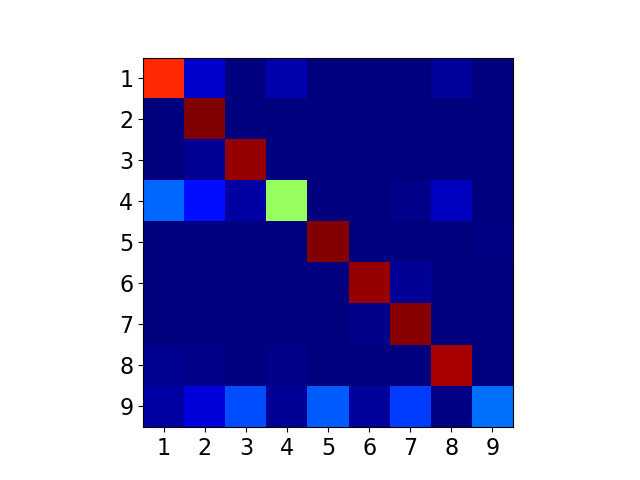}
	}
	\subfloat[DeepMorphNet]{
		\includegraphics[trim={6em 1.7em 4.5em 3em},clip,height=\exCM\textwidth]{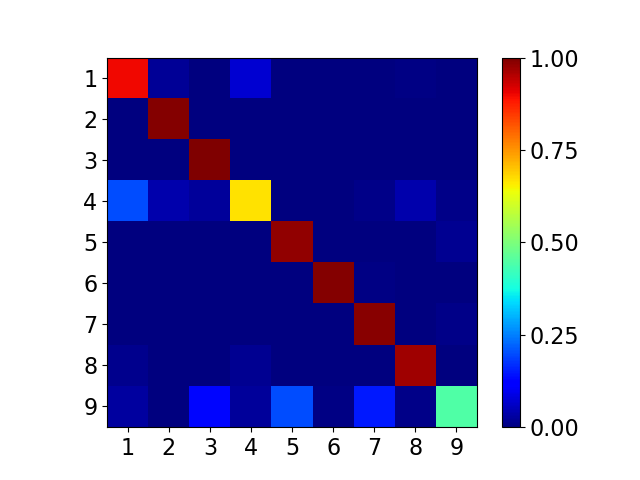}
	}
	\caption{Normalized Confusion matrices of the proposed method and baseline for both hyperspectral pixel classification datasets. Images (a) and (b) are related to the Pavia Centre dataset whereas images (c)-(d) are associated to the Pavia University dataset.
	}
	\label{fig:cms}
\end{figure*}

\section{Conclusion} \label{sec:conclusion}

In this proof of concept work, we proposed a new paradigm for deep networks where linear convolutions are replaced by non-linear morphological operations.
The proposed method, called Deep Morphological Network (DeepMorphNet), is able to perform morphological operations while optimizing their structuring elements.
% toward a better solution.
Technically, the proposed approach is composed of morphological layers, which consist of morphological neurons.
Such processing units are built upon a framework that is essentially based on depthwise convolution and pooling layers.
%(that process the input with decomposed binary weights) and pooling layers.
In fact, this framework provides support for the creation of the basic morphological neurons that perform erosion and dilation.
These, in turn, allow the creation of other more complex ones that perform opening, closing, top-hats, and (an approximation of) reconstructions.
The proposed approach is trained end-to-end using standard algorithms employed in deep networks.
%, including backpropagation and Stochastic Gradient Descent (SGD)~\cite{goodfellow2016deep}.
%Aside from the mere implementation, such layers are optimized, in terms of their parameters (i.e., filter or structuring elements), via backpropagation and Stochastic Gradient Descent (SGD)~\cite{goodfellow2016deep}.

Experiments were conducted using \textbf{six} datasets: two synthetic, two image, and two pixel classification ones.
%%%It is important to observe that the first two were only employed to analyze the feature learning of the proposed technique. % as well as its efficiency.
The first two were only employed to analyze the feature learning of the proposed technique whereas the other datasets were employed to assess the efficiency of the DeepMorphNets.
Results over the synthetic datasets have shown that the proposed DeepMorphNets are able to learn relevant structuring elements perfectly classifying them.
%In fact, the method could learn the expect (a priori) structuring element.
%Furthermore, the proposed approach learned a perfect classification of both datasets outperforming or producing equal results when compared to the ConvNets.
%%%This result shows the potential of DeepMorphNets, which can learn important filters.
% that are also different from those learned by the ConvNets.
%%%Then, the DeepMorphNet was analyzed using the two image classification datasets.
Considering the image classification datasets, the proposed DeepMorphNets outperform other similar deep morphological frameworks~\cite{mondal2020image,franchi2020deep} whereas producing competitive results when compared to ConvNets with equivalent architectures.
%(i) UCMerced Land-use Dataset~\cite{mercedlanduse}, composed of aerial high-resolution scenes in the visible spectrum, and
%(ii) WHU-RS19 Dataset~\cite{xia2010structural}, composed of high-resolution very distinct scenes.
%%%In both cases, the DeepMorphNets produced competitive results when compared to ConvNets with equivalent architectures.
Finally, for the pixel classification datasets, the proposed method outperformed (in approximately 5 percentage points of average accuracy) standard convolutional networks.
%%%Finally, the proposed technique was evaluated using two hyperspectral pixel classification datasets.
%%%In both cases, the proposed method outperformed standard convolutional networks. %, showing the ability of this technique in learning features.
In general, a conclusion common to all datasets is that the morphological networks are capable of learning relevant filters and extracting salient features.
%the obtained results shows that the morphological networks are capable of learning relevant filters and extracting salient features.

% and, consequently, distinct features.
%which shows the potential of the proposed technique.

%This would be opening the door to countless future developments, architectures, applications....
%The presented conclusions open some opportunities towards a better use and optimization of morphological operations, which may be exploited by several applications and tasks.
Although the presented conclusions may open opportunities towards a better use and optimization of non-linear morphological operations, learning the optimal shapes of the structuring elements used in each morphological neuron raises multiple scientific and technical challenges, including the high number of trainable parameters (which increase the complexity of the network), and so on.
% mainly because of their non-linearity.
%Because of the non-linearity of morphological operations, learning the optimal shapes of the structuring element used in each morphological neuron raises multiple scientific and technical challenges.
Following the proof of concept presented in this manuscript, we will continue our work trying to tackle these computational limitations in order to consider more complex architectures.
Furthermore, we plan to analyze the combination of DeepMorphNets and ConvNets and to test DeepMorphNets in different applications.

%\reftitle{References}
%
%\externalbibliography{yes}
%\bibliographystyle{unsrt}
%\bibliography{bibliography}

\bibliographystyle{IEEEtran}
\bibliography{bibliography}

\end{document}